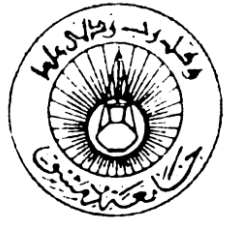

# تمييز الأنماط باستخدام النظام المناعي الاصطناعي
# Pattern Recognition using Artificial Immune System

رسالة أعدت للحصول على درجة الماجستير في هندسة الحاسبات و التحكم الآلي

إعداد :

م. محمد طارق المعلم

إشراف :

د.م. رؤوف حمـدان

د.م. رند القوتـلي

العام الدراسي
2008- 2009

بسم الله الرحمن الرحيم

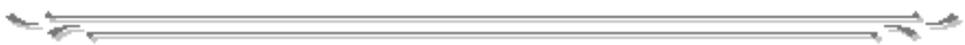

# ملخص الرسالة:
# Thesis Abstract

الهدف من هذا البحث هو دراسة أحدث التقنيات في مجال الذكاء الحوسبي Computational Intelligence (CI) هذه التقنية هي النظام المناعي الاصطناعي Artificial Immune System **(AIS)** إذ تقدم الرسالة فكرة عن ماهية النظام المناعي و كيفية الاستفادة منه في تصميم خوارزميات حاسوبية في مجال تمييز الأنماط ، حيث تعرض الرسالة عدداً من أهم النماذج و الخوارزميات المناعية المستخدمة في مجال تمييز الأنماط و منها: خوارزمية الانتقاء السلبي و خوارزمية الانتقاء النسيلي و الشبكات المناعية الاصطناعية. تضمن البحث شرحاً كاملاً لتلك الخوارزميات و تطبيقاتها و التحسينات التي تمت عليها ، كما قمنا أيضاً ضمن إطار هذا البحث باختبار تلك الخوارزميات المناعية مثل خوارزمية الانتقاء النسيلي و الشبكة المناعية الاصطناعية إضافة إلى دراسة اثر تغير معاملات تلك الخوارزميات على أداء الخوارزمية و مدى أهمية هذه المعاملات في تحسين الأداء. و بناءً على دراسة النظام المناعي الاصطناعي AIS و النماذج و الخوارزميات المناعية المطورة تم تصميم خوارزمية مناعية تعتمد على مبدأ الانتقاء النسيلي في النظام المناعي البيولوجي للقيام بعملية تصنيف غير موجه للمعطيات unsupervised classification. الخوارزمية الجديدة هي خوارزمية متكيفة مع المعطيات و تعمل علي تغيير معاملاتها أوتوماتيكياً مع المعطيات من اجل تحسين الأداء و تسريع الوصول إلى الحل و قد تم اختبار أدائها و مقارنته مع الخوارزميات المعروفة في مجال التصنيف غير الموجه مثل *K*-means حيث وجدنا من نتائج الاختبار أن الخوارزمية الجديدة كانت أكثر دقة في التصنيف و أكثر وثوقية.

# كلمة شكر:
# Acknowledgements

أتوجه بالشكر إلى من أخذوا بيدي و سددوا خطاي و كانوا لي آباءً في العلم و الإرشاد ، إلى جميع أساتذتي و أخص بالذكر:

الدكتور المهندس رؤوف حمدان

و الدكتور المهندس رند القوتلي

الذين تكرموا بالإشراف على البحث و كانوا لي خير معين................

كما أتوجه بالشكر إلى رئيس قسم هندسة الحواسيب و الأتمتة السابق الدكتور المهندس مازن المجايري و رئيس القسم الحالي الدكتور المهندس سالم مرزوق لمساعدتهم و دعمهم المتواصل......................

## الفهرس:

# لائحة الأشكال:

## لائحة الجداول:

## لائحة الاختصارات:

| | | |
|---|---|---|
| AI | Artificial Intelligence | الذكاء الصنعي |
| CI | Computational Intelligence | الذكاء الحوسبي |
| FS | Fuzzy Systems | الأنظمة العائمة |
| ANN | Artificial Neural Networks | الشبكات العصبونية الاصطناعية |
| EC | Evolutionary Computation | الحوسبة التطورية |
| AIS | Artificial Immune System | النظام المناعي الاصطناعي |
| MHC | major histocompatibility complex | معقد التوافق النسيجي الكبير |
| CDRs | complementarity-determining regions | مناطق تحديد المتمم |
| APCs | antigen presenting cells | الخلايا المقدمة للمستضد |
| TCRs | T cell receptors | مستقبلات الخلايا T |
| GCs | germinal centers | المراكز المنتشة |
| NS | Negative Selection | الانتقاء السلبي |
| CSA | Clonal Selection Algorithm | خوارزمية الانتقاء النسيلي |
| CLONALG | Clonal Algorithm | |
| CLONCLAS | Clonal Classification | التصنيف النسيلي |
| aiNet | Artificial Immune Network | الشبكة المناعية الاصطناعية |
| opt-CLONALG | optimization CLONALG | CLONALG من اجل الأمثَلة |
| opt-IA | optimization Immune Algorithm | خوارزمية مناعية للأمثَلة |
| opt-IMMALG | Ver2 of optimaization Immune Algorithm | نسخة ثانية من خوارزمية مناعية للأمثَلة |
| RECSA | receptors edidting Clonal Selection Algorithm | خوارزمية الانتقاء النسيلي مع تحرير للمستقبلات |
| FCA | Fast Clonal Algorithm | خوارزمية الانتقاء النسيلي السريعة |
| UCSC | Unsupervised Clonal Selection Classification | تصنيف الانتقاء النسيلي غير الموجه |

# لائحة الرموز:
## List of Variables and Notations

**رموز عامة:**

| | |
|---|---|
| $V_\varepsilon$ | منطقة التمييز |
| m | الشكل العام لجزيء (سواء كان ضد أو مستضد) |
| $L$ | عدد أبعاد الفضاء |
| Ab | ضد |
| Ag | مستضد |
| $S^L$ | فضاء الشكل |
| $\varepsilon$ | عتبة التآلف |
| $D$ | المسافة |
| $K$ | عدد العناقيد |
| $x$ | المعطيات |
| $x_i^{(K)}$ | نمط ينتمي إلى العنقود $K$ |
| $C_K$ | العنقود $K$ |
| $n_k$ | عدد الأنماط ضمن العنقود $K$ |
| $m^{(k)}$ | مركز العنقود $K$ |
| $E^2{}_k$ | الخطأ التربيعي |

**رموز الانتقاء السلبي:**

| | |
|---|---|
| M | مجموعة الكواشف (الذاكرة المناعية) |
| P | المجموعة المحمية |
| C | مجموعة من العناصر المقترحة |
| P* | مجموعة من العناصر الجديدة |

**رموز الانتقاء النسيلي:**

| | |
|---|---|
| Ag | مجموعة المستضدات |
| m | حجم مجموعة المستضدات |
| Ab | مجموعة الأضداد (الحلول المقترحة) |
| N | حجم مجموعة الأضداد |
| $Ab_m$ | مجموعة الذاكرة المناعية |
| $Ab_r$ | مجموعة باقي الحلول |
| C | مجموعة مؤقتة من النسائل |
| C* | مجموعة الخلايا الناضجة |
| f | شعاع قيم التآلف |
| $Ab_{mc}$ | المرشح ليكون الذاكرة المناعية الجديدة |
| n | عدد الحلول المنتقاة للاستنسال |
| Nc | حجم النسيلة |
| β | معامل الاستنسال |
| gen | عدد الأجيال |
| $\alpha$ | معدل احتمال الطفرة |
| $\rho$ | معامل الهبوط لمنحني الطفرة |
| d | عدد الحلول العشوائية الجديدة |
| k | عدد الخلايا المستبدلة |

## رموز aiNet:

| | |
|---|---|
| Np | حجم المعطيات |
| c | مصفوفة خلايا الشبكة |
| Nt | حجم مجموعة خلايا الشبكة |
| M | مصفوفة خلايا ذاكرة |
| N | حجم مصفوفة خلايا ذاكرة |
| Nc | عدد الخلايا المستنسلة من الخلية المحفزة (المفعلة) |
| D | مصفوفة عدم التماثل |
| S | مصفوفة التماثل |
| n | عدد الخلايا المنتقاة للاستنسال |
| ى | نسبة الخلايا المنتقاة من الخلايا الناضجة |
| $\sigma_d$ | عتبة الموت الطبيعي |
| $\sigma_s$ | عتبة التثبيط |
| $N_{gen}$ | عدد الأجيال |

## رموز **UCSC**:

| | |
|---|---|
| $D$ | معيار العنقدة |
| $m_i$ | مركز العنقود |
| $C_i$ | العنقود |
| $P$ | مجموعة الحلول |
| $N$ | حجم مجموعة الحلول |
| $P_C$ | مجموعة الخلايا المستنسلة |
| $P_m$ | مجموعة الخلايا الناضجة |
| $d$ | عدد الحلول العشوائية الجديد |
| $D_{intra}(C_j)$ | المسافة الداخلية لكل عنقود |
| aff | التآلف المقاس |
| $\beta$ | هو معامل الاستنسال |
| $Ab^*$ | الضد الناتج عن تطبيق الطفرة على الضد $Ab_i$ |
| $N(0,1)$ | مصفوفة من $L*K$ متحول غوصي |
| $\alpha$ | معامل يقييس مقدار الطفرة الغوصية |
| $\rho$ | معامل يحدد مجال $\alpha$ |
| $max_{data}$ | اكبر قيمة للخواص المشكلة للمعطيات |
| $min_{data}$ | اصغر قيمة للخواص المشكلة للمعطيات |
| $UL_i$ | الحدود العليا للخواص المحددة للمعطيات |
| $LL_i$ | الحدود الدنيا للخواص المحددة للمعطيات |
| rand | مصفوفة من $L*K$ متحول عشوائي ذو توزع احتمالي متساوي |
| $\sum_i$ | مصفوفات التغاير |

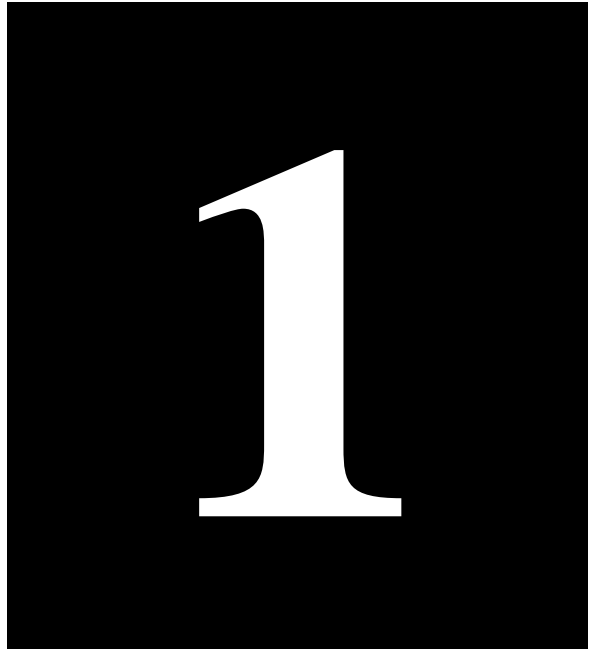

# مقدمـة

# 1- مقدمـة:

## 1-1- الدافع إلى البحث : Thesis Motivation

يعتبر الذكاء الصنعي Artificial Intelligence (AI) من أهم المجالات التي تتطور بشكل سريع وهو مجال له العديد من التطبيقات المهمة في الحياة العملية. يهتم الذكاء الصنعي بجعل الآلة تتصرف مثل الإنسان و جعلها قادرة على اتخاذ القرارات و بشكل عام يشمل AI التفكير و المعرفة و التخطيط و التعلم و التواصل و الإدراك و القدرة على التعامل مع الغرض.

يعتبر الذكاء الحوسبي Computational Intelligence (CI) [1] [2] احد فروع AI و هو يهتم بتطوير الأنظمة المتكيفة و القادرة على التعلم و اكتساب الخبرة من المعطيات المتوفرة بحيث تستطيع إعطاء الحلول الجديدة بدون الاعتماد الصريح على المعرفة الإنسانية و بحيث يمتلك النظام بعض صفات المنطق مثل : التعميم و الاكتشاف و ربط الذكريات و التجريد [1] [2].

تَستخدم أنظمة الذكاء الحوسبي فعلياً جميع الأبحاث في مجال الأنظمة العائمة Fuzzy Systems (FS) و الشبكات العصبونية الاصطناعية Artificial Neural Networks (ANN) و الحوسبة التطورية Evolutionary Computation (EC). و حالياً تمت إضافة مجال جديد ضمن مجالات أبحاث CI هو النظام المناعي الاصطناعي Artificial Immune System (AIS) و هو يهتم بتطوير خوارزميات و أنظمة حاسوبية مستوحاة من عمل النظام المناعي (عند ذكر عبارة النظام المناعي فقط يقصد بها البيولوجي) [3] [4].

و قد أظهرت تلك الخوارزميات جدارتها في العديد من المجالات و منها مجال تمييز الأنماط و تعلم الآلة و ذلك لكون النظام المناعي هو أصلاً نظام قادر على التعلم. و ما تزال هذه التقنية الجديدة في بداياتها و قيد البحث و ليس لها حتى الآن إطار عام للعمل.

و من هذا المنطلق كان هدف البحث هو دراسة AIS في مجال تمييز الأنماط كبداية للتعرف على هذه التقنية الحديثة و من ثم محاولة تطوير و تصميم خوارزميات مناعية تفيد في مجال تمييز الأنماط.

## 1-2- فكرة عن الرسالة:

يعتمد البحث على دراسة النظام المناعي البيولوجي و كيفية الاستفادة منه في تصميم خوارزميات حاسوبية في مجال تعلم الآلة. و تعرض الرسالة أهم النماذج و الخوارزميات المناعية المستخدمة

في مجال تمييز الأنماط ، وهي تحتوي على شرح كامل لتلك الخوارزميات و تطبيقاتها و التحسينات التي تمت عليها ، و من هذه الخوارزميات: خوارزمية الانتقاء السلبي و خوارزمية الانتقاء النسيلي و الشبكات المناعية الاصطناعية.

كما قمنا أيضاً ضمن إطار هذا البحث باختبار تلك الخوارزميات المناعية مثل خوارزمية الانتقاء النسيلي و الشبكة المناعية الاصطناعية إضافة إلى دراسة اثر تغير معاملات تلك الخوارزميات على أداء الخوارزمية و مدى أهمية هذه المعاملات في تحسين الأداء.

و بناءً على دراسة النماذج و الخوارزميات المناعية المطورة تم تصميم خوارزمية تعتمد على مبدأ الانتقاء النسيلي في النظام المناعي البيولوجي للقيام بعملية تصنيف غير موجه للمعطيات unsupervised classification ، الخوارزمية المطورة هي خوارزمية متكيفة مع المعطيات و تعمل على تغيير معاملاتها أوتوماتيكياً من اجل تسريع الوصول إلى الحل و قد تم اختبارها و مقارنة أدائها مع الخوارزميات المعروفة في مجال التصنيف غير الموجه مثل K-means و وجد من النتائج أن الخوارزمية المطورة (تصنيف الانتقاء النسيلي غير الموجه Unsupervised Clonal Selection Classification) كانت أكثر دقة من حيث التصنيف و أكثر وثوقية.

### 1-3- موجز لمحتويات الفصول القادمة:

في الفصل الثاني تم شرح النظام المناعي البيولوجي و مكوناته و طبقات الحماية التي يوفرها ومن ثم آليات عمله و النظريات الأساسية للاستجابة المناعية حيث شكل مدخلاً إلى الفصل الثالث الذي يعطي فكرة عن النظام المناعي الاصطناعي و مجالات استخدامه و كيفية تصميم نظام مناعي اصطناعي لتطبيق معين ، في الفصل الرابع قمنا بشرح الطرق الأساسية لتمييز الأنماط ، في الفصل الخامس تمت دراسة الخوارزميات المناعية المستخدمة في مجال تمييز الأنماط ، في الفصل السادس تم وضع نتائج الاختبارات التي قمنا بها على الخوارزميات المناعية المستخدمة في مجال تمييز الأنماط و من ثم قمنا بشرح الخوارزمية المناعية التي قمنا بتصميمها لتقوم بعملية التصنيف غير الموجه كما تم اختبار أدائها و مقارنته مع أداء الخوارزميات المستعملة في هذا المجال حيث أعطت الخوارزمية الجديدة نتائج جيدة.

# الفصل الثاني: النظام المناعي البيولوجي

# 2- النظام المناعي البيولوجي: The Immune System (IS)

## 2-1- مقدمة:

النظام المناعي هو نظام معقد من الخلايا و الجزيئات و الأعضاء و مهمته الدفاع عن الجسم و حمايته من أي جسيم غريب و مؤذي مثل: البكتيريا و الطفيليات و الفيروسات ، تدعى الجسيمات الغريبة عن الجسم بالمستضدات Antigens ، يمتلك النظام المناعي القدرة على إدراك أي خلل يصيب الجسم أو مرض سواء كان هذا المرض ذو منشأ داخلي infectious self (ناتج عن خلل في خلايا الجسم) أو ذو منشأ خارجي infectious nonself (ناتج عن جسيم غريب عن الجسم) [5]. بدون النظام المناعي فإن مصير الجسم هو الموت المحتم نتيجة أي مرض يصيب الإنسان.

كما أن النظام المناعي قادر على التمييز بين الجزيئات الغريبة عن الجسم nonself و الجزيئات من الجسم self molecules و هذا الأمر مهم جداً و هو يمنع النظام المناعي من تخريب خلايا الجسم إذ أن النظام المناعي يهاجم الخلايا و الأجسام الغريبة فقط و لا يهاجم خلايا الجسم [5].

سيتم في هذا الفصل إعطاء فكرة عن النظام المناعي البيولوجي و مكوناته و طبقات الحماية التي يوفرها ومن ثم آليات عمله و النظريات الأساسية للاستجابة المناعية. سوف نحيط بمعظم الجوانب في النظام المناعي لأنها تشكل أساساً للمعلومات ضمن الفصول القادمة (حول النظام المناعي الاصطناعي).

## 2-2- عناصر النظام المناعي:

### 2-2-1- أعضاء النظام المناعي:

يتكون النظام المناعي من عدة أعضاء إضافة إلى مجموعة من النسج و هي موزعة في الجسم و تدعى الأعضاء اللمفية lymphoid organs ومهمتها توليد الكريات البيضاء و اللمفاوية.

تنقسم أعضاء النظام المناعي إلى قسمين [4] [5]:

1. رئيسي: يعمل على توليد الخلايا المناعية الجديدة و مساعدتها على النضوج مثل نقي العظم Bone Marrow و التيموس thymus.
2. ثانوي: لتسهيل عملية التفاعل بين الخلايا و المستضد مثل الطحال spleen و العقد اللمفاوية lymph nodes.

يبين الشكل (2-1) أعضاء النظام المناعي و توزعها في الجسم.

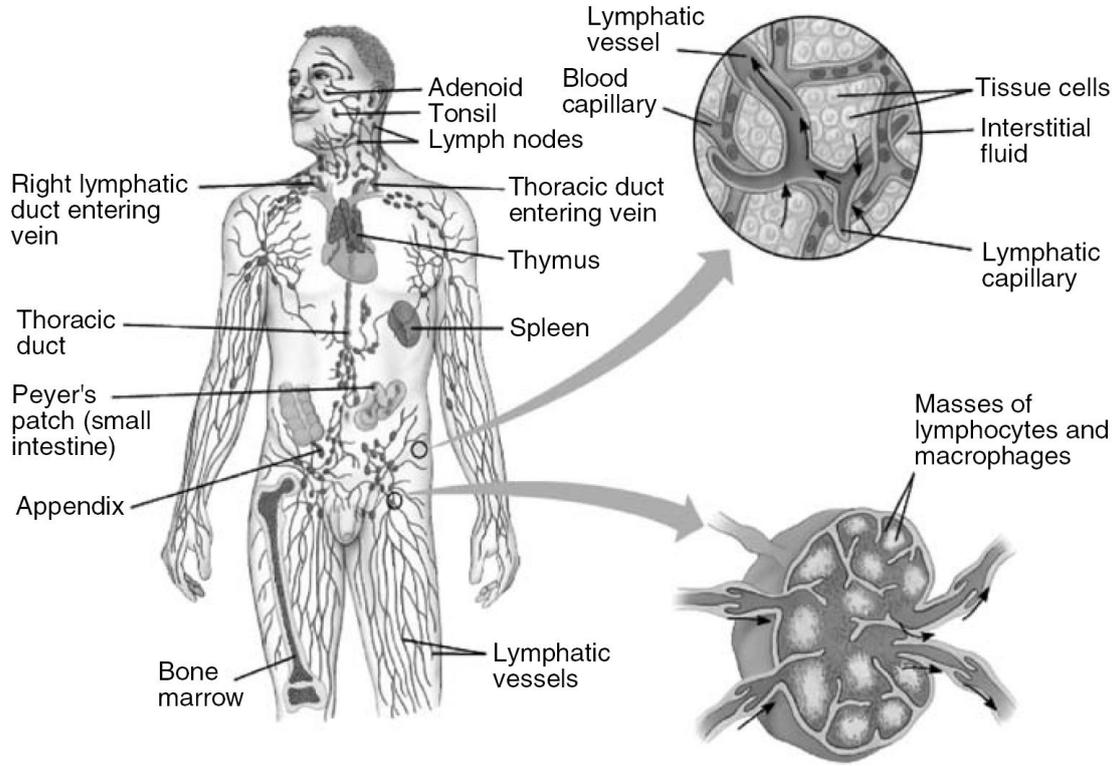

الشكل (2-1) أعضاء النظام المناعي [4]

### 2-2-2 - الخلايا المناعية: The Immune Cells

يحوي النظام المناعي العديد من الخلايا المناعية المختلفة و هي مختلفة من حيث البنية و الوظيفة لكنها تتفاعل مع بعضها البعض لتحقيق الاستجابة المناعية المطلوبة ، تقسم الخلايا المناعية إلى قسمين رئيسين هما: الخلايا اللمفاوية و البلعميات [4] [5].

يوضح الشكل (2-2) الخلايا المناعية و تقسيماتها.

### 1-2-2-2- الخلايا اللمفاوية : Lymphocytes

الخلايا اللمفاوية هي أهم خلايا النظام المناعي لأنها مسؤولة عن الاستجابة المناعية و هي تضم الخلايا T و الخلايا B و الخلايا الفاتكة الطبيعية [5].

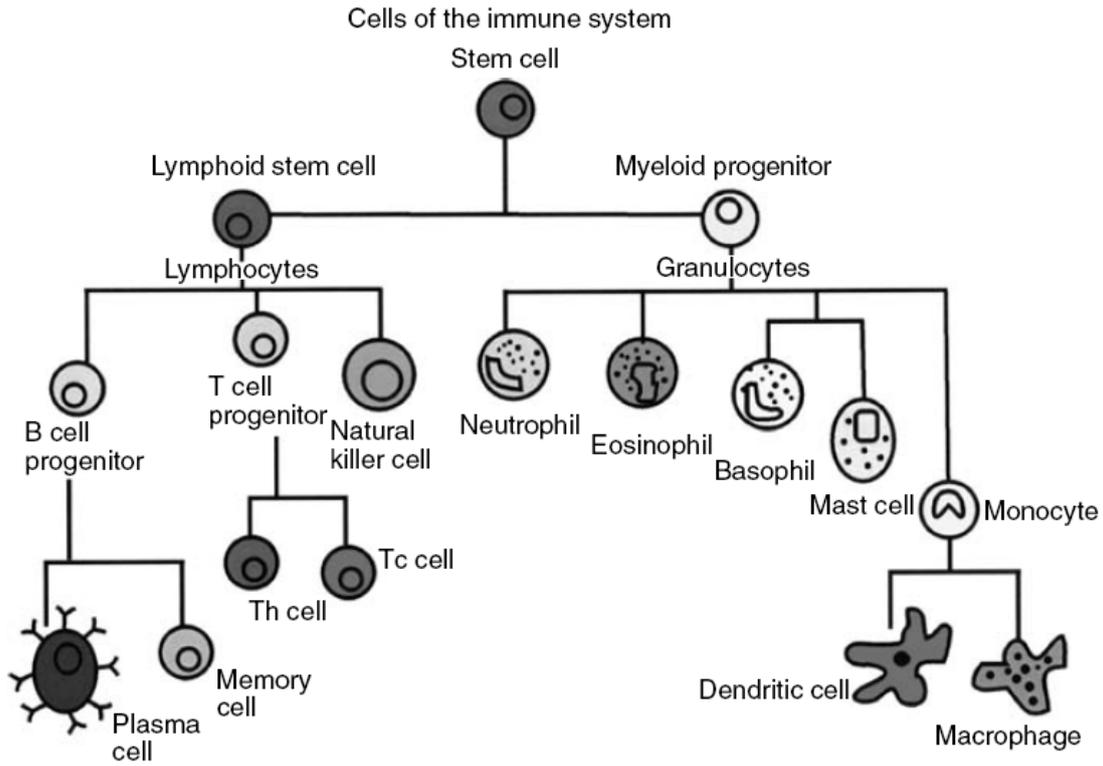

الشكل (2-2) الخلايا المناعية [4]

1. **الخلايا B** : مهمتها إنتاج نوع خاص من البروتينات هو الضد الذي يتحد مع بروتينات أخرى موجودة على العامل الممرض مما يحيد العامل الممرض و يوقفه عن العمل و يحفز النظام المناعي على ابتلاع الرابط الناتج [5].

2. **الخلايا T** تقسم إلى ثلاث أقسام [5]:

- **خلايا T المساعدة**: و هي تساعد في تفعيل خلايا النظام المناعي و تحفزها على العمل و هي مسؤولة عن تحفيز الخلايا التالية: الخلايا B و الخلايا T و الخلايا البلعمية Phagocytes و الخلايا الفاتكة الطبيعية.

- **خلايا T القاتلة**: و مهمتها القضاء على المكروبات و الفيروسات و الخلايا السرطانية إذ عند تفعيلها تقوم بحقن مادة كيميائية ضارة تخرب الخلايا.

- **خلايا T الكابتة**: مهمتها تنظيم الاستجابة المناعية حيث تعمل على تثبيط الخلايا المناعية الأخرى.

### 3. الخلايا الفاتكة الطبيعية :

هي احد أنواع الخلايا اللمفاوية ، تحوي حبيبات مملوءة بمادة كيماوية سامة ، سميت بالفاتكة الطبيعية لأنها تهاجم من دون الحاجة إلى المستضد لبدء العمل حيث تهاجم الأورام و تفرز كمية كبيرة من اللمفوكينات Lymphokines و تساهم في تنظيم المناعة [5].

### 2-2-2-2- البلعميات و المحببات : Phagocytes & Granulocytes

البلعمية كرة بيضاء قادرة على ابتلاع و هضم المكروبات و المستضدات ، لها عدة أشكال منها: الوحيدة monocytes و البالعات الكبيرة macrophages . تدخل الوحيدة الأنسجة ثم تتطور إلى بالعة كبيرة التي تقوم بتقديم المستضد إلى خلايا T بعد هضمه و تصبح خلايا مقدمة للمستضد لذلك تلعب دور مهم في بداية الاستجابة المناعية [3] [5]. يوضح الشكل(2-3) بعض الخلايا البلعمية.

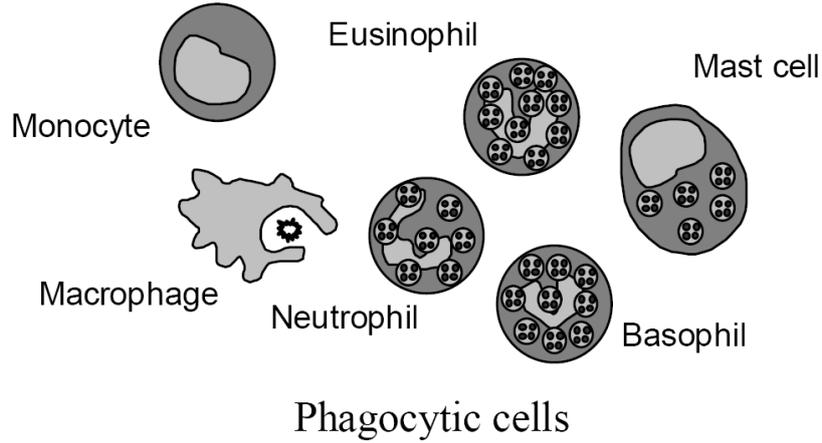

الشكل (2-3) الخلايا البلعمية [5]

### 2-2-3- نظام المتممة: The Complement System

يتألف نظام المتممة من مجموعة من البروتينات التي تدور ضمن البلازما في الدم (حوالي 25 بروتين) تقوم هذه البروتينات بالارتباط مع سطح الجسم الغازي و تخربه و من ثم تقوم سلسلة من التفاعلات تؤدي إلى ارتباطها مع بعضها لتشكل سلسة طويلة و هذا يسهل كشفها و ابتلاعها من قبل البلعميات Phagocytes [3] [4] . يوضح الشكل (2-4) عملية تفاعل نظام المتممة.

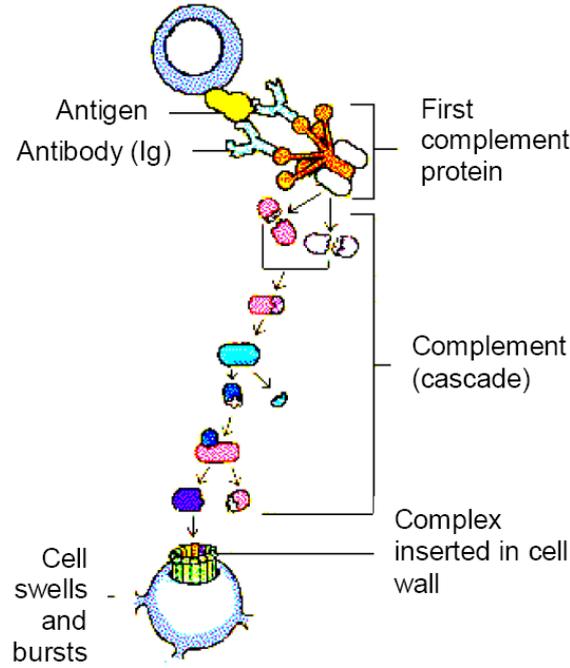

الشكل (2-4) نظام المتممة [3]

## 2-2-4- الجزيئات المناعية : Immune Molecules

### 2-2-4-1- السيتوكينات و اللمفوكينات : Cytokines & Lymphokines

السيتوكينات Cytokines هي مجموعة من البروتينات التي تفرزها بعض الخلايا المناعية و ذلك بغرض السيطرة على الخلايا الأخرى و هي غالباً تشكل رسائل كيماوية بين الخلايا حيث تتصل مع غشاء الخلية الهدف. يتم إفراز السيتوكينات Cytokines من قبل خلايا أخرى غير الخلايا المناعية لكنها تفرز بشكل أساسي من قبل الخلايا اللمفاوية و تدعى عندها باللمفوكينات Lymphokines و هي ذات تأثير كبير على سير عملية الاستجابة المناعية. اللمفوكينات التي يتم إفرازها بغرض التخاطب مع الخلايا الأخرى تدعى انترلوكينات Interleukins [4].

### 2-2-4-2- معقد التوافق النسيجي الكبير : MHC complex

معقد التوافق النسيجي الكبير MHC هو مركب يوجد في الغشاء الخلوي لخلايا الجسم و يلعب دور هاماً في عملية التمييز بين خلايا الجسم و العناصر الغريبة عن الجسم ، حيث يحدد هذا المركب خلايا الجسم الشيء الذي يمنع خلايا T من مهاجمتها لكن عند دخول مستضد للخلية يحدث خلل ضمن هذا المركب و تدخل بنية المستضد ضمنه ليصبح MHC-Ag و بالتالي تميز خلايا T الخلل و تقوم بمهاجمة الخلية [4] [5].

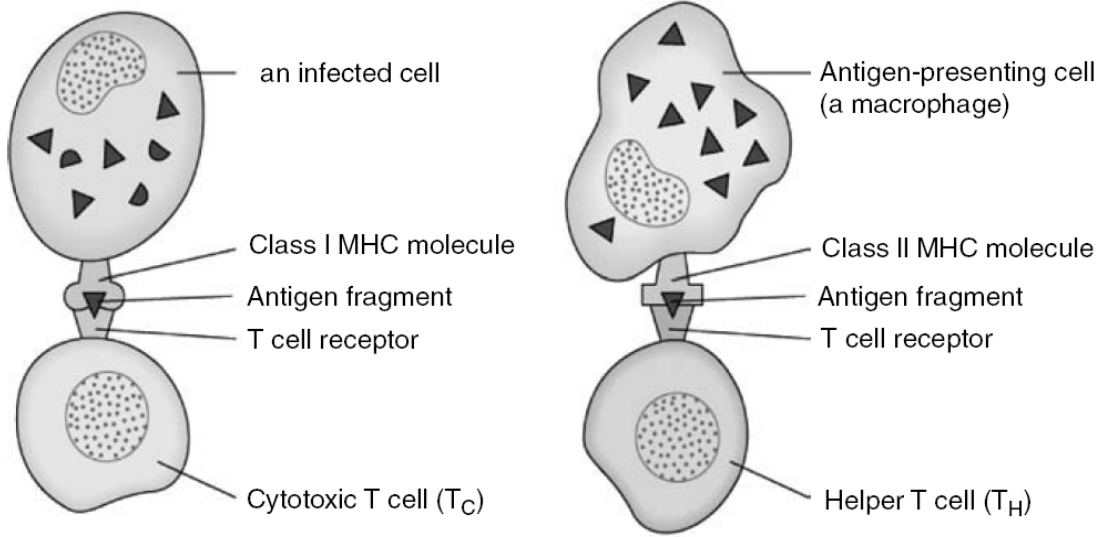

الشكل (2-5) أصناف معقد التوافق النسيجي الكبير [4]

هناك نوعان لـ MHC هما MHC-I و MHC-II [4] [5]:

1. الصنف I : يوجد في أي خلية تصاب بفيروس و تقوم خلايا T السامة للخلايا بالتفاعل مع هذا الصنف و تميز الرابط بين المستضد و MHC-I.
2. الصنف II : يوجد فقط في الخلايا المقدمة للمستضد مثل البالعات الكبيرة التي تهضم جزئيا المستضد و تفككه إلى ببتيدات و تربطه مع MHC-II ، تتفاعل خلايا T المساعدة مع هذا المركب.

يوضح الشكل (2-5) أصناف معقد التوافق النسيجي الكبير و الخلايا التي تصنعه.

### 2-2-4-3- جزيء الضد : The Antibody Molecule

الأضداد هي بروتينات أو غلوبولينات مناعية immunoglobulins توجد في الدم و هي تنتج بواسطة الخلايا B الناضجة أو ما يدعى الخلايا البلازمية، تعمل الاضداد على تمييز نمط جزيئي خاص من المستضد [3] [4] [5].

يتألف الضد من أربع سلاسل متعددة الببتيد polypeptide : سلسلتان خفيفتان (L) و سلسلتان ثقيلتان (H) و كل سلسلة مؤلفة من منطقة ذات بنية متغيرة (V) و منطقة ذات بنية ثابتة (C). المنطقة المتغيرة V هي المسؤولة عن تمييز المستضد و تحوي مناطق فرعية متغيرة هي التي تتصل بالمستضد تدعى مناطق تحديد المتمم complementarity-determining regions (CDRs). المنطقة الثابتة C لها دور في تثبيت الضد على المستضد[3] [4] [5].

يوضح الشكل (2-6) بنية الضد و منطقة الارتباط بين الضد و المستضد.

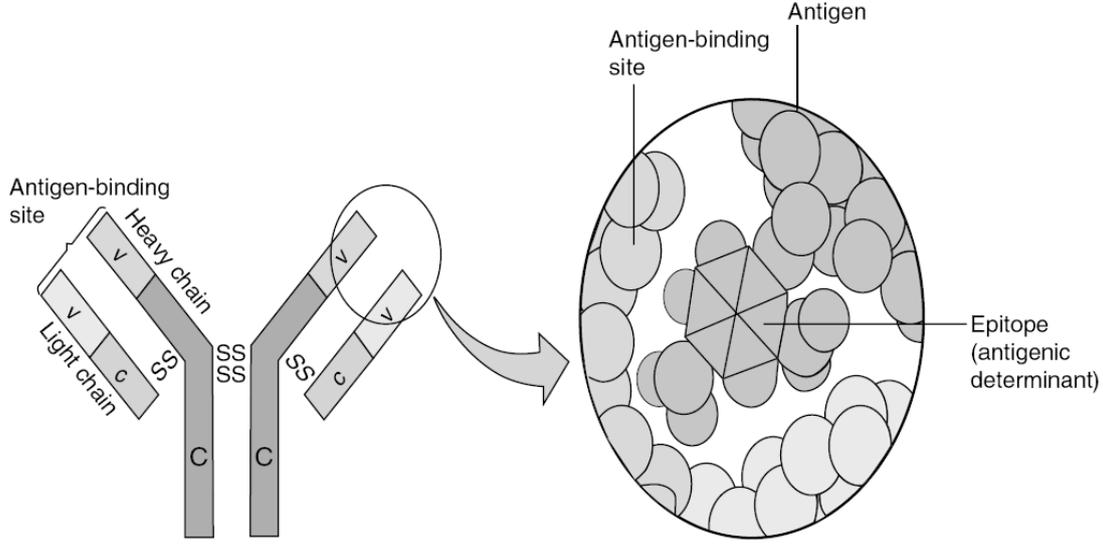

الشكل (2-6) بنية الضد و المنطقة المتغيرة [4]

## 2-3- طبقات النظام المناعي:

يمكن النظر إلى النظام المناعي على انه نظام متعدد الطبقات حيث كل طبقة تحوي نوعاً مختلفاً من آليات الدفاع عن الجسم. يتألف النظام المناعي من ثلاث طبقات رئيسية هي: الغشاء التشريحي Anatomic Barrier و المناعة الفطرية و المناعة المكتسبة [3]. يوضح الشكل (2-7) طبقات الحماية التي يقدمها النظام المناعي.

### 2-3-1- الغشاء التشريحي: Anatomic Barrier

الغشاء التشريحي هو الغشاء الذي يغلف الجسم و يشكل نقطة الاتصال بين الجسم و البيئة الخارجية و هو حاجز الدفاع الاول عن الجسم ، و هو يتألف من الجلد اضافة الى بعض السوائل مثل الدمع و العرق و اللعاب و الاحماض المعوية ، يعمل الجلد على صد العديد من العوامل الممرضة ، كما تحوي سوائل مثل الدمع و اللعاب و العرق أنزيمات تعمل على قتل البكتيريا كما أن الأحماض المعوية تعمل على قتل معظم العوامل الممرضة الموجودة في الطعام [3] [4].

## 2-3-2 - نظام المناعة الفطرية: innate immune system

نظام المناعة الفطرية هي مجموعة من آليات الدفاع التي تولد مع الإنسان و هي قادرة على تمييز عدد قليل من المكروبات و العوامل الممرضة. يتكون النظام المناعي الفطري من البلعميات Phagocytes و نظام المتممة Complement System. تكمن أهمية النظام المناعي الفطري في قدرته على التمييز بين خلايا الجسم و الجسيمات الغريبة عن الجسم و هو ما يدعى التمييز بين الذاتي و غير الذاتي self/nonself كما انه يضم خلايا تدعى الخلايا المقدمة للمستضد (antigen presenting cells (APCs التي تقوم بقتل و هضم العامل الممرض و من ثم تقوم بتفعيل الخلايا T من النظام المناعي المكتسب لتبدأ عملية التفاعل المناعي بالتالي النظام المناعي الفطري هو من يعطي الأمر إلى النظام المناعي المكتسب ليبدأ الاستجابة المناعية [3] [5].

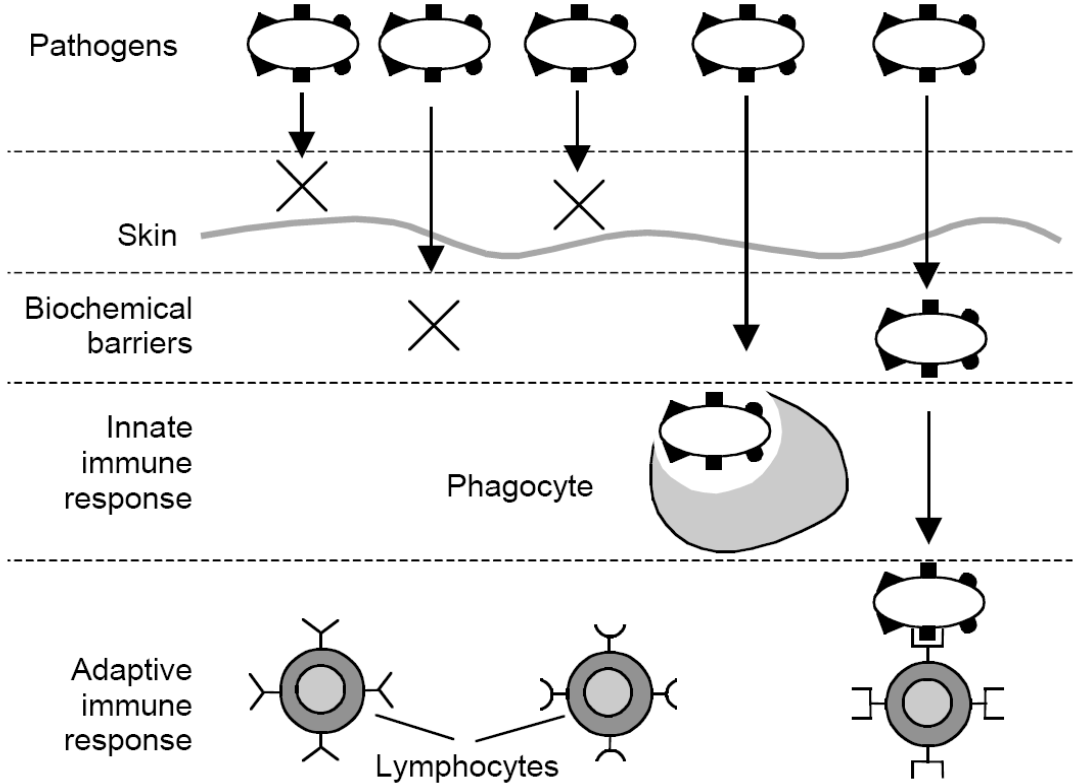

الشكل (2-7) طبقات النظام المناعي [3] [5]

## 2-3-3- نظام المناعة المكتسبة: adaptive immune system

نظام المناعة المكتسبة هو قسم من النظام المناعي له القدرة على التعرف على العوامل الممرضة الغير معروفة مسبقاً من قبل النظام المناعي و اكتساب المناعة ضدها و التخلص منها [3] [5] ، تقسم المناعة المكتسبة إلى قسمين [3] [5]:

1. المناعة الخلوية: Cellular immunity و هي تعتمد على نوعين من الخلايا اللمفاوية هما الخلايا B و الخلايا T التي تتفاعل مع العامل الممرض من اجل القيام بعملية التخلص منه.
2. المناعة الخلطية: Humoral immunity و هي تعتمد على الأضداد المولدة من قبل الخلايا B للتخلص من العامل الممرض.

## 2-4- آليات عمل النظام المناعي:

سيتم شرح الآليات التي يتبعها النظام المناعي من اجل حماية الجسم.

### 2-4-1- التمييز المناعي: Immune Recognition

جميع العمليات التي يقوم بها النظام المناعي تتطلب منه القدرة على تمييز المستضد (العامل الممرض) ، تتم عملية التمييز بواسطة المستقبلات الموجودة على غشاء الخلايا اللمفاوية (سواء كانت الخلايا B أو الخلايا T) و يعتمد التمييز على مقدار ارتباط هذه المستقبلات بالمستضد ، يتم الارتباط (التمييز) على المستوى الجزيئي و هو يعتمد على مقدار التتام (التعاكس) بين مقر الرابط للمستقبلات ما يدعى مستوقع paratope و جزء من المستضد يدعى حاتمة epitope [3] [4] [5].

يوضح الشكل (2-8) ارتباط المستقبلات بالمستضد و التتام بين المستقبل و الحاتمة.

كلما ازداد التتام بين المستقبل و الحاتمة زادت قوة الرابطة بينهما و تدعى هذه الرابطة بالتآلف ، بالتالي مقدار التمييز يتعلق بمقدار التآلف بين المستقبل على غشاء الخلية اللمفاوية و الحاتمة على المستضد و كلما زاد مقدار التآلف زادت قدرة التمييز. يوجد عناصر أخرى في النظام المناعي قادرة على تمييز المستضد و هي الأضداد التي لها نفس بنية المستقبلات على غشاء الخلية B لكنها توجد بشكل منفصل عن الخلايا و تدور ضمن الدم. بما أن المستضد يحوي عدة حواتم epitopes على سطحه و بأشكال مختلفة يتم تمييز كل مستضد من قبل عدة خلايا مختلفة [3] [5] و هذا موضح بالشكل (2-8) .

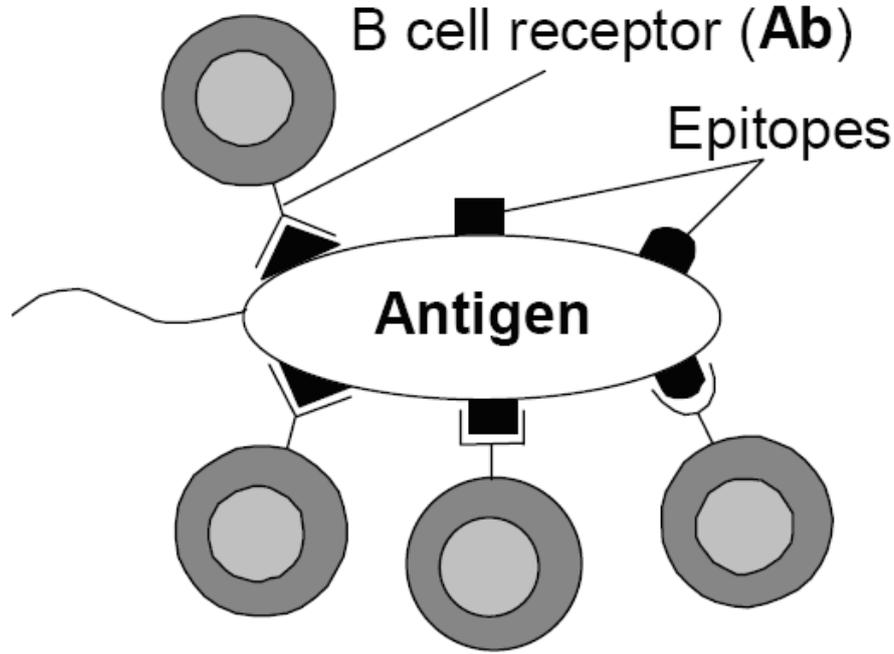

الشكل (2-8) ارتباط المستقبلات بالمستضد   [3] [5]

### 2-4-2- الاستجابة المناعية:

الاستجابة المناعية هي العمليات التي يقوم بها النظام المناعي من اجل تمييز المستضد و التخلص منه و اكتساب المناعة ضده ، تبدأ الاستجابة المناعية عند دخول مستضد للجسم إذ تطوف الجسم الخلايا المقدمة للمستضد كالبلعمية Phagocyte و عندما تعثر على مستضد تعمل على ابتلاعه و هضمه و تجزئته إلى ببتيد مستضدي ، أجزاء من هذه الببتيدات تجمع مع معقد التوافق النسيجي الكبير major histocompatibility complex (MHC) و يتم عرضه على سطح الخلية ، تمتلك خلايا T مستقبلات على سطحها تمكنها من تمييز المركب ببتيد- MHC ، عند التمييز تتفعل خلية T المساعدة و تفرز لمفوكينات Lymphokines أو انترلوكينات Interleukins الشيء الذي يؤدي إلى تحفيز عناصر النظام المناعي [3] [5] [6].

تمتلك خلايا B على سطحها مستقبلات قادرة على تمييز جزء من المستضد بدون MHC و عند تحفيز الخلايا B تعمل على تغير بنية المستقبلات على غشائها محاولة تمييز المستضد و عند حصول التمييز تتفعل الخلية B و تنقسم و تتحول إلى خلية بلازمية تفرز الأضداد التي لها نفس

بنية المستقبلات على سطح الخلية B لكن عددها أكثر بكثير من عدد الخلايا B مما يزيد من فعاليتها في مكافحة المستضد ، تتصل هذه الأضداد مع المستضد و تعدله ثم يتم تفكيك مركب ضد-مستضد الناتج بواسطة أنزيم متمم أو يخرب بواسطة خلية كاسحة scavenging cells [3] [5] [6]. يوضح الشكل (2-9) الاستجابة المناعية.

بعد أن يتم تدمير المستضد تتفعل خلايا T الكابتة و تقوم بإفراز لمفوكينات Lymphokines تؤدي إلى تثبيط عمل الخلايا المناعية و بالتالي تعمل خلايا T الكابتة على تعديل عمل النظام المناعي مانعة الاستجابة المفرطة [6] .

تتحول بعض خلايا T و B إلى خلايا ذاكرة تحفظ بنية المستقبلات التي تم التوصل لها بنتيجة التفاعل المناعي و ذلك من اجل تسريع الاستجابة في التعرض التالي لنفس المستضد حيث تكون هذه الخلايا موجودة في الدم و تميز المستضد مباشرة في حال دخوله مرة أخرى للجسم [3] [5].

### 2-4-3- نضوج الخلايا T : T Cell Maturation

يتم إنتاج الخلايا T في نقي العظم حيث تكون خاملة ولا تقوم بأي دور في النظام المناعي ثم تهاجر إلى التيموس حيث تخضع إلى عملية إنضاج لتصبح بعدها خلايا T الناضجة Mature T Cell ثم يتم إطلاقها لتدور في الدم و تمارس عملها [4].

بعد عملية النضوج تحصل كل خلية من خلايا T على مستقبلات فريدة على سطحها تدعى مستقبلات الخلايا T (TCRs) T cell receptors ، هذه المستقبلات قادرة على تمييز المركب MHC- بيبتيد ، و الغاية من عملية الإنضاج هو الحصول على خلايا قادرة على تمييز الأجسام الغريبة عن الجسم و بنفس الوقت لا تميز خلايا الجسم أي القدرة على التمييز بين الذاتي و غير الذاتي ، تدعى الخلايا الناتجة خلايا متحملة للذات self-tolerance .

يتم تشكيل المستقبلات بإعادة تركيب قطع الجينات المختلفة الموجودة في الخلية و ذلك بشكل عشوائي ثم تخضع الخلية الناتجة لعملية انتقاء [3] [4] [5].

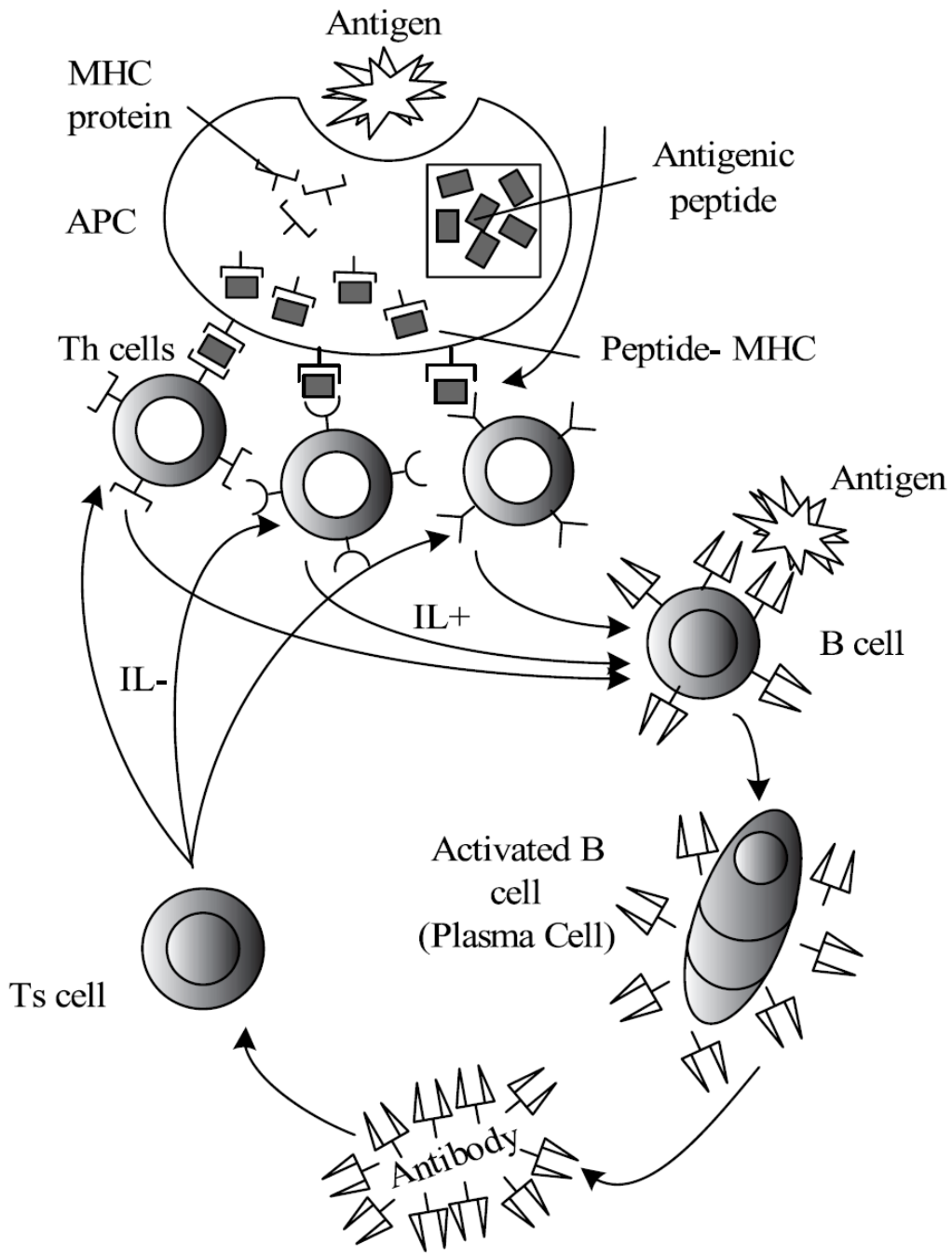

Immune response process.

الشكل (2-9) الاستجابة المناعية [6]

**مبدأ الانتقاء السلبي** : و هو يصف الآلية التي يتم إتباعها في التيموس لانتقاء خلايا T الناضجة و السماح لها بدخول الدم ، يتم اختبار خلايا T الناتجة في التيموس و عندما تقوم أي خلية بتمييز مركب ببتيد ذاتي-MHC يتم رفضها و التخلص منها ، أما التي لا تميز هذا المركب يحتفظ بها لذلك تدعى العملية بالانتقاء السلبي [3] [4] [5].

يوضح الشكل (2-10) عملية الانتقاء السلبي و كيف يتم التخلص من الخلايا التي تميز الذات.

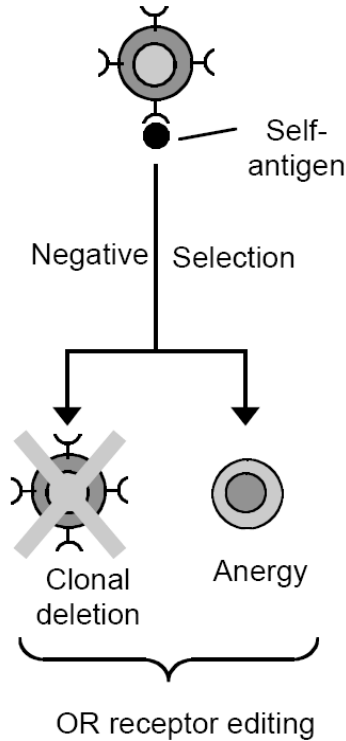

الشكل (2-10) الانتقاء السلبي [4]

### 2-4-4- مبدأ الانتقاء النسيلي : Clonal Selection Principle

الانتقاء النسيلي هو العملية التي تصف تكاثر الخلايا B أثناء الاستجابة المناعية ، و كما ذكرنا سابقاً فإنه عند تحفيز الخلايا B أثناء الاستجابة المناعية تبدأ بتغير بنية المستقبلات على سطحها محاولة تمييز بنية المستضد ، و حسب مبدأ الانتقاء النسيلي فان الخلايا التي تتعرف على المستضد بشكل اكبر تتكاثر أكثر من الخلايا التي تتعرف على المستضد بشكل اقل.

بالتالي مقدار تكاثر الخلايا الناتجة يعتمد على مقدار التآلف بين المستقبلات الموجودة على سطحها و المستضد و كلما زاد التآلف ازداد تكاثر الخلايا، تتم هذه العملية في العقد اللمفاوية

في المراكز المنتشة germinal centers (GCs) الغنية بالخلايا المقدم للمستضد [3] [4] [5]. يوضح الشكل (2-11) فكرة الانتقاء النسيلي.

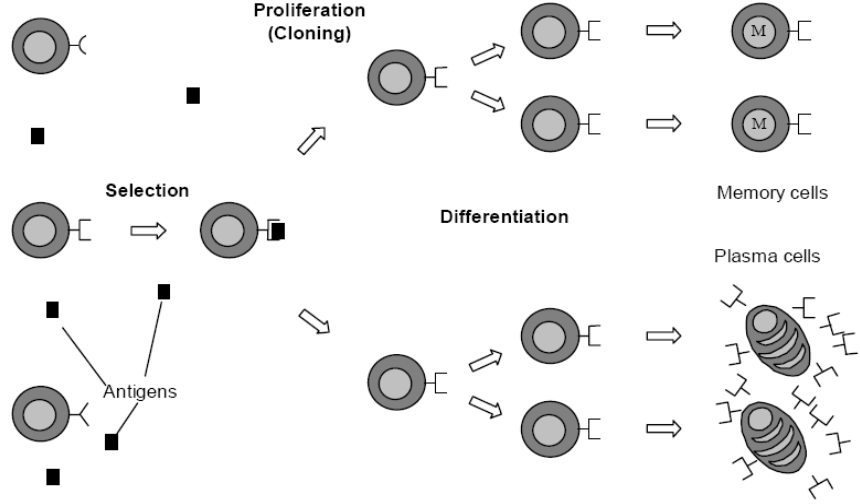

الشكل (2-11) الانتقاء النسيلي [3] [5]

العمليات الأساسية في الانتقاء النسيلي هي [3] [5]:
1. تشكيل خلايا جديدة هي نسخة من الخلية الأم (نسائل من الأم).
2. العمل على تغيير بنية المستقبلات في الخلايا الأبناء.
3. عدد الخلايا المستنسلة عن كل خلية يعتمد على مقدار التآلف الناتج بين المستقبلات و المستضد.
4. التخلص من الخلايا التي تحوي مستقبلات تتفاعل مع الجسم .

النتيجة النهائية لعملية الانتقاء النسيلي هي خلايا B ناضجة أو ما يدعي خلايا بلازمية إضافة إلى خلايا B تشكل ذاكرة مناعية.

### 2-4-5- تشكيل الأضداد و نضوج التآلف : affinity maturation

نضوج التآلف يصف العملية التي تؤدي إلى زيادة مقدار التآلف بين المستقبلات على سطح الخلايا اللمفاوية و المستضد و هذه العملية متعلقة بآليات توليد هذه المستقبلات. لا فرق بين الضد و المستقبل على سطح الخلية B إذ لهما نفس البنية لكن الضد منفصل عن الخلية و لذلك فان عملية توليد المستقبل تكافئ فعلياً عملية توليد الضد. تتم عملية نضوج التآلف في المراكز

المنتشة (GCs) germinal centers و ذلك لتحقيق تغيير في بنية الضد من اجل محاولة البحث عن تمييز اكبر لبنية المستضد و من ثم انتقاء الخلايا ذات المستقبلات عالية التآلف [3] [4] [5].

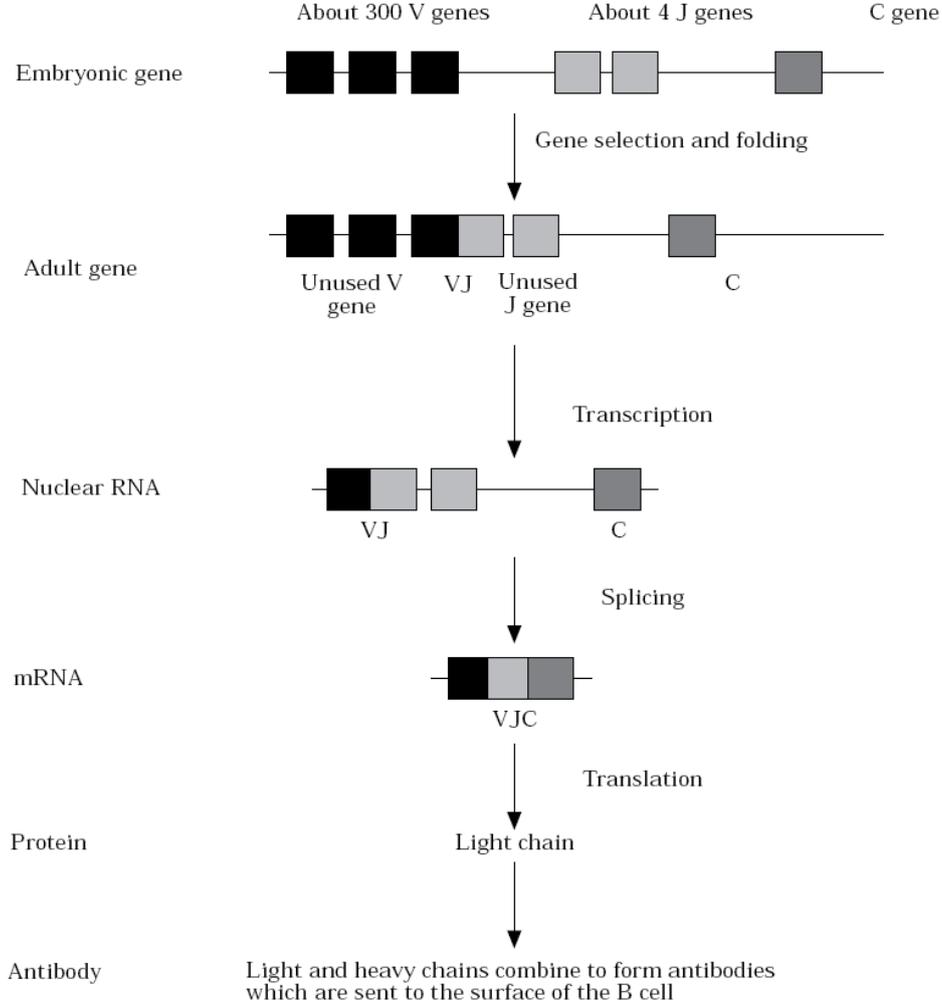

الشكل (2-12)توليد الضد [7]

إن سلاسل متعدد الببتيد الخاصة بالضد مرمزة بقطع جينات متعددة مبعثرة داخل الكروموسوم و يتم توليد الأضداد و إنضاج التآلف (تحسينه) وفق العمليتين التاليتين: الطفرة المعززة جسدياً somatic hypermutation و عملية تحرير المستقبلات أو إعادة التجميع recombination . إن هذه العملية (توليد الأضداد) ذات طبيعة عشوائية و ذلك بغرض تحقيق تنوع كبير من الأضداد إذ يتم أولا تجميع قطع الجينات لتوليد عدد كبير من الخلايا

المختلفة و عندها عدة خلايا فقط ستتعرف على المستضد ثم نلجأ بعدها إلى الطفرة لتحسين تمييز هذه الخلايا المنتقاة (عملية ضبط دقيق) [7].

يوضح الشكل (2-12) عملية تجميع قطع الجينات من اجل توليد بنى مختلفة للضد. كما يوضح الشكل (2-13) تأثير كل من الطفرة و تحرير المستقبلات على مقدار التآلف للضد آخذين بعين الاعتبار البنى المختلفة لمواقع الارتباط على المستضد (عملياً ما نريد تمييزه في المستضد).

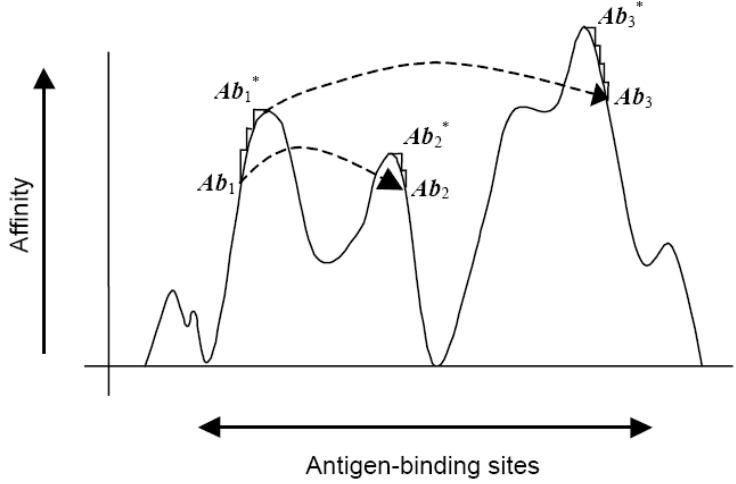

الشكل (2-13) تأثير الطفرة و تحرير المستقبلات على التآلف [3] [5]

تؤدي عملية الطفرة إلى تغير صغير و تعطي انتقال صغير وصولاً إلى حل امثل محلي ، أما عملية تحرير المستقبلات تنقل الأضداد بشكل قفزات كبيرة الشيء الذي يسمح بالانتقال إلى حل محلي آخر. من العمليات الأخرى التي تساهم في تنوع الخلايا هي توليد خلايا جديد ضمن نقي العظام حيث يتم استبدال نسبة 5-8% من الخلايا ذات التآلف الأدنى بخلايا جديدة قادمة من نقي العظم [3] [5].

### 2-4-6- الذاكرة المناعية:

بعد عملية الاستجابة المناعية و تميز النظام المناعي للمستضد و التخلص منه يحتفظ النظام المناعي بعدد من الخلايا ذات التآلف العالي مع المستضد لتشكل الذاكرة المناعية المكتسبة الخاصة بذلك المستضد ، و هذه العملية مهمة جداً ، إذ عند تعرض الجسم لنفس المستضد مرة ثانية تتفاعل خلايا الذاكرة المناعية معه مباشرة و تتعرف عليه بسرعة لأنها تخزن بنيته التي تم التعرف عليها عند التعرض الأول للمستضد [3].

عند التعرض الثاني للمستضد يعمل النظام المناعي على تحسين مقدار التآلف لخلايا الذاكرة مرة أخرى و بالتالي مع التعرض المستمر لنفس المستضد تتحسن قدرة النظام المناعي على تمييز ذلك المستضد مما يسرع استجابة النظام المناعي عند التعرض للمستضد مرة تلو المرة. عملياً يدل تركيز الأضداد في الدم على مقدار تعلم بنية المستضد حيث كلما زاد تمييز المستضد زاد إفراز الأضداد و بالتالي زاد تركيز الأضداد في الدم [3] [5].

يوضح الشكل (2-14) تغير تركيز الأضداد عند التعرض المتكرر للمستضد و كما نرى من الشكل فإنه عند دخول مستضد $Ag_1$ تتم الاستجابة بعد زمن تأخير بسيط و يتم زيادة نسبة الأضداد الخاصة بالمستضد $Ag_1$ ثم بعد القضاء على المستضد $Ag_1$ تتناقص نسبة الأضداد الخاصة به ، عند دخول المستضد $Ag_1$ مرة ثانية نلاحظ الاستجابة السريعة معه و ذلك بسبب الذاكرة المناعية التي تكونت من التعرض الأول. عند دخول مستضد آخر $Ag_2$ إضافة إلى المستضد $Ag_1$ عندها تتكيف خلايا B اللمفاوية إلى نوع خاص يميز المستضد $Ag_1$ و المستضد $Ag_2$ معا بآن واحد ، و هذا يدعى التفاعل المناعي المتصالب cross-reactive response [3] [5].

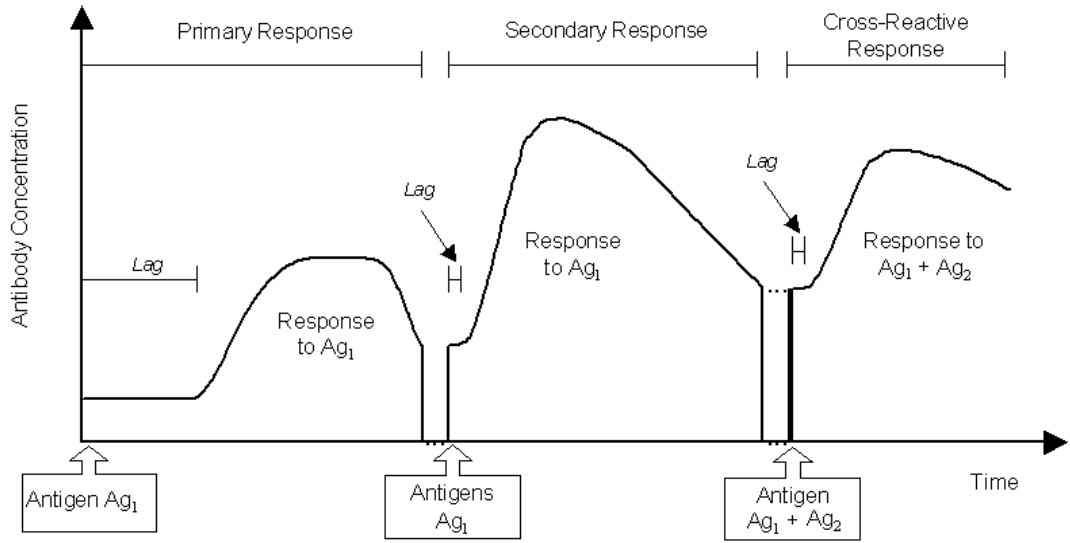

الشكل (2-14) تغير تركيز الأضداد عند التعرض للمستضدات [3] [5]

## 2-4-7- نظرية الشبكة المناعية : Immune Network Theory

تقدم هذه النظرية طريقة جديدة و متطورة لوصف نشاط الخلايا اللمفاوية و النظام المناعي , إذ تَعتبر أن النظام المناعي هو عبارة عن شبكة منظمة من الخلايا و الجزيئات التي تميز بعضها بعضاً حتى في غياب المستضد و بالتالي تُعتَبر هذه النظرية متعارضة مع نظرية الانتقاء النسيلي السابقة التي تفترض أن النظام المناعي ما هو إلا مجموعات منفصلة من النسائل التي تكون بوضع الراحة و لا تستجيب إلا عندما يتم تفعيلها من قبل مستضد [3] [4] [5].

إن الطبيعة العشوائية لعملية إنتاج الأضداد المبينة في الفقرة (2-4-5) هو ما يقودنا إلى فكرة أن الأضداد المطورة يجب أن ترى كعنصر غريب عن الجسم و سيتم التعامل معها على أنها مستضدات (البنية الناتجة عشوائياً غالباً ما تكون غريبة عن الجسم) [3] .

إن الحاتمة epitope و المستوقع paratope هما العنصران الأساسيان في عملية التمييز , و قد وجد بالتجربة أن جزيء الضد يقدم حواتم epitopes أيضا و سميت الحاتمة الخاصة بالضد بالمكنان idiotope و مجموعة الأنماط الذاتية المقدمة من المنطقة المتغيرة من الضد هي idiotype و تحدد بواسطة سلاسل الببيتيد في المنطقة المتغيرة من الضد و التي هي نفسها تحدد المستوقع paratope [3] [4]. و هذا موضح في الشكل (2-15).

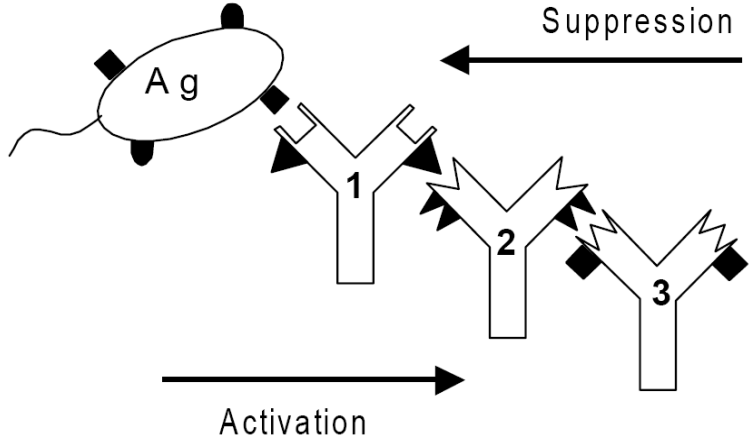

Idiotypic network representations

الشكل (2-15) تشكل الشبكة المناعية [3]

عند دخول مستضد للجسم فإن الحواتم epitopes للمستضد تميز من قبل مجموعة من المستوقعات paratopes ندعوها P1 و يوجد مع هذه المستوقعات مجموعة من المكانين idiotype المتلازمة معها ندعوها i1 عندها P1i1 هي مجموعة الأضداد التي استجابت مع المستضد. إن المجموعة P1 قادرة على تمييز مجموعة مكانين idiotype أخرى إضافة للمستضد تدعى هذه المجموعة i2 و تسمى الصورة الداخلية للمستضد و هذه المجموعة بدورها متلازمة مع مجموعة من المستوقعات P2 paratopes و تشكل معاً مجموعة من الأضداد P2i2 [3] [4].

أيضاً المجموعة i1 يتم تمييزها من قبل مجموعة من المستوقعات P3 paratopes المتلازمة مع i3 و بالتالي ستتشكل مجموعة كبيرة من المميَز و المميِز داخل الشبكة المناعية و يوضح الشكل (2-16) ذلك. هناك مجموعة على التفرع مع P1i1 تدعى Pxi1 لها نفس المكانين idiotype لكنها لا تميز نفس المستضد أي مختلفة في المستوقعات paratopes [3] [4].

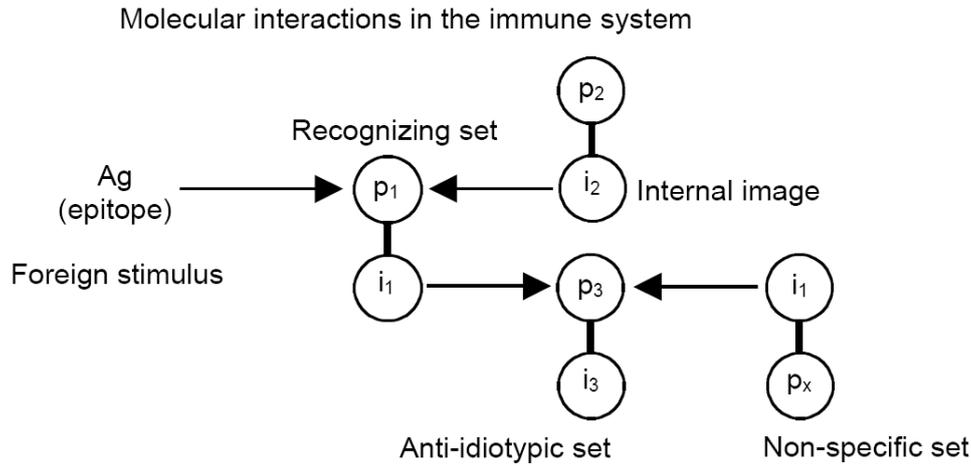

الشكل (2-16) تفاعل الجزيئات ضمن الشبكة المناعية [3]

## 2-5- خاتمة:

قدم هذا الفصل فكرة عامة عن النظام المناعي و العناصر التي يتكون منها ، حيث من أهم مكوناته: الخلايا اللمفاوية و جزيئات الضد. ثم تم شرح آليات عمل النظام المناعي و النظريات المهمة التي تشرح الاستجابة المناعية و التفاعلات بين عناصر النظام المناعي حيث من أهم هذه النظريات: الانتقاء النسيلي و الانتقاء السلبي و الشبكة المناعية. تشكل المعلومات الموجودة في هـــذا الفصـــل أساســاً مهمــاً للمعلومــات ضـــمن الفصـــول القادمـــة.

# 3

# الفصل الثالث: النظام المناعي الاصطناعي

# 3- النظام المناعي الاصطناعي:
# ARTIFICIAL IMMUNE SYSTEMS

## 1-3- مقدمة:

في السنوات الأخيرة ازداد الاهتمام بدراسة الأنظمة المستوحاة من الأنظمة الحيوية ومن بين هذه الأنظمة الشبكات العصبونية الاصطناعية (ANN) Artificial Neural Networks ، و الحوسبة التطورية (EC) Evolutionary Computation ، و النظام المناعي الاصطناعي Artificial Immune System (AIS). بناءً على مبادئ علم المناعة تم تطوير تقنية جديدة للحوسبة computational techniques و هي لا تهدف فقط إلى فهم أفضل للنظام المناعي بل تهدف أيضاً إلى حل المسائل الهندسية[8] .

و لقد ازداد الاهتمام بدراسة النظام المناعي في السنوات الأخيرة من قبل علماء الحاسب و المهندسين والرياضيين و جميعهم اهتموا بمقدرة هذا النظام ، حيث أن تعقيده يقارن بتعقيد الدماغ و بذلك ظهر حقل جديد للبحث هو النظام المناعي الاصطناعي Artificial Immune System (AIS) الذي استفاد من جوانب علم المناعة لإنشاء أدوات لحل عدة مشاكل هندسية مثل تعلم الآلة machine-learning و غيرها من المسائل.

يعتبر AIS حالياً احد فروع الذكاء الحوسبي computational intelligence [1]. و مع ذلك ليس هناك إطار عام رسمي لهذا النظام حتى الآن. يبين الشكل (3-1) علاقة AIS بالذكاء الحوسبي.

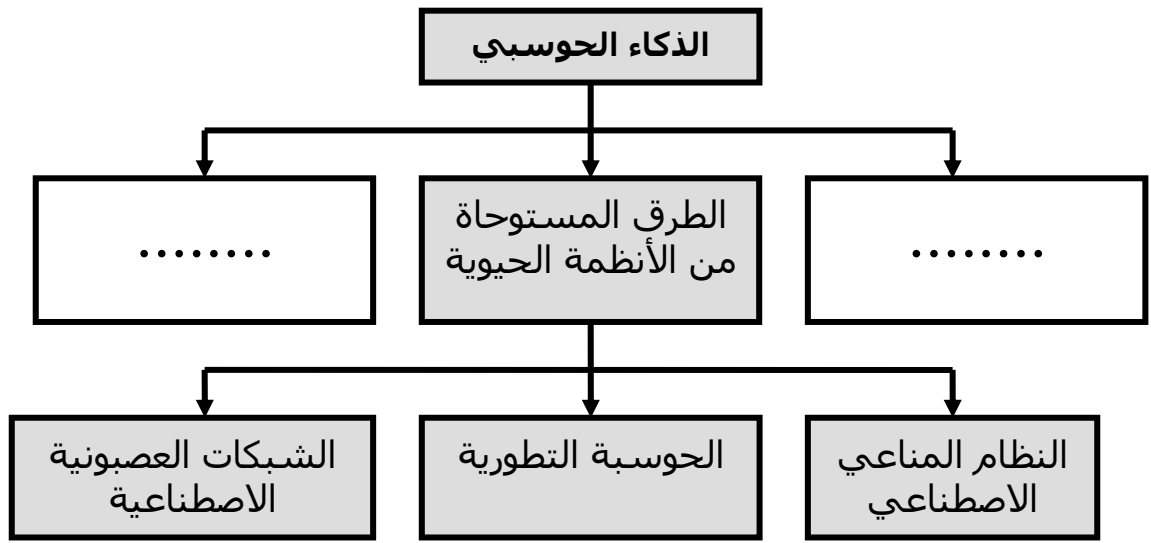

الشكل (3-1) علاقة AIS بالذكاء الحوسبي

يحوي هذا الفصل فكرة عن ماهية AIS إضافة إلى بعض الأفكار الأساسية و المهمة لبناء نظام مناعي اصطناعي AIS و هو يشرح أسباب اهتمام الباحثين و المهندسين بالنظام المناعي بغرض تصميم خوارزميات حاسوبية و يضع تعريف للنظام المناعي الاصطناعي و من ثم يذكر مجالات استخدام AIS كما يتضمن شرح لهيكلية تصميم و هندسة AIS و الخطوات العملية لتصميم نظام مناعي اصطناعي AIS لحل مسألة هندسية ما.

## 3-2- لماذا النظام المناعي:

يتمتع النظام المناعي بالعديد من الخواص التي لفتت اهتمام علماء و مهندسي الحواسيب [3] [8] مثل:

1. تمييز الأنماط : إن جزيئات و خلايا النظام المناعي قادرة على تميز الأنماط بعدة طرق.
2. التفرد Uniqueness: حيث أن كل فرد (شخص) يملك نظام مناعي خاص به و مختلف من حيث الحساسية و المقدرة.
3. تمييز الغريب: حيث أن أي جزيء مؤذي و غريب عن الجسم يتم تمييزه و التخلص منه بواسطة النظام المناعي.
4. التنوع Diversity: يوجد عدة أنواع من العناصر في النظام المناعي مثل (الخلايا، الجزيئات ، البروتينات ، ... ) و جميعها معا مسؤولة عن حماية الجسم.
5. تعدد الطبقات Multilayered: يتكون النظام المناعي عملياً من عدة طبقات من الآليات المختلفة التي تتعاون معاً وتتكامل للحصول على الحماية الأكبر.
6. كشف غير المعتاد (الشذوذ) Anomaly detection: يستطيع النظام المناعي اكتشاف و التعامل مع العوامل الممرضة التي لم يواجهها من قبل أبداً.
7. الاستقلالية Autonomy: لأن خلايا النظام المناعي موزعة في أنحاء الجسم بالكامل وهي غير تابعة لأي عنصر تحكم مركزي.
8. الاكتشاف غير الكامل و تحمل الضجيج noise tolerance: لا حاجة للتمييز الكامل للعامل الممرض و بالتالي النظام يتمتع بالمرونة.
9. التذكر و التعلم المعزز reinforcement learning and memory:  يمكن للنظام المناعي تعلم بنية العامل الممرض بحيث يستجيب لنفس العامل الممرض بشكل أسرع و أقوى في المستقبل.

## 3-3- ما هو النظام المناعي الاصطناعي:

يعتبر النظام المناعي الاصطناعي طريقة جديدة للذكاء الحوسبي computational intelligence و يمكن وضع تعريف للنظام المناعي الاصطناعي بأنه نظام متكيف مستوحى من المناعة النظرية و من النماذج و المبادئ و الآليات المناعية و يستخدم لحل مشاكل العالم الحقيقي [3] [4].

يجب التمييز بين AIS و علم المناعة النظري الرياضي Mathematical theoretical immunology حيث أن الأخير يعمل على وضع نماذج رياضية للنظام المناعي من اجل دراسة النظريات المناعية و الفهم الأعمق للنظام المناعي و محاولة تطوير تقنيات جديدة للتعامل معه مثل تطوير اللقاح و تصبح الغاية هنا هي فقط محاكاة النظام المناعي الطبيعي بغرض الدراسة و التنبؤ بتصرف النظام المناعي في الواقع. بالتالي هناك تشابه بين AIS و علم المناعة النظري الرياضي و الفرق هو في أن AIS يستخدم لحل المشاكل الهندسية كما انه يستخدم بعض الأفكار المأخوذة من النظام المناعي و هو ليس نمذجة كاملة للنظام المناعي و ذلك على عكس المناعة النظرية الرياضية التي تسعى إلى نمذجة كاملة للنظام المناعي [3].

## 3-4- مجالات استخدام النظام المناعي الاصطناعي:

تم استخدام AIS في العديد من المجالات و التطبيقات حيث يبين الشكل (3-2) معظم مجالات AIS كنسبة لعدد الأبحاث المنجزة في كل مجال و نلاحظ أن معظم مجالات استخدام AIS هي في تمييز الأنماط و الأمثلة و كشف الشذوذ [9].

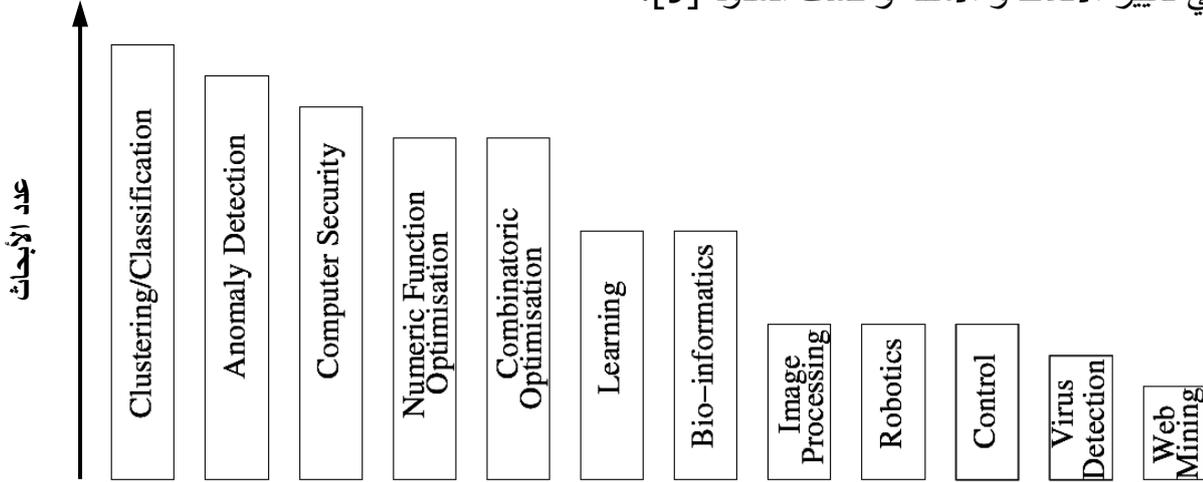

الشكل (3-2) مجالات AIS كنسبة لعدد الأبحاث المنجزة في كل مجال [9]

إن معظم أبحاث AIS تتعلق بمجال تمييز الأنماط [8] و سبب ذلك يعود إلى ارتباط النظام المناعي الوثيق بعملية تمييز الأنماط [10] و في السنوات القليلة السابقة تم استخدام AIS بشكل واسع في العديد من المجالات منها:

1. التحكم [11] [12] [13].
2. معالجة الإشارة [14].
3. تشخيص الأمراض [15].
4. تصميم الدارات الرقمية [16].
5. تطوير العديد من الخوارزميات في مجال التصنيف (تمييز الأنماط) [17] [18] [19].

## 3-5- هيكلية تصميم و هندسة AIS :

من اجل تصميم خوارزمية مستوحاة من النظم الحيوية مثل الشبكات العصبونية أو الخوارزميات التطورية فإننا نحتاج على الأقل لإتّباع الخطوات التالية[3]:

1. تمثيل عناصر النظام.
2. وضع مجموعة من الآليات لتقييم التفاعل بين عناصر النظام.
3. وضع اجرائيات للتكيف اعتمادا على ديناميكية النظام الحيوي.

على سبيل المثال من اجل تصميم نظام للشبكات العصبونية بشكل اصطناعي و وفقا لهذه الخطوات فإننا نحتاج إلى وضع تمثيل للعصبون ثم تحديد شبكة العصبونات و طريقة اتصالها مع بعضها البعض و من ثم تحديد خوارزمية التعلم التي تغير من معاملات النظام.

بالنسبة للنظام المناعي الاصطناعي فان التصميم يبدأ بالتمثيل حيث يتم صنع نموذج مجرد للأعضاء المناعية و الخلايا و الجزيئات و من ثم وضع مجموعة من التوابع التي تحدد التآلف affinity لوصف تفاعل العناصر بشكل كمي و من ثم وضع عدد من الخوارزميات ذات الأغراض العامة بحيث تعطي ديناميكية النظام المناعي [3].

يمكن تمثيل هيكلية التصميم لـ AIS بالطبقات الموضحة بالشكل (3-3) :

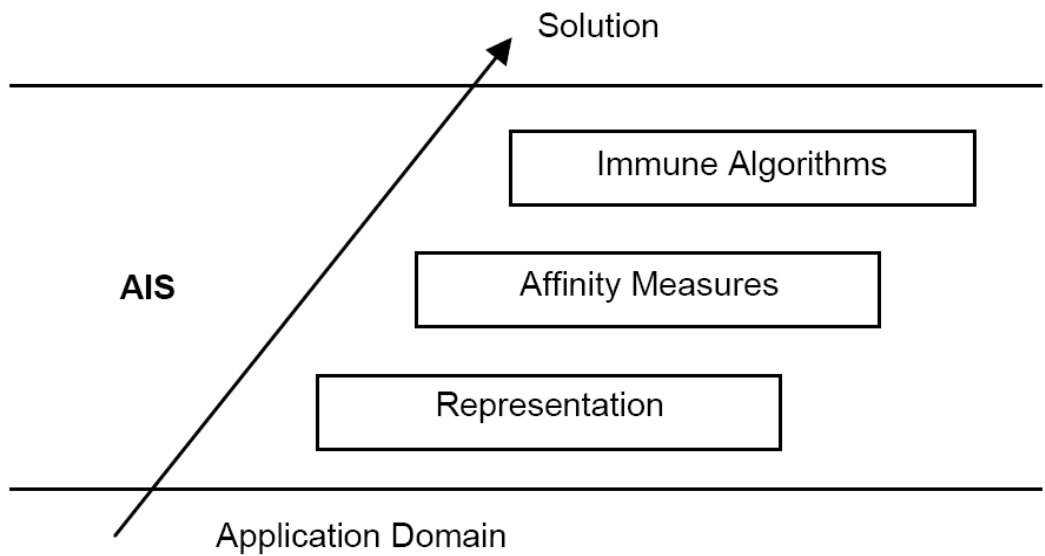

الشكل (3-3) هيكلية التصميم لـ AIS [3] [5]

للانتقال من مجال التطبيق إلى مجال الخوارزمية الحاسوبية يجب تمثيل معطيات المسألة بالشكل الأمثل لكي نستطيع التعامل معها عبر الخوارزميات المناعية و من ثم نحتاج إلى آلية لقياس التآلف affinity لتحديد مدى التتام بين الضد و المستضد ثم يتم تطبيق الخوارزمية المناعية من اجل حل المسألة [3].

### 3-5-1- التمثيل: Representation

الغاية من التمثيل هو وضع نماذج رياضية مجردة للخلايا و الجزيئات المناعية بحيث تمثل تلك النماذج معطيات و حلول المسألة الهندسية و هذا التمثيل يختلف من مسألة إلى أخرى و ذلك حسب طبيعة المسألة. في المناعة النظرية يتم تمثيل النظام المناعي و وصف التفاعل بين المستضد و جزيئات الخلايا المناعية بواسطة فكرة فضاء الشكل Shape-Space Model [3] [4] [5]. حيث يتم وصف منطقة ارتباط الضد بالمستضد (منطقة التتام) بعدة خصائص مثل الطول ، العرض ، الشحنة الكهربائية ، توزع الروابط الكهربائية (العوامل المهمة للارتباط بالمستضد) ، و تدعى مجموعة الخواص تلك بالشكل العام للجزيء generalized shape .

إذا افترضنا انه يتم وصف الشكل العام لموقع الربط لضد(الذي يدعى مستوقع paratope ) بـ L بارامتر مثل : الطول ، العرض ، التحدب ، التقعر ، الشحنة ، عندها سيصبح الضد نقطة مرسومة في فضاء ذو L بعد و يدعى هذا الفضاء بفضاء الشكل. إذا افترضنا أن الجسم يملك ذخيرة مناعية ذات حجم N (يقصد بالذخيرة المناعية الأنواع المختلفة من الأضداد التي يمتلكها

الجسم) فإن فضاء الشكل لها يحوي N نقطة و هذه النقاط موجودة في فضاء V محدود (بسبب وجود مجال محدد للخواص). يتم وصف موقع الربط للمستضد (الذي يدعى حاتمة epitope) بنفس الطريقة حيث يؤخذ متممه الذي يجب أن يكون في نفس المنطقة V. أن التفاعل بين الضد و المستضد يقاس بتتام مناطق الربط بالتالي في فضاء الشكل يحدث هذا التتام عند تطابق النقطة المقابلة لضد و متمم المستضد. في حال عدم التتام بشكل كامل بين مواقع الربط يحدث الارتباط لكن مع تآلف صغير بالتالي كل مستوقع paratope سيميز منطقة $V_\varepsilon$ محيطة به حسب المعامل $\varepsilon$ الذي هو عتبة التآلف اللازمة للارتباط [3] [5].

يبين الشكل (4-3) العملية و كما نرى كل مستوقع يمكن أن يميز عدة حواتم و يمكن للمناطق $V_\varepsilon$ أن تتقاطع لتشكل التفاعل المتصالب المذكور سابقا.

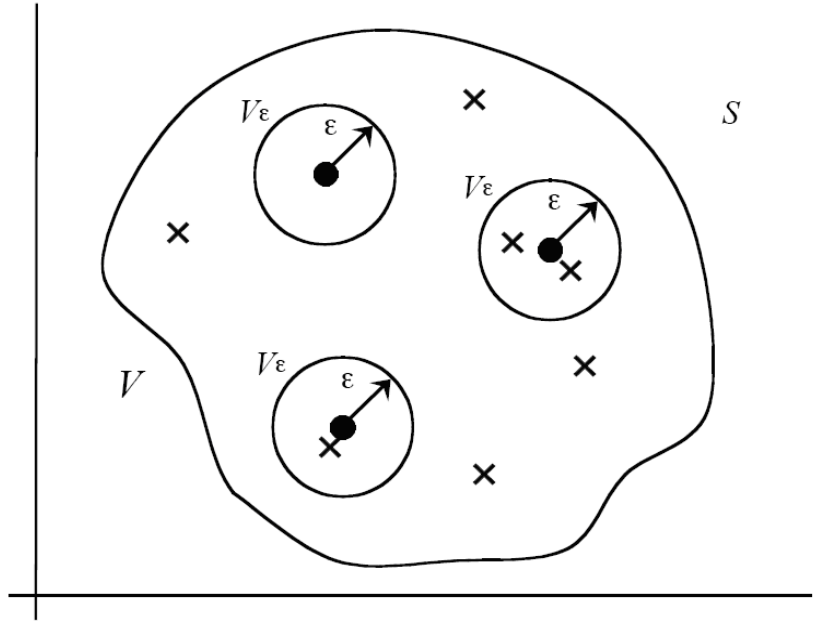

الشكل (4-3) فضاء الشكل و مناطق التمييز [3] [5]

حيث (•)ترمز للضد و(x)ترمز للمستضد

يمكن استخدام هذه الطريقة المستخدمة في المناعـة النظريـة (طريقـة فضـاء الشـكل) لتمثيـل الأضداد و المستضدات التي ستمثل بدورها معطيات المسألة و عندها يكون الشكل العام لجزيء m (سواء كان ضد أو مستضد) هو شعاع $m = [m_1, m_2, ..., m_L]$ مؤلف من مجموعة من الواصفات attribute (الإحداثيات) التي تمثل نقطة في فضاء ذو $L$ بعد ,

$m \in S^L$ حيث $S^L$ فضاء الشكل. هذه الإحداثيات ممكن أن تكون مكونة من أي نوع من الواصفات مثل قيم حقيقية أو صحيحة أو عبارة عن بتات أو حتى مجرد رموز و يتم استخراج هذه الواصفات (الخواص) من المسألة و هي عملية مهمة[3] [5].

### 3-5-2- تمثيل الرابط بين الضد و المستضد Ab-Ag و قياس التآلف:

نوعية فضاء الشكل تتبع تمثيل الخواص و آلية قياس التآلف. يمكن تمثيل الشكل العام لجزيء m (سواء كان ضد Ab أو مستضد Ag) بشعاع $m = [m_1, m_2, ..., m_L]$ مجموعة من الإحداثيات الحقيقية التي تمثل نقطة في فضاء ذو $L$ بعد , $m \in S^L \subseteq \Re^L$ حيث $S^L$ فضاء الشكل , و عندها يقاس التآلف بين الضد و المستضد بالمسافة بينهما و كلما ازدادت المسافة بينهما ازداد التآلف (إذ أننا نبحث عن التعاكس أو التتام) [3] [5].

هناك عدة توابع لقياس المسافة:

بافتراض أن

(3-1)    Ab=[ab$_1$, ab$_2$, ..., ab$_L$]

(3-2)    Ag=[ag$_1$, ag$_2$, ..., ag$_L$]

عندها نستخدم المسافة الاقليدية التي تعطى بالعلاقة:

$$D = \sqrt{\sum_{i=1}^{L}(Ab_i - Ag_i)^2} \quad (3\text{-}3)$$

و يصبح فضاء الشكل اقليدي .

كما يمكن استخدام مسافة مانهاتن التي تعطى بالعلاقة :

$$D = \sqrt{\sum_{i=1}^{L}|Ab_i - Ag_i|} \quad (3\text{-}4)$$

و يصبح فضاء شكل مانهاتن و هو مفيد عند تطبيق الخوارزمية بواسطة كيان صلب hardware.

عندما تكون Ab و Ag ممثلة بتتابع من الرموز المأخوذة من أبجدية و التي يمكن أن تترجم على أنها ببتيدات نستخدم مسافة هامينغ.

$$D = \sum_{i=1}^{L} \delta, \text{ where } \delta = \begin{cases} 1 & \text{if } ab_i \neq ag_i \\ 0 & \text{otherwise} \end{cases} \quad (3\text{-}5)$$

ويدعى عندها فضاء شكل هامينغ.

يجب تحديد العلاقة بين المسافة و منطقة التمييز و عتبة التآلف $\varepsilon$, عندما تكون المسافة أعظمية يكون التتام أعظمي و التآلف أعظمي. قيمة الربط هو مقدار يحدد فيما إذا كان هناك ارتباط بين الجزيئات أو لا و للحصول على قيمة الربط بناءً على المسافة بين الجزيئين يمكن استخدام عدة توابع مثل تابع قفزة أو تابع بشكل حرف S في الحالات المستمرة , يتحقق الربط من اجل مسافة اكبر من $\varepsilon$.

يوضح الشكل (3-5) التوابع المستخدمة لحساب قيمة الربط.

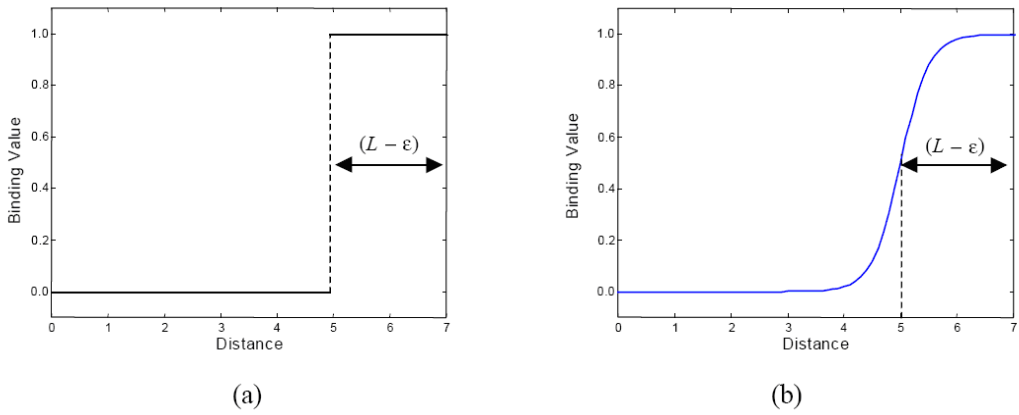

الشكل (3-5) توابع قيمة الربط [3] [5]

إن معيار قياس التآلف يختلف حسب طبيعة المسألة و طريقة تمثيل الحلول للمسألة و ليس من الضروري استخدام المعايير السابقة بل يمكن استخدام معايير للتآلف مختلفة تماماً بحيث تلائم طبيعة المسألة المراد حلها ، إن المعايير السابقة غالباً ما تستخدم في مسائل تمييز الأنماط.

### 3-5-3- الخوارزميات المناعية:

بعد عملية التمثيل للخلايا و الجزيئات المناعية و وضع آلية لقياس التفاعل بين تلك الجزيئات (عملياً آلية لقياس التآلف) يجب تصميم الخوارزمية المناعية التي ستعمل على إيجاد الحل للمسألة. يتم بناء الخوارزمية بأخذ بعض الأفكار و الآليات التي يتبعها النظام المناعي البيولوجي أثناء عمله و تحويلها إلى خوارزميات حاسوبية [3].

أهم تلك الآليات التي تم تحويلها إلى خوارزميات حاسوبية هي [3]:

- آلية عمل التيموس التي أعطت خوارزمية الانتقاء السلبي.
- مبدأ الانتقاء النسيلي الذي أعطى خوارزمية الانتقاء النسيلي.
- نظرية الشبكة المناعية التي أعطت خوارزمية الشبكة المناعية الاصطناعية.

سيتم شرح هذه الخوارزميات بالتفصيل في الفصول القادمة إضافة إلى ذكر المبادئ المناعية الأساسية التي أخذت منها.

## 3-6- خطوات تصميم AIS:

يمكن تلخيص الخطوات اللازمة لتصميم نظام مناعي اصطناعي AIS من اجل حل مسألة ما كما يلي [3]:

1. وصف المسألة و الغاية هنا تعريف كل عناصر AIS اللازمة للمسألة و ذلك يشمل كل محددات المسألة و التوابع و المتحولات بغرض الانتقال من مجال المسألة إلى مجال AIS.
2. تحديد العناصر المناعية التي ستُستخدم في النظام المناعي الاصطناعي AIS مثل الضد أو الخلية B اللمفاوية أو خلية T اللمفاوية أو التيموس.
3. وضع تمثيل رياضي لعناصر النظام المناعي الاصطناعي AIS المستخدمة.
4. تحديد المبادئ المناعية التي ستُستخدم لحل المسألة ثم تحويلها إلى خوارزميات بغرض حل المسألة بواسطتها.
5. بعد الحصول على الحل فك ترميز الحلول الناتجة و نقلها من جديد إلى مجال المسألة الأصلية.

## 3-7- الخاتمة:

أعطى هذا الفصل فكرة عامة عن ماهية النظام المناعي الاصطناعي AIS و أسباب البحث في مجال النظام المناعي بغرض الوصول لخوارزميات حاسوبية ، كما تم ذكر المجالات المتعددة التي استخدم فيها AIS ، و تضمن هذا الفصل شرح للهيكلية العامة لـ AIS و خطوات تصميم مثل هذه الأنظمة لحل المسائل الهندسية. تشكل الأفكار المضمنة ضمن هذا الفصل مبادئ مهمة لتصميم الخوارزميات المناعية بشكل عام.

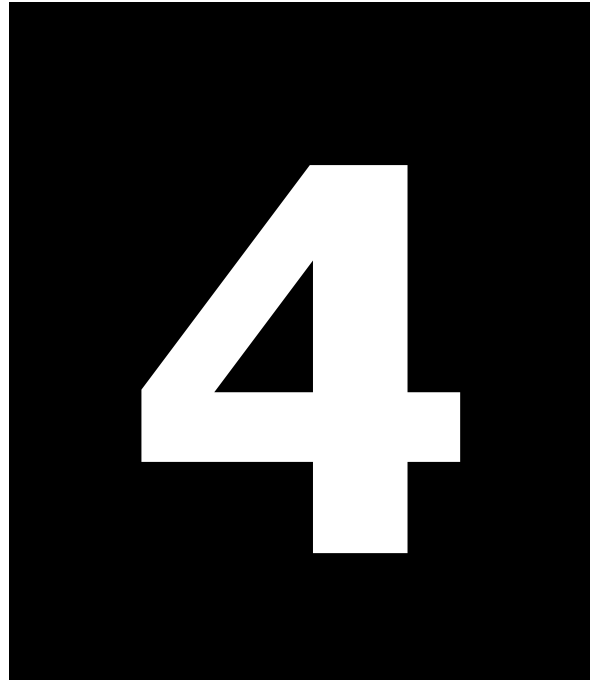

# الفصل الرابع: تمييز الأنماط

# 4- تمييز الأنماط: Pattern Recognition

## 4-1- مقدمة:

يتميز الإنسان بقدرته على التمييز (تمييز الكلام ، الأشخاص ، الأرقام ، الأشكال ، الأشياء بشكل عام) حيث تشكل هذه الأشياء أنماط ذات خواص متقاربة و هذه الخواص تميزها عن غيرها من الأنماط الأخرى. تمييز النمط هو قدرة الآلة على مراقبة البيئة و تعلم التمييز بين الأشياء موضوع الاهتمام عن باقي الأشياء في الخلفية و من ثم اتخاذ قرارات معقولة عن المجموعات التي تشكلها هذه الأنماط [20].

يعرف النمط بأنه أي غرض أو عملية أو حدث يمكن تحديده و إعطائه اسم ، و يكون عملياً تمييز الأنماط هو تصنيف معطيات الدخل (الأنماط) إلى احد أصناف الأنماط المحددة مسبقا. الصنف هو مجموعة من الأنماط التي تشترك في مواصفات عامة ، و ندعو الآلة التي تقوم بعملية التصنيف بالمصنف.

غالبا عملية التمييز لا تعتمد على الشيء بحد ذاته بل على الخواص المقاسة للشيء و لضمان نجاح عملية التصنيف يجب الحصول على معلومات كاملة ومسبقة عن أصناف الأنماط. الغاية من هذا الفصل إعطاء فكرة عامة عن تمييز الأنماط و الطرق المختلفة المستخدمة لتحقيق ذلك.

## 4-2- تصميم نظام لتمييز الأنماط:

عملية تصميم نظام لتمييز الأنماط تتضمن بشكل أساسي ثلاث جوانب [20]:

1. تحصيل المعطيات و المعالجة الأولية لها.
2. تمثيل المعطيات بالشكل المناسب.
3. تحديد آلية اتخاذ القرار.

## 4-3- طرق تمييز الأنماط:

أشهر الطرق المعروفة لتمييز الأنماط هي [20]:

1. المطابقة بالقوالب Template Matching .
2. التصنيف الإحصائي Statistical Approach .
3. المطابقة البنيوية أو النحوية Syntactic Approach .
4. الشبكات العصبونية Neural Networks [21].

### 4-3-1- المطابقة بالقوالب:

و هي من ابسط الطرق و أقدمها في تمييز الأنماط و هي عملية عامة في التمييز و تستخدم لتحديد مقدار التشابه بين شيئين من نفس النوع (نقاط ، منحنيات ، أشكال). في هذه العملية يجب توفير قالب أو نموذج أولي عن النمط المراد التعرف عليه , و عندها يتم مطابقة النمط المراد تمييزه مع القوالب المخزنة آخذين بالحسبان جميع التوضعات الممكنة (الانتقال و الدوران و تغير المقياس) [20] [22].

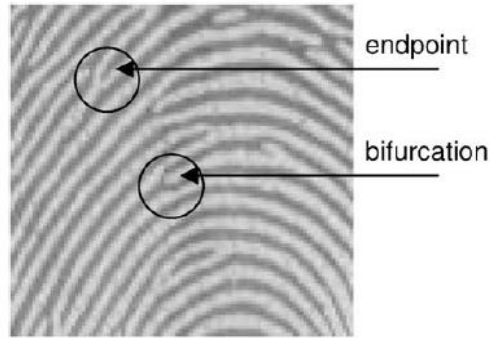

الشكل (4-1) التفرعات و النهايات في البصمة [22]

احد الأمثلة على المطابقة بالقوالب هي عملية تمييز بصمة الإصبع [22] حيث يتم تشكيل القالب من مجموعة من النقاط التي تحدد التفرعات و النهايات للخطوط المشكلة للبصمة مثل الموضحة في الشكل (4-1) و من ثم نعمل على تحديد مقدار التطابق بين بصمة الدخل و البصمة المخزنة مسبقا و ذلك عن طريق تحديد مقدار التطابق بين هذه النقاط مع القيام بعمليات التدوير و النقل و التقييس لضمان اكبر تطابق. يوضح الشكل (4-2) عملية المطابقة بين القالب و النقاط المستخرجة من البصمة. من مساوئ هذه الطريقة التشوهات اللاخطية للنمط [19] [20].

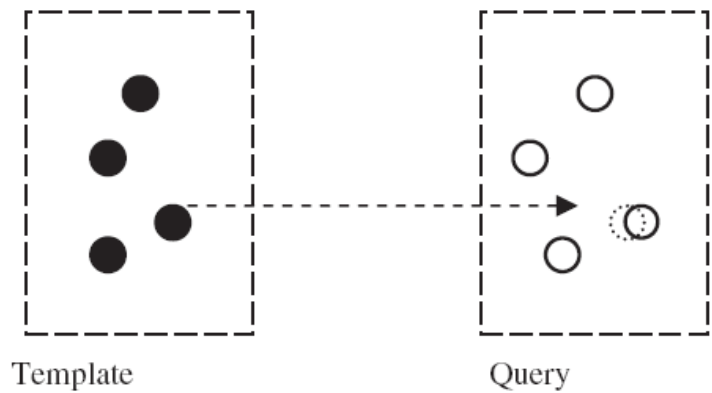

الشكل (4-2) عملية المطابقة [22]

### 4-3-2- المطابقة البنيوية:

في العديد من مسائل التمييز التي تحتوي أنماط معقدة فانه من الأفضل استخدام نموذج هرمي حيث ينظر إلى النمط على انه مؤلف من أنماط فرعية بسيطة و هذه الأنماط الفرعية أيضا مؤلفة من أنماط فرعية أخرى أبسط و هكذا ... . العناصر الأبسط تدعى التراكيب الأولية و عندها النمط يتم تمثيله بالقواعد التي تربط التراكيب الأولية أي أن عدد كبير من الأنماط المعقدة يمكن وصفها بواسطة عدد صغير من العناصر الأولية و القواعد ، عندها تصبح الغاية تحديد العناصر الأولية و استخراج القواعد بينها [20].

تتم مقارنة هذه الطريقة بالصياغة النحوية في اللغة حيث العناصر الأولية ترى على أنها أبجدية اللغة و يتم توليد الجمل بناءً على القواعد و بالتالي ترى الأنماط على أنها جمل تنتمي إلى لغة. استخدام هذه الطريقة يقود إلى صعوبات مثل اكتشاف العناصر الأولية (و يتم ذلك غالباً بالتجزئة ضمن نماذج ذات ضجيج) و من ثم استخراج القواعد من المعطيات [20].

### 4-3-3- التصنيف الإحصائي:

تعمل هذه الطريقة على تشكيل نموذج إحصائي للشيء المراد تمييزه بدلاً من تشكيل قالب له و ذلك بتحصيل مجموعة من الأنماط (النماذج) لتشكيل مجموعة معطيات إحصائية و بناءً على هذه المعطيات يتم تحديد النموذج الإحصائي بغرض بناء آلية لاتخاذ القرار (آلية التمييز) ، في هذه الطريقة يتم تمثيل النمط بواسطة شعاع مؤلف من الخواص المستخرجة من العنصر المراد تمييزه بحيث يظهر النمط كنقطة في فضاء متعدد الأبعاد و عدد الأبعاد هو عدد الخواص المشكلة للنمط [20].

تشكل مجموعة الأنماط معاً للعنصر صنف واحد لتعطي معلومات عن ذلك العنصر. من المهم اختيار أو إيجاد خواص تسمح لأشعة الأنماط بان تنتمي إلى مجموعات مختلفة و أن تحتل مناطق مضغوطة و غير متداخلة في فضاء الخواص [20]. يبين الشكل (4-3) مثالاً عن مجموعات الخواص الجيدة الفصل و الخواص السيئة الفصل. بعد عملية التمثيل يتم تحديد حدود القرار في فضاء الخواص بحيث تفصل الأنماط التي تنتمي إلى أصناف مختلفة.

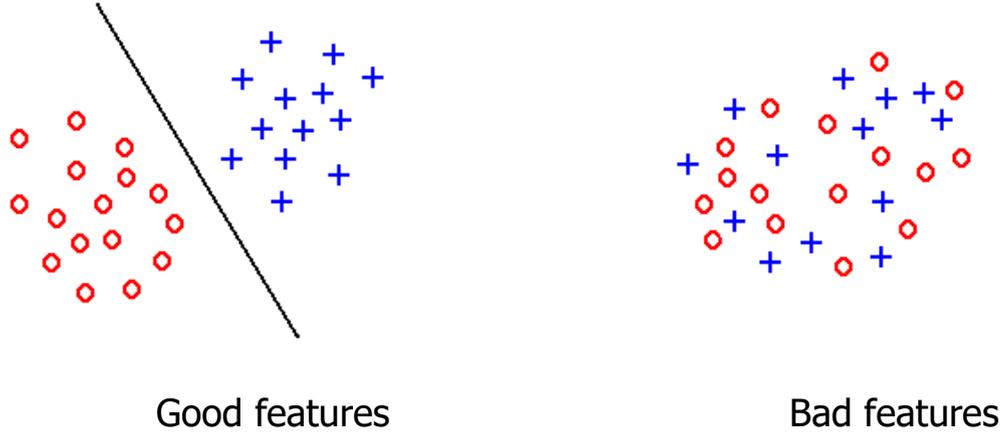

الشكل (3-4) مجموعات الخواص

في طريقة القرار الإحصائي النظري يتم تحديد حدود القرار بواسطة التوزع الاحتمالي للأنماط التي تنتمي لنفس الصنف لكن إذا كان التوزع الاحتمالي للأصناف غير معروف يتم بناء حدود القرار باستخدام خوارزميات التعلم و التي تقسم إلى [20]:

1. التعلم الموجه و تدعى عملية التمييز عندها بالتصنيف الموجه.
2. التعلم غير الموجه و تدعى عملية التمييز عندها بالتصنيف غير الموجه.

في التصنيف الموجه supervised يتم تحديد حدود القرار من مجموعة أنماط للتدريب (مجموعة معطيات معروفة الأصناف) بحيث يفترض أن معطيات التدريب هذه تعطي معلومات كافية عن التوزع المفترض لأصناف الأنماط في الفضاء [20] .

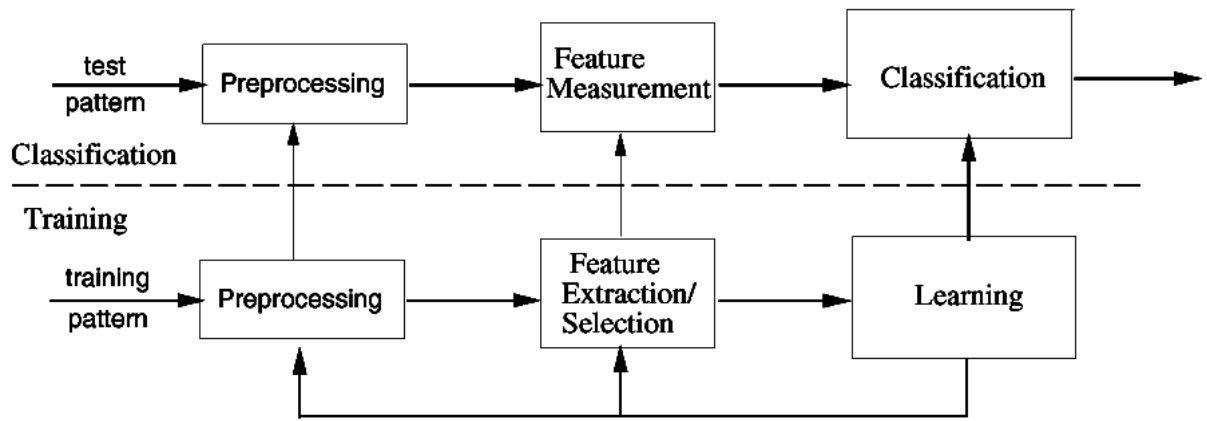

الشكل (4-4) التصنيف الموجه [20]

يوضح الشكل (4-4) عملية التصنيف الموجه الذي يتم على مرحلتين:

المرحلة الأولى مرحلة التدريب (التعلم) و المرحلة الثانية مرحلة الاختبار (التصنيف). و تحتوي كل مرحلة على عمليات المعالجة الأولية حيث يتم فصل النمط موضوع الاهتمام عن الخلفية و من ثم إزالة الضجيج و تقييس النمط ضمن مجال ملائم و القيام بأي عملية تساعد في التمثيل الأفضل للنمط. يسمح خط التغذية الخلفية للمصمم بأمثلة عمليات المعالجة الأولية و استراتيجيات استخراج و انتقاء الخواص. يعمل المصنف المدرب في مرحلة التصنيف (بعد عملية التدرب) على إسناد أنماط الدخل غير معروفة الصنف إلى احد أصناف الأنماط و ذلك بناءً على اعتبارات تعتمد على الخواص المقاسة [20].

يعمل المصنف في التصنيف غير الموجه unsupervised على المعطيات مباشرة دون معلومات مسبقة عن الأصناف أو أمثلة للتدريب (معطيات تدريب) و يقوم بتصنيف المعطيات حسب معيار للتماثل بينها [20] [23] [24].

في عملية تميز الأنماط الإحصائية هناك العديد من الاستراتيجيات المستخدمة لتصميم المصنف و هي تعتمد على مقدار المعلومات المتوفرة عن التوزع الاحتمالي الشرطي للأصناف. يوضح الشكل (4-5) الطرق المستخدمة في تمييز الأنماط الإحصائي.

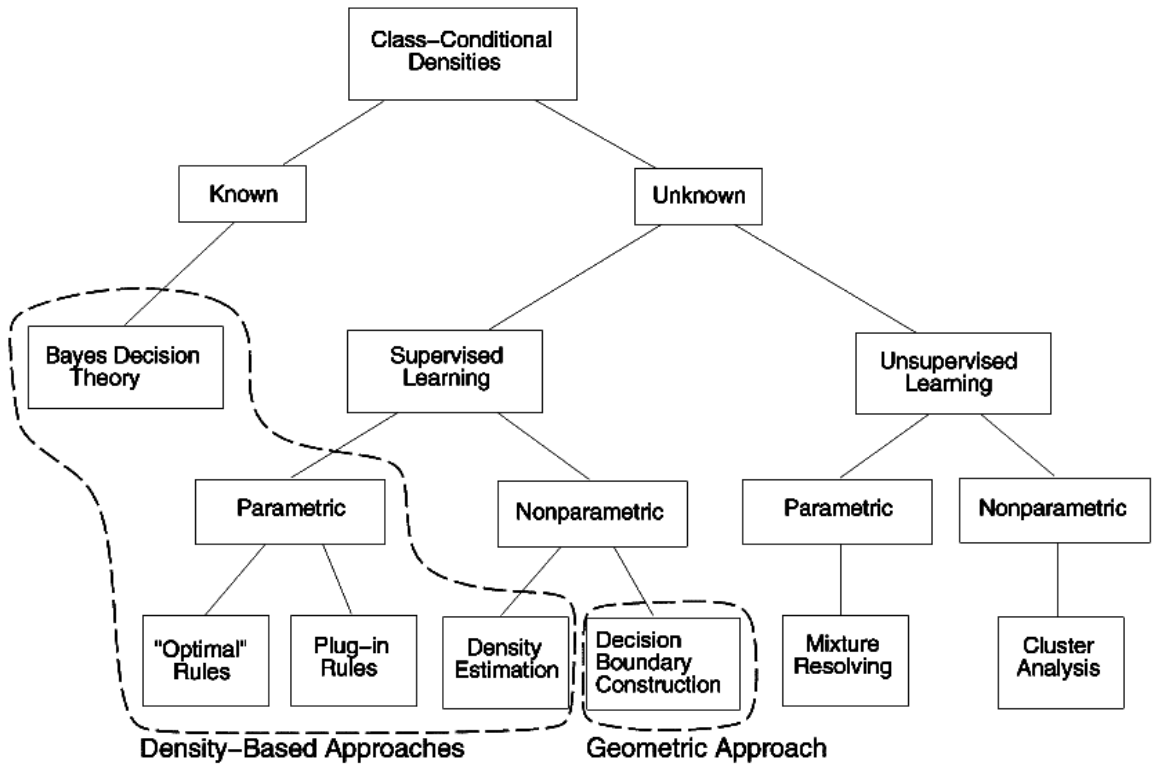

الشكل (4-5) الطرق المستخدمة لتمييز الأنماط إحصائياً [20]

نطبق قاعدة Bayes Decision لتصميم المصنف في حال التوزع الاحتمالي معروف تماماً و لكن إذا كان التوزع الاحتمالي غير معروف عندها نلجأ إلى تعلم التوزع الاحتمالي من معطيات التدريب المتوفرة. إذا كان الشكل العام للتوزع الاحتمالي معروف (مثلاً غوصي) لكن محدداته غير معروفة تصبح الغاية من عملية التعلم هي تحديد بارامترات التوزع فقط و المسألة هي قرار بارامتري parametric decision. لكن إذا كان الشكل العام للتوزع الاحتمالي غير معروف عندها نعمل في نمط غير بارامتري وعندها يجب تقدير التوزع الاحتمالي [20]. هناك العديد من الخوارزميات المعروفة جيداً و المستخدمة في التمييز الإحصائي مثل K-nearest neighbor classifiers و kernel classifiers [25].

بشكل عام الخطوات الأساسية لعملية تميز الأنماط إحصائياً هي [26]:

1. المعالجة الأولية.
2. استخراج الخواص و انتقاء الخواص.
3. تقدير التوزع الاحتمالي سواء بارامتري أو غير بارامتري (التعلم).
4. عملية اتخاذ القرار (التصنيف).
5. تقييم الأداء.

تعتبر عملية تحليل صور الأقمار الصناعية من أهم تطبيقات تمييز الأنماط إحصائياً و السبب في ذلك يعود إلى مقدار المعطيات الكبير الموجود في تلك الصور و عملياً في تمييز الأنماط الإحصائي فإننا بحاجة لعدد كبير من المعطيات لتقدير الاحتمالات بشكل دقيق [26].

## 4-4- التصنيف غير الموجه Unsupervised Classification:

غالباً في العديد من تطبيقات تمييز الأنماط يكون من الصعب الحصول على معطيات تدريب معلمة (معروف مسبقاً الأصناف التابعة لها) و مثال على ذلك عملية تحليل صور الأقمار الصناعية فللحصول على معلومات حقيقة يجب زيارة الموقع المدروس و تحديد مجموعة من نقاط علام تابعة لكل صنف بواسطة نظام معلومات جغرافي و يكون ذلك غالباً صعب جداً و يلغي الفائدة من استخدام صور الأقمار الصناعية. لذلك يلجأ إلى عملية التصنيف غير الموجه حيث يتم بناء حدود القرار بناءً على معطيات تدريب غير معلمة (غير معروف مسبقاً إلى أي صنف تتبع).

تدعى هذه العملية أيضاً بعنقدة المعطيات data clustering و هي مجموعة من العمليات و الإجرائيات الغاية منها الحصول على المجموعات الطبيعية أو العناقيد clusters ضمن المعطيات و ذلك بالاعتماد على قياس مقدار التماثل بين المعطيات ضمن العنقود [23] [24].

عملياً العنقدة أو التصنيف غير الموجه هو عملية صعبة جداً لأن المعطيات يمكن أن تشكل عناقيد أو مجموعات بأشكال مختلفة و أحجام مختلفة [20]. يوضح الشكل (4-6) ذلك.

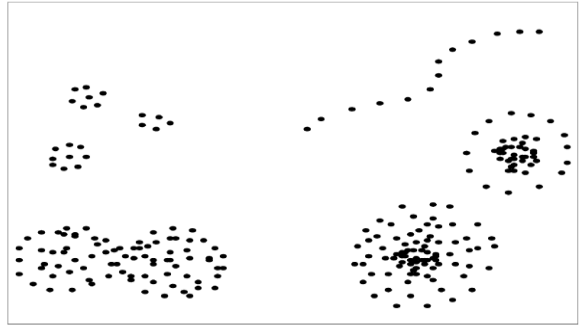

الشكل (4-6) الأشكال المختلفة لمجموعات المعطيات [20]

بناءً على ذلك تم اعتماد عدة تعاريف للعنقود [20] [23]:

1. الأنماط (النماذج) ضمن العنقود (مجموعة المعطيات المترابطة) تتشابه معاً أكثر من تشابهها مع المعطيات ضمن عنقود آخر.
2. العنقود يشكل منطقة ذات كثافة عالية من النقاط مفصولة عن العناقيد الأخرى بمناطق ذات كثافة منخفضة من النقاط.

و بالرغم من وجود هذه التعاريف فإنه من الصعب عملياً تحديد العناقيد (مجموعات المعطيات المترابطة) كما أن معيار التشابه مهم جداً لتعريف العنقود.

استُخدم التصنيف غير الموجه أو تحليل العناقيد cluster analysis كأداة فعالة و قوية في العديد من المجالات مثل : التنقيب عن المعطيات [27] و تجزئة الصورة [28] و استيفاء المعلومات [29] و ترميز و ضغط الإشارة [30] و تعلم الآلة [31]. و بناء على ذلك تم تطوير العديد من خوارزميات العنقدة clustering algorithms و تطوير الخوارزميات الجديدة ما يزال مستمراً حتى الآن [20].

إن معظم هذه الخوارزميات يعتمد على تقنيتين أساسيتين للعنقدة [20] [23] [24]:

1. تقنية العنقدة التقسيمية Partitional clustering.
2. تقنية العنقدة الهرمية Hierarchical clustering.

غالباً ما يتم استخدام تقنية العنقدة التقسيمية أكثر من تقنية العنقدة الهرمية.

إن الدراسات الحديثة حول التصنيف غير الموجه تقترح على مستخدم خوارزميات العنقدة مراعاة ما يلي [20]:

1. ستعمل خوارزمية العنقدة على إيجاد العناقيد ضمن مجموعة المعطيات سواء وجدت فعلياً هده العناقيد أم لم توجد . لذلك يجب دراسة إمكانية العنقدة قبل تطبيق الخوارزمية و من ثم دراسة صلاحية العناقيد الناتجة عن الخوارزمية.
2. لا وجود لخوارزمية العنقدة المثلى لذلك يجب تجريب عدة خوارزميات من اجل مجموعة المعطيات المدروسة.
3. عملية جمع المعطيات و تمثيلها و تقيسها و صلاحية العناقيد هي مواضيع مهمة مثل أهمية اختيار الخوارزمية (اختيار إستراتيجية العنقدة).

### 4-4-1- العنقدة التقسيمية Partitional clustering:

عملية العنقدة التقسيمية هي كالتالي: لدينا $n$ نمط في فضاء ذو $L$ بعد و الغاية هي تحديد التقسيم بين الأنماط إلى $K$ مجموعة (عنقود) cluster بحيث تكون الأنماط ضمن العنقود الواحد متشابهة مع بعضها البعض أكثر من تشابهها مع الأنماط في العناقيد الأخرى [23] [24].

عدد العناقيد $K$ (المجموعات) قد يكون معروف أو غير معروف. للقيام بالتقسيم الأمثل يجب اعتماد معيار للعنقدة clustering criterion و يمكن أن يكون هذا المعيار عام أو محلي: معيار عام مثل الخطأ التربيعي , معيار محلي يعطي البنية المحلية للمعطيات.

يجب التفريق بين معيار العنقدة و خوارزمية العنقدة حيث أن الخوارزمية هي استعمال عملي لمعيار العنقدة و بالتالي و على سبيل المثال يوجد العديد من خوارزميات عنقدة الخطأ التربيعي و كل منها تستخدم معيار الخطأ التربيعي بشكل مختلف عن الأخرى [20] [24].

- عنقدة الخطأ التربيعي Square-Error Clustering:

تعتبر عنقدة الخطأ التربيعي من أكثر استراتيجيات العنقدة التقسيمية شيوعاً و تستخدم للحصول على التقسيم بين المعطيات و ذلك من اجل عدد محدد من العناقيد بحيث نقلل الخطأ التربيعي [20] [24].

من اجل $n$ نمط (نموذج) في فضاء بـ $L$ بعد فإنه يتم التقسيم إلى $K$ عنقود (مجموعة مترابطة) $(C_1, C_2, ..., C_K)$ بحيث يحوي $C_K$ $n_k$ نمط و كل نمط يوجد فقط في عنقود واحد فقط و عندها $m^{(k)} = (\frac{1}{n_k}) \sum_{i=1}^{n_k} x_i^{(k)}$ هو مركز العنقود حيث $x_i^{(k)}$ هو نمط ينتمي إلى العنقود (المجموعة) $K$.

الخطأ التربيعي للعنقود $C_K$ هو مجموع مربعات المسافة الأقليدية بين كل نمط في $C_K$ و مركز هذا العنقود $m^{(k)}$.

يدعى الخطأ التربيعي أيضاً بالتغاير ضمن العنقود و يعطى بالمعادلة (4-1).

$$(4-1) \quad e^2{}_k = \sum_{i=1}^{n_k} (x_i^{(k)} - m^{(k)})^T (x_i^{(k)} - m^{(k)})$$

و عندها الخطأ التربيعي لعملية التجميع كاملة من اجل $K$ عنقود هو مجموع التغايرات الداخلية لكل العناقيد و يعطى وفق المعادلة (4-2).

$$(4-2) \quad E^2{}_k = \sum_{k=1}^{K} e^2{}_k$$

و تصبح الغاية من عملية عنقدة الخطأ التربيعي هي إيجاد التقسيم الذي يعطي اقل قيمة لـ $E^2{}_k$ و التقسيم الناتج للمعطيات يدعى تقسيم التغاير الأصغري minimum variance partition.

الخوارزمية التكرارية العامة للعنقدة التقسيمية هي كالتالي [20]:

1. اختيار تقسيم أولي للمعطيات إلى $K$ عنقود (مجموعة). ثم تكرار العمليات من 2 إلى 5 حتى نحصل على عناقيد مستقرة.
2. تشكيل تقسيمات جديدة بإسناد كل نمط إلى اقرب مركز للعناقيد السابقة.
3. حساب مراكز العناقيد الجديدة و ذلك بحساب المتوسط.
4. تكرار الخطوة 2 و 3 حتى الوصول إلى امثل قيمة لتابع معيار العنقدة المختار (مثلاً معيار الخطأ التربيعي).
5. تعديل عدد العناقيد بدمج أو تقسيم العناقيد الناتجة أو بإلغاء العناقيد الصغيرة أو إلغاء النقاط الشاذة.

الخوارزمية السابقة من دون الخطوة 5 تدعى خوارزمية K-mean. يوضح الشكل (4-7) النقاط الشاذة.

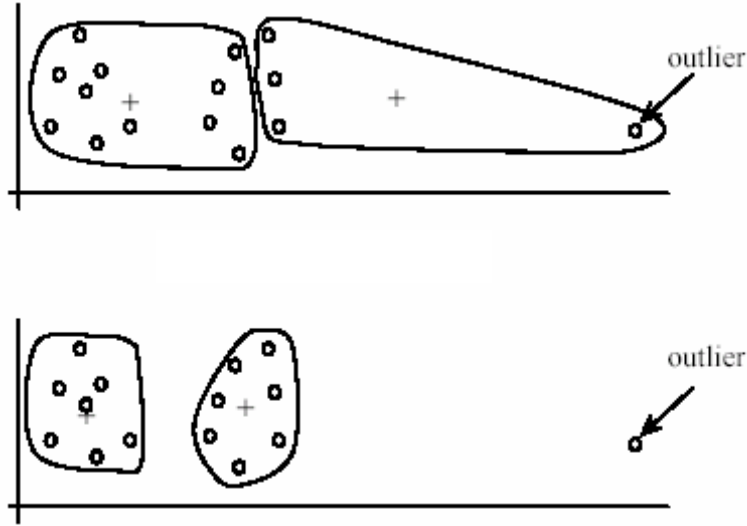

الشكل (7-4) النقاط الشاذة

خوارزمية *K-mean* هي خوارزمية فعالة حاسوبياً و تعطي نتائج ممتازة عندما تكون العناقيد مضغوطة و مفصولة جيداً في فضاء الخواص و ذات شكل شبه كروي hyperspherical في الفضاء متعدد الأبعاد. يمكن استخدام مسافات أخرى غير المسافة الأقليدية لتعريف الخطأ التربيعي مثل مسافة مانهاتن.

## 4-5- الخاتمة:

تضمن هذا الفصل معلومات عامة حول تمييز الأنماط مع شرح سريع للطرق المعروفة من اجل القيام بهذه العملية مثل مطابقة القوالب و المطابقة البنيوية إضافة إلى آليات التصنيف الإحصائي و تم التركيز على شرح التصنيف غير الموجه و بالأخص تقنية العنقدة التقسيمية و ذلك لأهمية التصنيف غير الموجه في تطبيقات تمييز الأنماط.

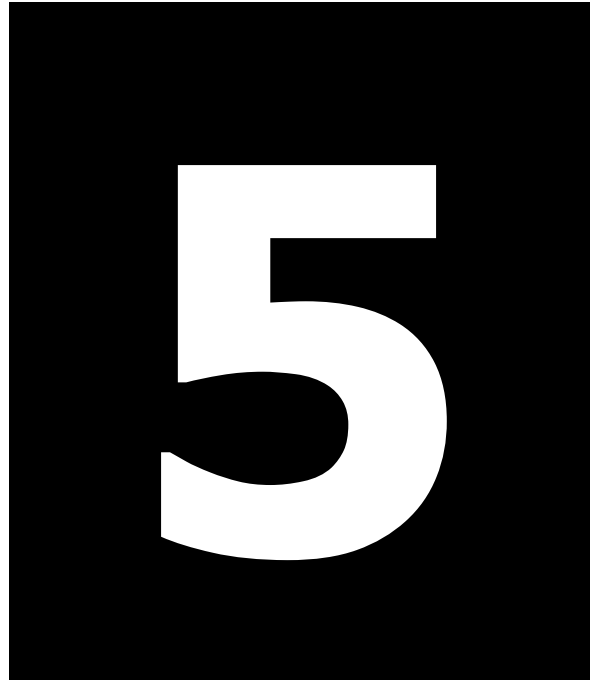

# الفصل الخامس: الخوارزميات و النماذج المناعية

# 5- الخوارزميات و النماذج المناعية:

## 5-1- مقدمة:

يتم تمييز الأنماط في النظام المناعي بخلق ذاكرة مناعية قادرة على تمييز أنماط المستضدات (العوامل الممرضة) التي يتعرض لها النظام المناعي. و عند التعرض لنفس المستضدات أو أي مستضدات مشابهة لها بنسبة معينة يتم تمييزها مباشرة و ذلك باستخدام الذاكرة المناعية و عندما لا يتم تمييز المستضد من قبل الذاكرة المناعية يعمل النظام على تعلم بنية المستضد الجديد (وفق آليات المناعة المكتسبة) و بناء ذاكرة مناعية خاصة بهذا المستضد لمكافحته عند التعرض له في المرة المقبلة.

يوجد العديد من الخوارزميات المناعية المطورة في مجال تمييز الأنماط و معظمها تم تطويره من اجل تطبيق مخصص. قام Yanfei Zhong [32] بتطوير مصنف يستخدم المبادئ المناعية للقيام بعملية التحليل لصور الأقمار الصناعية و قام بمقارنة النتائج للخوارزمية المطورة مع العديد من الخوارزميات التقليدية المستعملة في نفس المجال و اظهر أن الخوارزمية الجديدة تتفوق على تلك الخوارزميات التقليدية ، و يوجد العديد من الخوارزميات المناعية المطورة في مجال تحليل صور الأقمار الصناعية [33] [34] [35]. كما تم استخدام العديد من الخوارزميات المناعية في مجال معالجة الصور [36] [18] حيث استخدمت من اجل عملية تجزئة الصور. و هناك من دمج المبادئ المناعية مع الشبكات العصبونية [19] من اجل عملية التمييز و هي عملية مهمة. استخدم Kemal Polat [37] خوارزمية مناعية لتشخيص سرطان الثدي. و تم استعمال النظام المناعي الاصطناعي لتحديد الأعطال في نظام توزيع القدرة الكهربائية [38].

سيتم في هذا الفصل شرح الخوارزميات و النماذج المناعية العامة المستخدمة لتمييز الأنماط بشكل عام و التي تصلح لعدة تطبيقات في مجال تمييز الأنماط وليس لتطبيق مخصص. من أهم الخوارزميات العامة في مجال تمييز الأنماط:

1. خوارزمية الانتقاء السلبي.
2. خوارزمية الانتقاء النسيلي.
3. الشبكات المناعية الاصطناعية.

من اجل كل خوارزمية سيتم ذكر المبادئ المناعية الأساسية المستوحاة منها الخوارزمية و من ثم شرح الخوارزمية الحاسوبية. يوضح الشكل(5-1) الإطار العام لعمل الخوارزميات المناعية ، إن جميع الخوارزميات المناعية غايتها بناء ذاكرة مناعية خاصة بغرض تمييز المستضد الذي

سيشكل النمط المراد تمييزه. سنذكر ضمن هذا الفصل بعض الخوارزميات المناعية المستخدمة في مجال الأمثلة و ذلك يعود إلى علاقة الأمثلة بعملية تمييز الأنماط (عملية التصنيف غير الموجه).

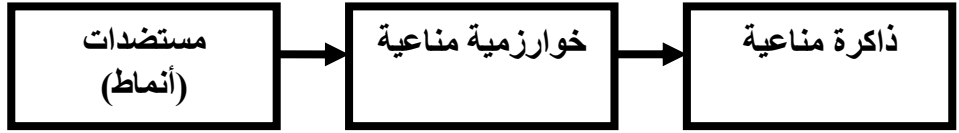

الشكل(5-1) الإطار العام لعمل الخوارزميات المناعية

## 5-2- خوارزمية الانتقاء السلبي: Negative selection algorithm

هذه الخوارزمية مستوحاة من آلية عمل التيموس التي تقوم بإنتاج خلايا لمفاوية متحملة للذات (قادرة على التمييز بين الذاتي و غير الذاتي) و هي تستخدم بشكل أساسي في عملية كشف الشذوذ Anomaly Detection ، حيث غالباً ما يكون لدينا مجموعة من الأمثلة التي تدل على الحالة الطبيعية للنظام و ليس لدينا أي معلومات عن الحالات الأخرى (الغير طبيعية) عندها يُعتمد على مبدأ الانتقاء السلبي لتشكيل مجموعة من الكواشف التي يتم تدريبها على معطيات الحالة الطبيعية و يصبح بذلك لديها القدرة على كشف الحالات غير الطبيعية للنظام (الحالات الشاذة) [3] [4] [10] [39].

تم تطوير أول خوارزمية انتقاء سلبي (NS) Negative Selection من قبل Forrest [39] بغرض القيام بحماية الحاسوب من الفيروسات و المستخدمين غير المرخص لهم.

تعمل خوارزمية الانتقاء السلبي NS على مرحلتين:

1. مرحلة تشكيل الكواشف.
2. مرحلة المراقبة .

**المرحلة الأولى:** الغاية منها تشكيل مجموعة كواشف M لا تميز المجموعة المحمية P (و التي تمثل الحالات الطبيعية للنظام). يتم ذلك بتوليد مجموعة C من العناصر المقترحة ثم اختبارها على P و عند المطابقة يتم إهمال العنصر الذي يميز أي عنصر من P , في نهاية العملية نحصل على مجموعة الكواشف M التي لا تميز P [3] [4] [10] [39]. يوضح الشكل (5-2) مرحلة تشكيل الكواشف.

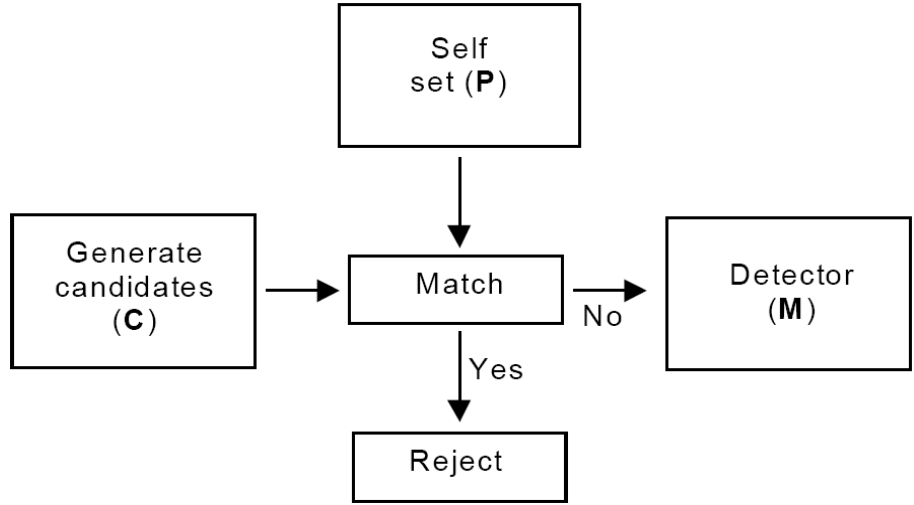

الشكل (5-2) مرحلة تشكيل الكواشف [40]

**المرحلة الثانية**: المراقبة و ذلك بفحص مجموعة ما $P^*$ و عند اكتشاف عنصر غير ذاتي (دخيل) يتم تفعيل عملية التخلص منه [3] [4] [10] [39]. يوضح الشكل (5-3) مرحلة المراقبة.

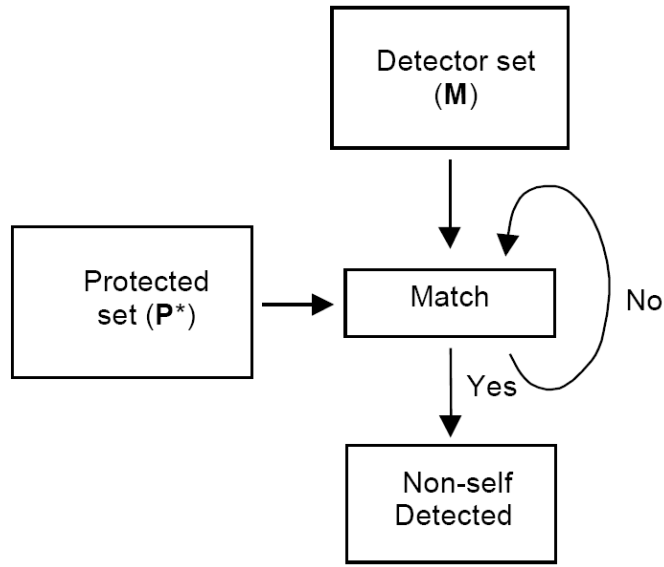

الشكل (5-3) مرحلة المراقبة [40]

من أهم مساوئ هذه الخوارزمية الكلفة الحسابية العالية لتوليد الكواشف و التي تزداد أسياً مع ازدياد حجم المجموعة الذاتية (المحمية). لحل هذه المشكلة طور **D'haeseleer** [41] خوارزميات لتوليد الكواشف بحيث تكون الكلفة الحسابية متناسبة خطياً مع حجم معطيات الدخل.

### 5-2-1- علاقة كشف الشذوذ بعملية التصنيف:

أن عملية كشف الشذوذ المشروحة سابقاً ما هي فعلياً إلا عملية تصنيف موجه supervised classification إذ أن جوهر عملية كشف الشذوذ يعتمد على معطيات التدريب و التي تحوي أمثلة عن صنف وحيد هو الصنف الذاتي أو الطبيعي في حين أن معطيات الاختبار تحوي صنفين يعبر احدهم عن الذاتي و الآخر عن غير الذاتي. إذاً الاختلاف بين كشف الشذوذ و عملية التصنيف التقليدية هو انه في الطريقة التقليدية معطيات التدريب تحوي أمثلة عن كل الأصناف (كل من الذاتي و غير الذاتي) [42].

بالرغم من هذا الاختلاف يكمن التشابه في ما يلي [42]:

1. كل منهما يقسم المعطيات إلى معطيات تدريب و معطيات اختبار و الخوارزمية يجب أن تتعلم من معطيات التدريب و تطبق النتائج على معطيات الاختبار .
2. نتيجة التدريب هي نموذج التصنيف classification model (النموذج الذي يصنف معطيات الاختبار إلى عدة أصناف).
3. يتم تقييم كشف الشذوذ (أو فعلياً في النظام المناعي هو انتقاء سلبي NS) بنفس الطرق التي تقيم بها عملية التصنيف التقليدية مثل مقدار الدقة في التصنيف أو منحني ROC Receiver Operating Characteristic Curve.

### 5-2-2- تطبيقات خوارزمية الانتقاء السلبي NS :

معظم تطبيقات NS في مجال كشف الشذوذ و قد استخدمت في العديد من التطبيقات من أهمها مجال حماية الحاسوب و الشبكة [39] [43] [44] ، كما استخدمت خوارزمية الانتقاء السلبي NS لتجزئة الصورة و ذلك ببناء مجموعة من الكواشف التي ستتخلص من كل شيء في الصورة ما عدا الصنف المراد تمييزه [45] حيث تم تطبيق الخوارزمية على صور للأرض مأخوذة من الفضاء بواسطة حساس أشعة تحت الحمراء حيث بينت التجارب أن الخوارزمية أعطت نتائج ممتازة ، استعملت NS لكشف التغير في معطيات متسلسلة زمنياً حيث تم توليد الكواشف اعتماداً على الحالة الطبيعية للإشارة و من ثم المراقبة و في حال حدوث تغير في الإشارة تتفعل مجموعة من تلك الكواشف المولدة حسب نسبة التغير الحاصل [46]. يوضح الشكل (5-4) مثال على الإشارة المختبرة و يبين الشكل (5-5) عدد الكواشف التي تفعلت عند حدوث الخلل. من المجالات الأخرى التي تم تطبيق خوارزمية NS عليها هو مجال تحليل صور الأسنان [47] ، كما استخدمت NS في كشف العطل في طائرة [48].

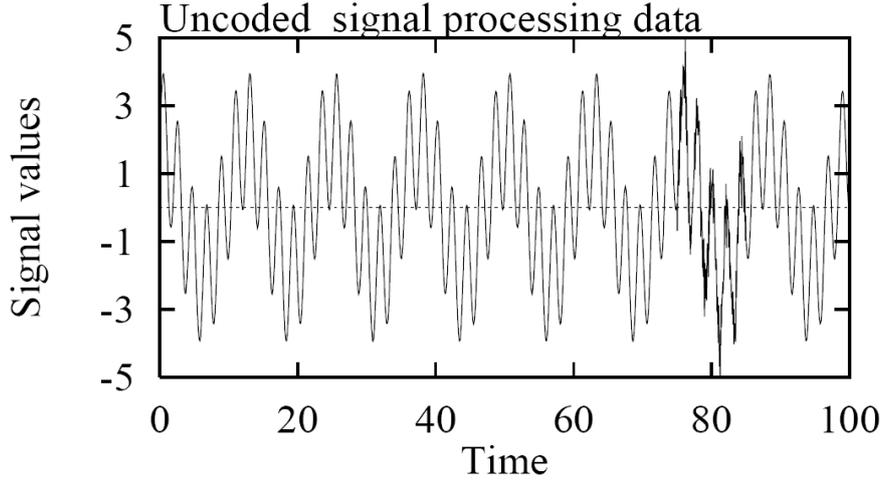

الشكل (4-5) الإشارة المختبرة [46]

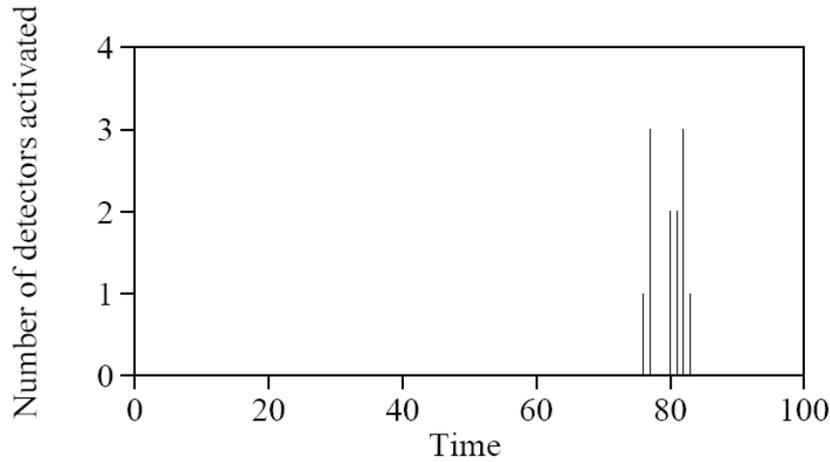

الشكل (5-5) عدد الكواشف التي تفعلت عند حدوث الخلل [46]

## 5-3- خوارزمية الانتقاء النسيلي:

## Clonal Selection Algorithm (CSA)

يتلخص مبدأ الانتقاء النسيلي المستخدم من قبل النظام المناعي في أن الخلية التي تميز المستضد يتم انتقائها للتكاثر و التغاير أما التي لا تميزه يتم إهمالها. أثناء عملية التكاثر يتم إخضاع الخلايا إلى عملية إنضاج للتآلف affinity maturation و هي عملية تعمل على زيادة التآلف للخلايا المنتقاة (زيادة قدرة التمييز).

تـم تطـوير أول خوارزميـة حاسـوبية تعتمـد علـى مبـدأ الانتقـاء النسـيلي عـام 2000 مـن قبـل Leandro N. de Castro [49] ، اعتمدت الخوارزمية على عمليتين أساسيتين في النظام المناعي البيولوجي بغرض إيجاد الحلول هما مبدأ الانتقاء النسيلي و عمليات إنضاج التآلف. استخدمت هذه الخوارزمية في تمييز الأنماط و الأمثلة [3] [4] [50]. كما طُورت العديد من خوارزميات الانتقاء النسيلي بغرض تحسين أداء الخوارزمية الأولى و من اجل القيام بعمليات هندسية مختلفة ، و سندرس منها الخوارزميات التالية:

1. خوارزمية CLONALG.
2. خوارزمية CLONALG التفرعية.
3. خوارزمية CLONCLAS.

### 5-3-1- المبادئ المناعية التي استخدمتها خوارزمية الانتقاء النسيلي:

تم استخدام عدة مبادئ من النظام المناعي البيولوجي لتحقيق خوارزمية الانتقاء النسيلي:
1. تحصيل ذاكرة مناعية مختصة حسب المستضد.
2. انتقاء و استنسال الخلايا ذات التحفيز الأكبر (التآلف الاكبر).
3. عمليات إنضاج التآلف.
4. صيانة التنوع.

ضمن جميع خوارزميات الانتقاء النسيلي يتم اعتبار أن الضد و الخلية اللمفاوية B (المستقبل على سطحها) هما شيء واحد و لا يتم التفريق بينهما.

### 5-3-2- الخطوات العامة لخوارزمية الانتقاء النسيلي:

بشكل عام تتبع جميع خوارزميات الانتقاء النسيلي الخطوات العامة التالية:
1. توليد مجموعة من الحلول المقترحة عشوائياً (مجموعة من الأضداد أو الخلايا).
2. طالما معايير التوقف لم تتحقق القيام بما يلي:
   1،2. استنسال الخلايا بناءً على مقدار التآلف لكل منها.
   2،2. تطبيق آليات إنضاج التآلف (مثل الطفرة المعززة).
   3،2. تطبيق آليات الانتقاء لتشكيل الذاكرة المناعية(مجموعة الحلول الجديدة).
   4،2. صيانة التنوع ضمن مجموعة الحلول.

و في نهايـة الخوارزميـة فـان الـذاكرة المناعيـة المطـورة سـتمثل الحلـول التـي حصـلت عليهـا الخوارزميـة. يوضـح الشـكل (5-6) المخطـط التـدفقي لخوارزميـة الانتقـاء النسـيلي العامـة المستخدمة لتمييز الأنماط.

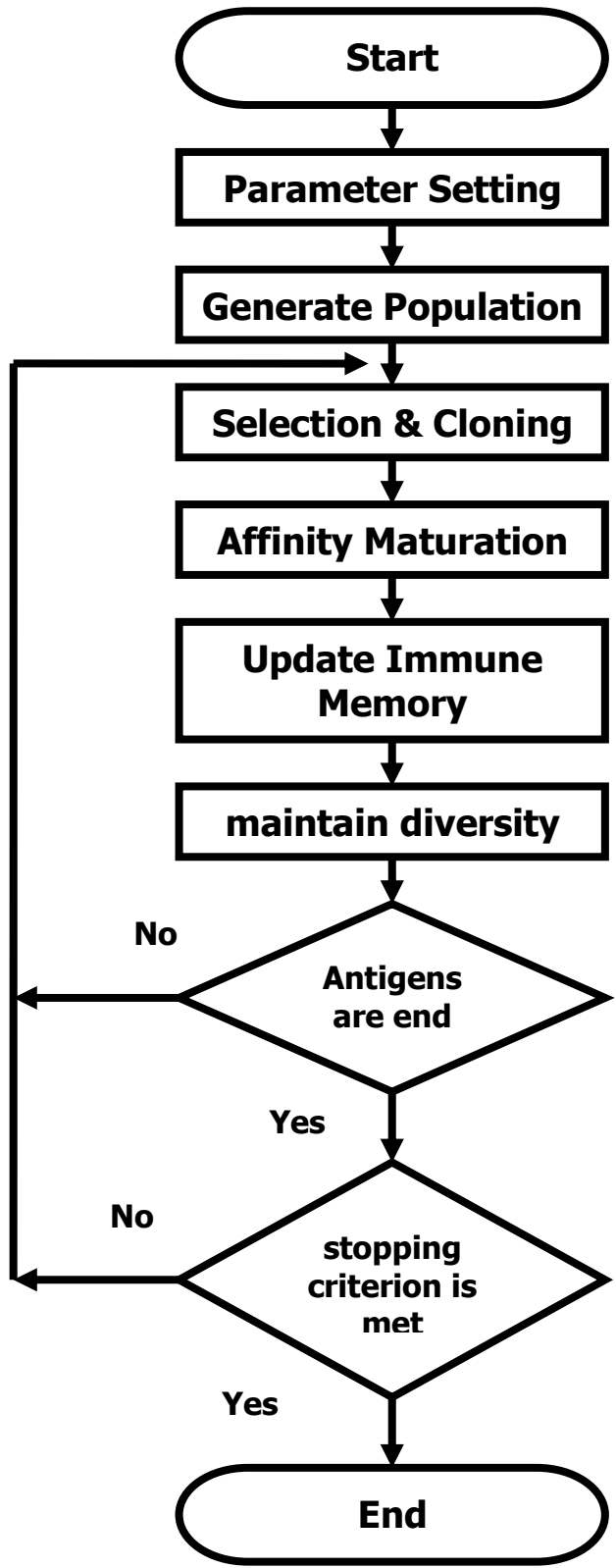

الشكل (6-5) المخطط التدفقي لخوارزمية الانتقاء النسيلي

### 5-3-3- خوارزمية CLONALG:

CLONALG [50] هي أول خوارزمية انتقاء نسيلي و صممت كأداة حاسوبية قادرة على القيام بأعمال مثل تعلم الآلة و تمييز الأنماط و تحصيل الذاكرة. تستخدم CLONALG في تمييز مجموعة من المستضدات $Ag$ عددها $m$ بحيث تولد مجموعة من الأضداد المطورة $Ab_m$ حيث كل ضد $Ab_{mi}$ يمثل الذاكرة المناعية المقابلة لكل مستضد $Ag_i$.

خطوات CLONALG هي كالتالي [3] [4] [5] [50]:

1. يتم في البداية توليد مجموعة $Ab$ من الأضداد (الحلول المقترحة) عددها $N$ و بشكل عشوائي بحيث تكون مؤلفة من مجموعات فرعية وهي: مجموعة خلايا الذاكرة $Ab_m$ و مجموعة باقي الحلول $Ab_r$ حيث يكون $Ab=Ab_m+Ab_r$ و $N=m+r$.

2. من اجل كل جيل يتم ما يلي:

    2،1. من اجل كل مستضد $Ag_i$ ضمن مجموعة المستضدات يتم ما يلي:

    2،1،1. يقاس مقدار التآلف بين كل عنصر من $Ab$ و $Ag_i$ و يوضع الناتج ضمن شعاع $f$.

    2،1،2. نختار أفضل $n$ ضد من المجموعة $Ab$ لتشكل المجموعة $Ab_n$ و هي الأضداد ذات التآلف الأكبر مع المستضد $Ag_i$.

    2،1،3. نقوم باستنسال المجموعة المنتقاة (أفضل $n$ فرد) و ذلك حسب التآلف المقاس لكل ضد بحيث يكون مقدار الاستنسال يتزايد حسب ازدياد التآلف مع المستضد و بذلك ينشأ لدينا مجموعة مؤقتة $C$.

    2،1،4. يطبق على $C$ عمليات إنضاج للتآلف affinity maturate بحيث تكون العمليات متناسبة عكساً مع مقدار التآلف و ينتج عندها المجموعة $C^*$.

    2،1،5. يتم قياس مقدار التآلف من جديد بين عناصر $C^*$ و $Ag_i$ و يوضع الناتج ضمن شعاع $f^*$.

    2،1،6. يتم اختيار من $C^*$ العنصر ذو التآلف الأعلى $Ab_{mc}$ كمرشح ليكون الذاكرة المناعية الجديدة.

    2،1،7. في حال تآلف $Ab_{mc}$ أعلى من تآلف الذاكرة المناعية $Ab_{mi}$ يتم استبدال الذاكرة المناعية به ليشكل الذاكرة المناعية الجديدة.

    2،1،8. يتم توليد $d$ حل عنصر عشوائياً و نستبدل بها اقل العناصر تآلفاً من $Ab_r$

    2،2. يتم تكرار العملية 2،1. حتى نهاية المستضدات.

3. نكرر العمليات في 2 حتى نهاية الأجيال اللازمة.

بنتيجة الخوارزمية نحصل على ذاكرة مناعية قادرة على تمييز مجموعة المستضدات.

قام Leandro N. de Castro [50] باختبار الخوارزمية باستخدامها في عملية تمييز مجموعة من الأعداد المرمزة بشكل ثنائي [50] حيث اعتبر كل عدد مؤلف من مصفوفة من الأرقام الثنائية 0,1 بحجم مصفوفة يساوي 12X10 ، و تم اعتماد ترميز للأضداد لتمثيل الحلول بحيث ان كل ضد مؤلف من مصفوفة سطر 1X120 بحيث يتألف السطر الوحيد من تتالي اسطر المصفوفة المرمزة للعدد. لقياس التآلف تم استخدم مسافة هامينغ لقياس مقدار التآلف بين الضد و المستضد ، إذاً عند تحقق التتام الكامل بين الضد و المستضد تكون المسافة العظمى هي 120 و يكون عندها الضد $Ab_{mi}$ هو نسخة معكوسة رقمياً عن المستضد $Ag_i$ و نحصل عند تحقق ذلك على تمييز تام للمستضد.

### 5-3-3-1- آلية الاستنسال المستخدمة في CLONALG:

تم افتراض أن مجموعة الأضداد المنتقاة للاستنسال ذات العدد n تم فرزها تنازلياً و ذلك وفقاً للتآلف مع المستضد ومن ثم تم تطبيق المعادلة (5-1) لتحديد عدد النسخ المولدة عن كل ضد ضمن المجموعة[3] [4] [5] [50].

$$(1-5) \quad nc_i = round(\frac{\beta.n}{i})$$

حيث $\beta$ هو معامل الاستنسال . بالتالي يكون عدد الحلول الكلية الناتجة ضمن المجموعة المؤقتة C هو وفق المعادلة التالية:

$$(2-5) \quad Nc = \sum_{i=1}^{N} round(\frac{\beta.n}{i})$$

مثلاً : من اجل N=100 و $\beta=1$ فان العنصر الأول (ذو التآلف الأعلى) يتم تكوين 100 نسخة له و العنصر الذي يليه يتم تكوين 50 نسخة له و هكذا حسب المعادلة (5-2).

### 5-3-3-2- عمليات إنضاج التآلف:

اعتمد نضوج التآلف لكل ضد على الطفرة المعززة جسدياً حيث تم جعل الطفرة متناسبة عكساً مع مقدار التآلف للضد. بسبب استخدام الترميز الثنائي لترميز الأضداد تم استخدام الطفرة متعددة النقاط multipoint mutation [3]. يوضح الشكل (5-7) آلية الطفرة متعددة النقاط حيث عند حدوث طفرة في نقطة هذا يؤدي إلى تغير الرقم من '1' إلى '0' أو بالعكس.

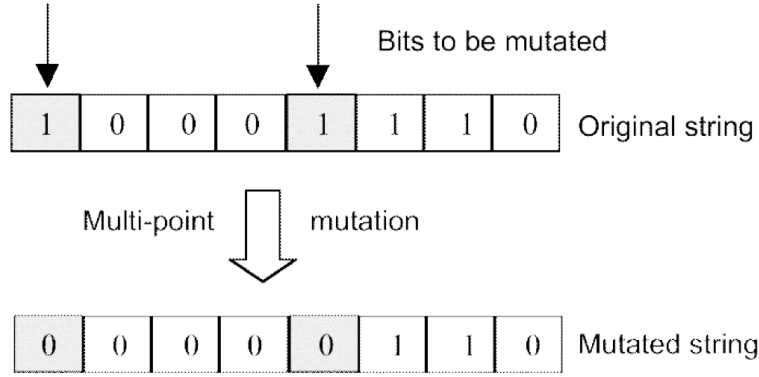

الشكل (5-7) آلية الطفرة متعددة النقاط [3]

تم ربط احتمال الطفرة بمقدار التآلف وفق المعادلة (5-3) و بناءً عليها فإن احتمال حدوث الطفرة متناسب عكساً مع التآلف و كلما ازداد التآلف أصبح الحل أفضل و ينقص عندها احتمال الطفرة [3] [4] [50]:

$$\alpha = e^{-\rho f} \qquad (5\text{-}3)$$

حيث $\alpha$ هو احتمال معدل الطفرة و $\rho$ هو معامل الهبوط أما $f$ هو مقدار التآلف الذي تم تقييسه normalized ضمن المجال [0,1]. يبين الشكل (5-8) المنحني ما بين معدل الطفرة و التآلف من اجل قيم مختلفة لمعامل الهبوط $\rho$ حيث مع ازدياد معامل الهبوط يزداد الفرق في معدلات الطفرة بين الحلول الجيدة و الحلول السيئة [3] [4] [5] [50].

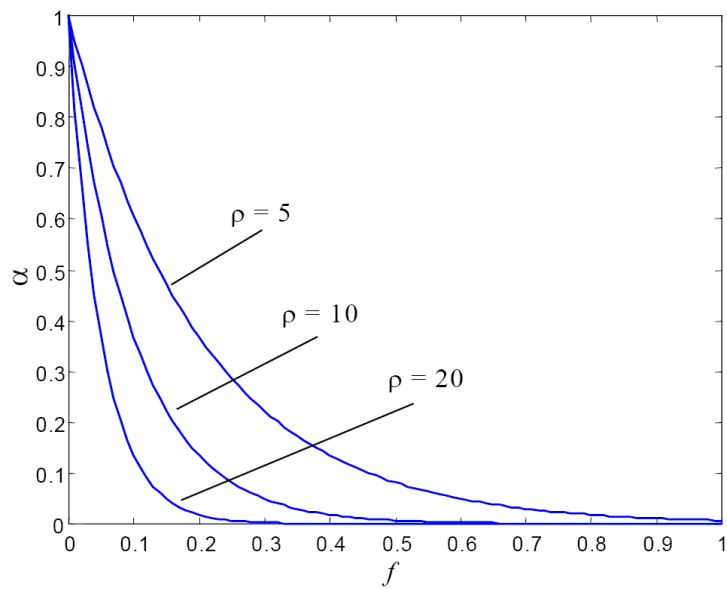

الشكل (5-8) منحني معدل الطفرة بالنسبة للتآلف [50]

بشكل عام يختلف تطبيق الخوارزمية من مسألة إلى أخرى و ذلك حسب نمذجة الضد و المستضد (عملية الترميز) و آلية قياس التآلف. من المهم جدا التحكم بالطفرة و يختلف أداء الخوارزمية حسب حجم مجموعة الحلول و حجم الذاكرة المناعية.

### 5-3-4- خوارزمية CLONALG التفرعية:

قام Andrew Watkins [51] بدراسة تنفيذ CLONALG بشكل تفرعي كمحاولة لتسريع الخوارزمية للوصول إلى الحل حيث تم تقسيم مجموعة المستضدات (الأنماط التي نريد تمييزها) إلى عدة مجموعات و تم تطبيق CLONALG على كل مجموعة على حدة بشكل تفرعي ثم في النهاية تم تجميع الذاكرة المناعية الناتجة عن كل عملية تفرعية لتكوين النتيجة النهائية. يوضح الشكل (5-9) عملية تنفيذ CLONALG بشكل تفرعي.

كانت الخوارزمية الجديدة [51] أسرع من الخوارزمية التقليدية بكثير و ذلك يعود إلى أن عملية تشكيل جزءٍ من الذاكرة المناعية كان أسرع بكثير من محاولة إنشاء الذاكرة المناعية كاملة.

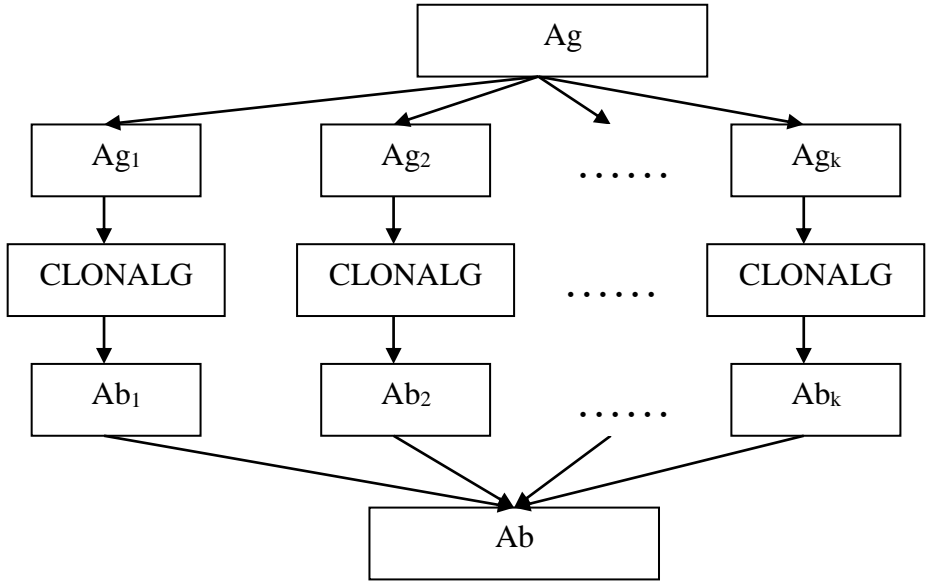

الشكل (5-9) خوارزمية CLONALG التفرعية

يبين الشكل (5-10) ازدياد سرعة الخوارزمية (نسبة الزمن اللازم للوصول للحل من اجل عملية واحدة إلى الزمن اللازم عند تطبيقها على عدة عمليات تفرعية) مع ازدياد عدد العمليات التفرعية.

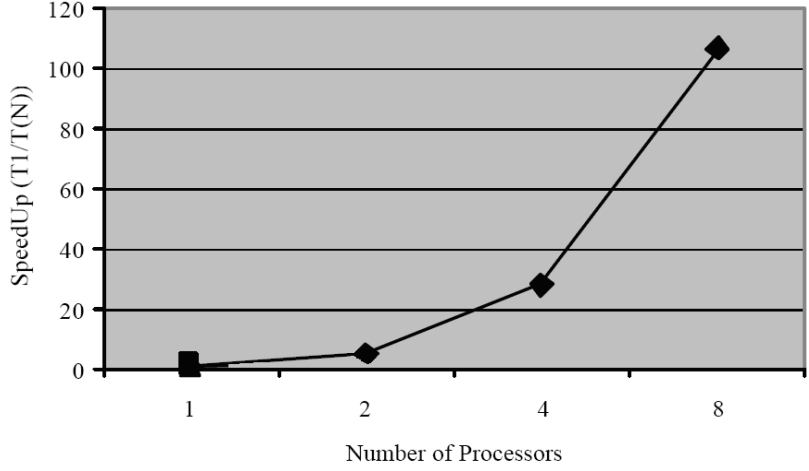

الشكل (5-10) ازدياد سرعة خوارزمية CLONALG التفرعية مع ازدياد عدد العمليات التفرعية [51]

## 5-3-5- خوارزمية التصنيف النسيلي:Clonal Classification (CLONCLAS)

قام White A. Jennifer [52] بتطوير نسخة معدلة من CLONALG سميت CLONCLAS حيث تم استخدامها في عملية التصنيف ، تم تعديل الخوارزمية لعدة أسباب نذكر منها [52]:

1. عند تمييز أحرف أو أرقام فإنه لا فائدة من محاولة تعلم الحرف كما هو في CLONALG لأنه لدينا مسبقاً نسخة من ذلك الحرف , كما أن هذا الحرف لديه عادة عدة أشكال مختلفة الشيء الذي يشكل صنف للنمط و بالتالي فانه من المفيد تكوين ذاكرة مناعية للحرف معممة عن جميع الأشكال المختلفة له أي الصنف و استخدام هذه الذاكرة المناعية للتعرف عليه عندما يكون بشكل غير محدد مسبقاً.
2. أن CLONALG لا تستفيد فعلياً من المجموعات المؤقتة الناشئة أثناء عمل الخوارزمية لأنها تعمل على تطوير مجموعة الذاكرة المناعية دفعة واحدة معاً و تنتقل من مستضد إلى المستضد التالي ضمن نفس الجيل.

لذلك تم تعديل الخوارزمية بحيث تحاول تعلم عدة أشكال للرقم معاً لتشكل صنف واحد حيث تم تعديل معيار التآلف ليشمل جميع الأمثلة عن الرقم ضمن الصنف الواحد و ذلك بجعل المعيار هو مجموع مسافات هامينغ بين الضد و جميع المستضدات ضمن الصنف الواحد و ذلك وفق المعادلة التالية [52]:

$$affinity = \sum_{i=1}^{E} D_i \qquad (4-5)$$

حيث $E$ عدد الأمثلة ضمن الصنف الواحد و $D_i$ هي مسافة هامينغ بين الضد $Ab_{mi}$ و المستضد $Ag_i$ من نفس الصنف و بذلك يشير المعيار إلى مقدار تآلف الضد مع كل المستضدات ضمن الصنف الواحد و بالتالي بدلاً من تعرض الضد لمستضد واحد نعرضه لمجموعة المستضدات ضمن الصنف الواحد و بالنتيجة ستتشكل ذاكرة مناعية معممة عن الصنف بالكامل [52]. يوضح الشكل (5-11) تكوين الخلايا المعممة عن الأصناف.

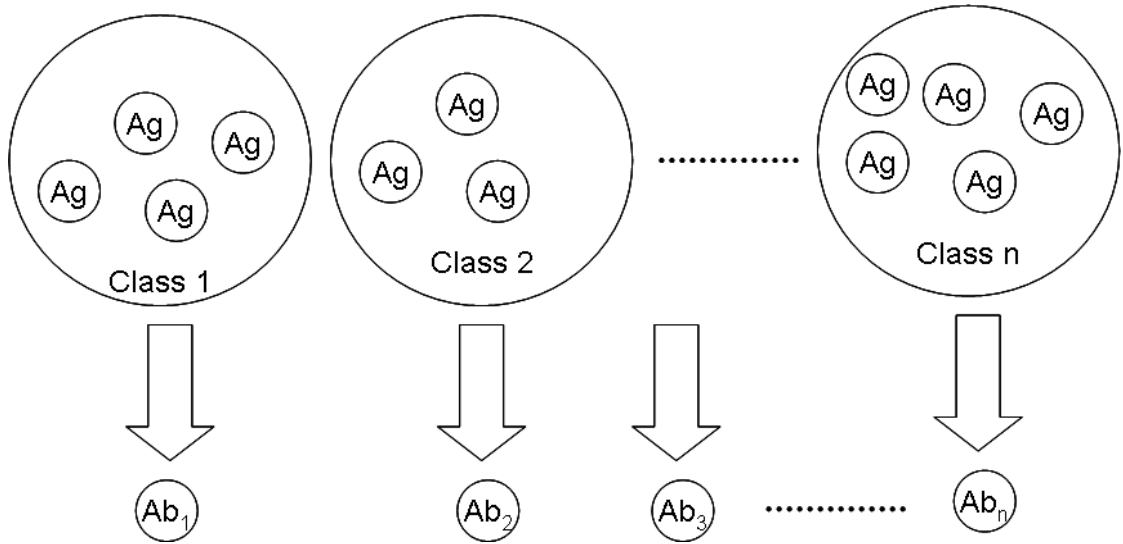

الشكل (5-11) تكوين الخلايا المعممة عن الأصناف

تستخدم الذاكرة المناعية الناتجة بعد عملية التعلم في عملية تمييز الأرقام أو الأحرف الجديدة غير المعروفة حيث يتم إسناد الحرف إلى الصنف الذي يحقق اكبر تآلف مع الذاكرة المناعية المقابلة له لكن بشرط أن يكون هذا التآلف فوق عتبة التآلف الدنيا $\varepsilon$ اللازمة و المحددة مسبقاً. و في حال كان التآلف من اجل جميع الأضداد ضمن الذاكرة المناعية تحت عتبة التآلف يصنف الحرف على انه غير معروف [52].

قام Jennifer A. White [52] بتعديل بنية خوارزمية CLONALG لكي تستطيع الاستفادة من الحلول الجيدة الموجودة في المجموعة المؤقتة *C و أصبحت الخطوات للخوارزمية الجديدة كالتالي:

1. في البداية يتم توليد مجموعة Ab من الحلول المقترحة ذات العدد N و ذلك بشكل عشوائي مؤلفة من مجموعات فرعية: مجموعة خلايا الذاكرة $Ab_m$ و مجموعة باقي الحلول $Ab_r$ و يكون $Ab = Ab_m + Ab_r$ و $N = m + r$ .

2. من اجل كل مستضد $Ag_i$ ضمن مجموعة المستضدات يتم ما يلي:
   
   2،1. من اجل كل جيل يتم :
   
   2،1،1. يقاس مقدار التآلف بين كل عنصر من Ab مع $Ag_i$ و يوضع الناتج ضمن شعاع f.
   
   2،1،2. يتم اختيار أفضل n فرد (ضد) من المجموعة Ab لتشكل المجموعة $Ab_n$ و هي الأضداد ذات التآلف الأكبر مع المستضد $Ag_i$ .
   
   2،1،3. يتم استنسال المجموعة المنتقاة (أفضل n فرد) و ذلك حسب التآلف المقاس لكل ضد بحيث مقدار الاستنسال يتزايد حسب ازدياد التآلف مع المستضد و بذلك ينشأ لدينا مجموعة مؤقتة C.
   
   2،1،4. يطبق على C عمليات إنضاج للتآلف affinity maturate بحيث تكون العمليات متناسبة عكساً مع مقدار التآلف و ينتج عندها المجموعة *C.
   
   2،1،5. يتم قياس مقدار التآلف من جديد بين عناصر *C و $Ag_i$ و نضع الناتج ضمن شعاع *f.
   
   2،1،6. يتم اختيار من *C العنصر ذو التآلف الأعلى $Ab_{mc}$ كمرشح ليكون الذاكرة المناعية الجديدة.
   
   2،1،7. في حال تآلف $Ab_{mc}$ أعلى من تآلف الذاكرة المناعية $Ab_{mi}$ يتم استبدال الذاكرة المناعية به ليشكل الذاكرة المناعية الجديدة.
   
   2،1،8. اختيار أفضل r حل من *C و نستبدل بها عناصر $Ab_r$ بالكامل.
   
   2،1،9. يتم توليد d حل عشوائياً و نستبدل بها اقل العناصر تآلفاً من $Ab_r$.
   
   2،2. يتم تكرار الخطوة 2,1 حتى نهاية جميع الأجيال و بذلك تتكون الذاكرة المناعية المقابلة للمستضد $Ag_i$.

3. يتم تكرار العمليات ضمن 2 حتى نهاية المستضدات و بذلك تتكون الذاكرة المناعية المقابلة لجميع المستضدات.

## 5-4- الشبكة المناعية الاصطناعية التطورية:
## (aiNet) An Evolutionary Artificial Immune Network

تم تطوير خوارزمية aiNet [53] [54] للقيام بعملية تحليل المعطيات ، حيث تستعمل هذه الخوارزمية نظرية الشبكة المناعية التي تفترض أن النظام المناعي هو عبارة عن شبكة منظمة من الخلايا و الجزيئات التي تميز بعضها بعضا حتى في غياب المستضد و بالتالي تفترض التفاعل بين جزيئات النظام المناعي (التفاعل بين الأضداد) بشكل مستمر و متغير مع الزمن و يعمل النظام المناعي على تنظيم عملية تشكيل الشبكات المناعية و ذلك بتثبيط أو تفعيل التفاعل بين الأضداد. تمثل الشبكة الناتجة الصورة الداخلية للمستضد و هي تعكس بنيته.

### 5-4-1- المبادئ المناعية التي استخدمتها aiNet :

تم استخدام عدة مبادئ من النظام المناعي البيولوجي من اجل بناء aiNet و هي:
1. تحصيل ذاكرة مناعية مختصة تعطي الصورة الداخلية للمستضد.
2. انتقاء و استنسال الخلايا ذات التحفيز الأكبر (التآلف الأكبر).
3. عمليات إنضاج التآلف.
4. صيانة التنوع .
5. التفاعل بين الأضداد (تمييز بعضها بعضاً) .
6. تثبيط الشبكة المناعية (تثبيط التفاعل بين الأضداد) .

ضمن aiNet يتم اعتبار أن الضد و الخلية اللمفاوية B هما شيء واحد و لا يتم التمييز بينهما. بناءً على ذلك فإن aiNet تستعمل مبدأ الانتقاء النسيلي إضافة لمبدأ الشبكة المناعية للقيام بتحليل المعطيات.

### 5-4-2- بنية aiNet :

aiNet هي عملياً مخطط ذو حواف موزونة graph edge-weighted و تتألف من مجموعة من العقد التي هي الخلايا أو الأضداد و مجموعة من الأزواج بين هذه الخلايا هي الحواف edges و مع كل حافة رقم مرفق هو الوزن أو قوة الاتصال [53] [54].

إن aiNet هي شبكة تطورية لأنها تعتمد على الإستراتيجية التطورية للتحكم بديناميكية الشبكة و مرونتها (تستعمل التغاير الجيني لمجموعة من الأفراد). مهمة الخوارزمية الحصول على الصورة الداخلية للعناقيد الموجودة ضمن المعطيات (مجموعات المعطيات الطبيعية) و ذلك بشكل عناقيد شبكية. بافتراض أن الدخل كما هو مبين في الشكل (5-12) (a) فإن الخرج المفترض سيكون كما هو مبين في الشكل (5-12) (b) [53] [54].

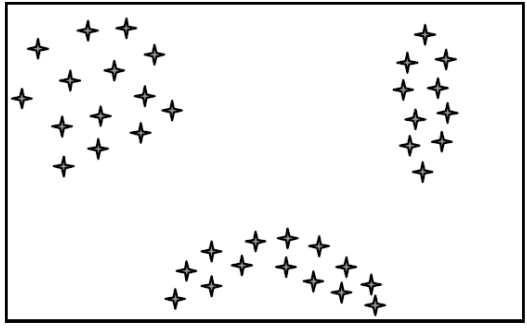 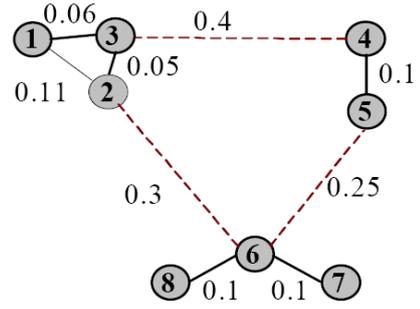

(a)            (b)

الشكل (5-12) بنية الشبكة المناعية الاصطناعية[53]

يجب ملاحظة أن عدد خلايا الشبكة اقل بكثير من عدد النماذج في المعطيات و هذا ملائم جداً لضغط المعطيات لذلك تستخدم aiNet لضغط المعطيات و التخلص من المعلومات الفائضة ضمن المعطيات.

### 5-4-3- آلية عمل خوارزمية aiNet :

وفقاً لنظرية الشبكة المناعية فإن الخلايا ستتنافس على تمييز المستضد و الخلية التي تنجح في التمييز أكثر سيتم تفعيلها و من ثم إكثارها أما التي تفشل سيتم التخلص منها إضافة إلى فكرة أن عملية التمييز Ab-Ab بين الأضداد ستؤدي إلى تثبيط الشبكة network suppression.

تتم عملية التثبيط ضمن نموذج aiNet بالتخلص من الخلايا التي تميز الذات (تميز بعضها) و ذلك بالاعتماد على عتبة للتثبيط $\sigma_s$، كذلك فإن العلاقة بين الضد و المستضد Ab-Ag ستحدد بناءً على التآلف بينهما dij (مقدار عدم التماثل) ، أما العلاقة بين الأضداد Ab-Ab ستكون وفقاً للتآلف بينها sij (مقدار التماثل) [53] [54]. يوضح الشكل (5-13) المخطط التدفقي الذي يحدد الخطوات العامة لخوارزمية aiNet. الغاية في aiNet بناء مجموعة من خلايا الذاكرة المناعية القادرة على تمييز و تمثيل بنية و تنظيم المعطيات بحيث يتم فعلياً ضغط المعطيات ، و كلما كانت هذه الخلايا معممة كانت نسبة الضغط اكبر.

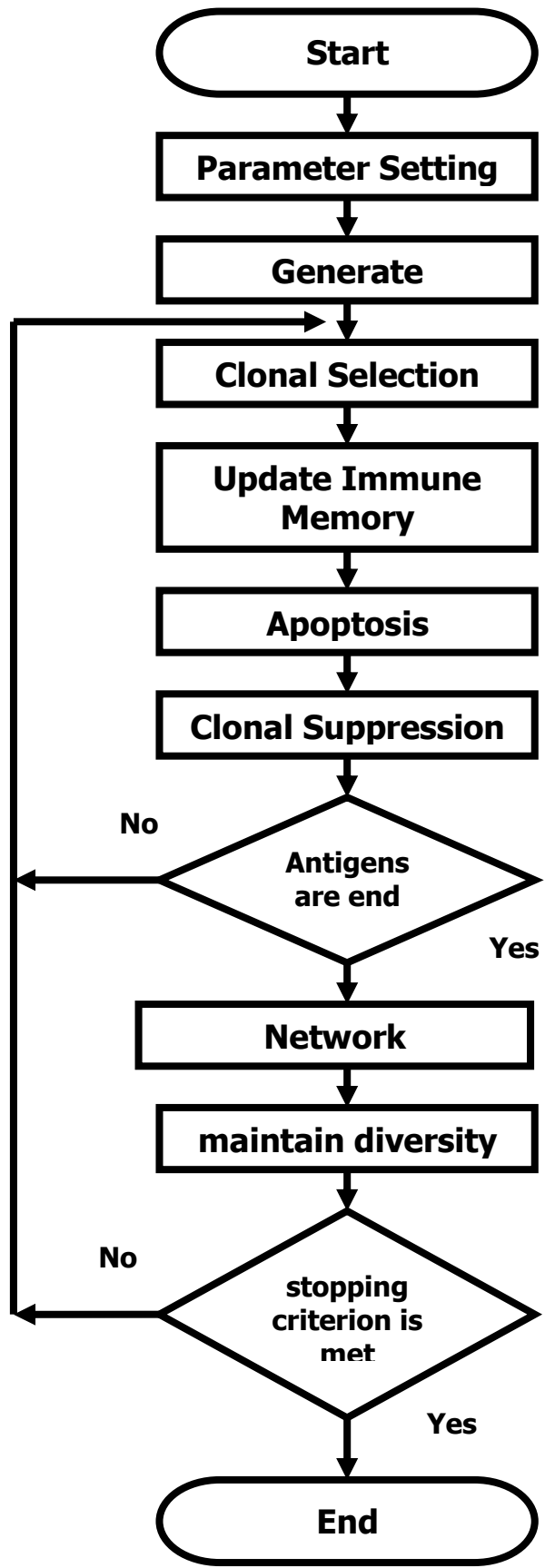

الشكل (5-13) المخطط التدفقي لخوارزمية aiNet

في الخوارزمية تم استخدام الرموز التالية:

- X المعطيات ذات العدد Np .
- c مصفوفة تحوي خلايا الشبكة ذات العدد Nt .
- M مصفوفة تحوي N خلية ذاكرة .
- Nc عدد الخلايا المستنسلة من الخلية المحفزة (المفعلة) .
- D مصفوفة عدم التماثل تحوي (Ag-Ab) dij .
- S مصفوفة التماثل تحوي (Ab-Ab) sij .
- n عدد الخلايا المنتقاة للاستنسال و عمليات الإنضاج .
- $ζ$ نسبة الخلايا المنتقاة من الخلايا الناضجة .
- $σ_d$ عتبة الموت الطبيعي .
- $σ_s$ عتبة التثبيط .

تعمل خوارزمية aiNet للتعلم كالتالي [53] [54]:

1. من اجل كل تكرار يتم ما يلي:
    1،1. من اجل كل مستضد i يتم ما يلي:
        1،1،1. قياس التآلف dij لكل خلايا الشبكة.
        2،1،1. اختيار n خلية من الشبكة ذات التآلف الأعلى.
        3،1،1. استنسال الخلايا المنتقاة بناءً على التآلف.
        4،1،1. تطبيق آلية الطفرة على الناتج.
        5،1،1. حساب D للخلايا المحسنة.
        6،1،1. نختار نسبة $ζ$% من الخلايا عالية التآلف لتشكل Mp مصفوفة خلايا ذاكرة جزئية.
        7،1،1. التخلص من الخلايا ذات التآلف اقل من $σ_d$ عتبة الموت و بذلك يقلل حجم Mp .
        8،1،1. قياس تآلف Ab-Ab للشبكة sij .
        9،1،1. التخلص من الأضداد ذات sij> $σ_s$ (تثبيط نسيلي).
        10،1،1. وصل Mp مع مجموعة الخلايا الكلية C ([C;Mp] ← C).
    2،1. حساب S و التخلص من الخلايا ذات sij> $σ_s$ (تثبيط شبكي).
    3،1. استبدال أسوء r% حل بحلول عشوائية جديدة.
2. اختبار معيار التوقف.

نلاحظ أن الخطوات من 1.1.1 حتى 7.1.1 هي عملية الانتقاء النسيلي أما العمليات الباقية فهي لإجراءات الشبكة المناعية. تم اعتماد المسافة الأقليدية من اجل قياس التآلف كمعيار للتماثل و معيار لعدم التماثل. بعد عملية التعلم الخلايا الناتجة تمثل الصورة الداخلية للمستضدات و ينتج عنها خرج الشبكة M و هو مصفوفة إحداثيات خلايا الذاكرة و S مصفوفة التآلفات بين هذه الخلايا و هي تحدد أي من الخلايا متصل مع الآخر و مقدار هذه الاتصال و بذلك تحدد البنية العامة للشبكة.

### 5-4-4- آلية الطفرة في aiNet :

تحدث الطفرة على أعداد حقيقية و ذلك بزيادة أو إنقاص العدد الحقيقي بمقدار معين و هو عمليا مقدار الطفرة. تم احداث الطفرة في aiNet باستخدام المعادلة التالية [34]:

$$c* = c - \alpha(c - Ag) \qquad (5-5)$$

توجيه الطفرة محكوم بإشارة الحد $(c - Ag)$ و ذلك يضمن اتجاه الطفرة نحو المستضد. و يحدد المعامل $\alpha$ فعلياً مقدار الإزاحة أي مقدار الطفرة الحاصلة و هو متحول عشوائي ينتمي إلى مجال محدد. تم ربط $\alpha$ بالتآلف فعندما يزداد التآلف تقل قيمة $\alpha$ و يتم تحقيق ذلك عن طريق ربط مجال المتحول العشوائي $\alpha$ عكساً مع التآلف [54].

### 5-4-5- استخراج المعلومة من بنية aiNET المدربة :

يمكن تحديد بنية الشبكة الناتجة ببساطة بواسطة وصلات المصفوفة S لكن ذلك سيجعل عملية تفسير الشبكة و استخراج المعلومة صعباً للغاية و خاصة عندما يكون عدد الأبعاد أكثر من ثلاثة. يتم استخراج المعلومة عن طريق تحديد عدد العناقيد و تحديد الخلايا التي تنتمي لكل عنقود آخذين بعين الاعتبار شكل كل عنقود. و ذلك باستخدام تحليل الشجرة ذات الامتداد الأصغري [55] minimal spanning tree للشبكة الناتجة للحصول على العناقيد ضمن المعطيات.

### 5-5- الخوارزميات المناعية المستخدمة في الأمثلة:

إن عملية الأمثلة ذات ارتباط وثيق بتمييز الأنماط فكما ذكرنا سابقاً إن عملية التصنيف غير الموجه (العنقدة التقسيمية) تعتمد على أمثلة معيار عنقدة معين بالتالي إن خوارزميات العنقدة التقسيمية التقليدية هي فعلياً خوارزميات أمثلة. تم تطوير العديد من الخوارزميات المناعية في مجال الأمثلة و استخدمت هذه الخوارزميات في العديد من المجالات و التطبيقات مثل التخطيط

و الجدولة و تصميم الدارات الرقمية و تصميم أنظمة التحكم. الغاية من هذه الفقرة إعطاء فكرة عن أهم الخوارزميات المناعية التي استخدمت من اجل الحصول على الحل الأمثل لتابع ما أو مسألة ما , معظم هذه الخوارزميات اعتمدت على مبدأ الانتقاء النسيلي.

### 5-5-1- خوارزميات الانتقاء النسيلي المستخدمة في الأمثلة:

قام Leandro N. de Castro [50] بتعديل خوارزمية CLONALG المستخدمة لتميز الأنماط لتقوم بعملية الأمثلة و سميت الخوارزمية المعدلة opt-CLONALG. لا يوجد في عملية الأمثلة مستضد ليتم تمييزه بل تابع $g(.)$ ليتم أمثلته و هو يعبر عن جودة الحل و بالتالي مقدار التآلف يصبح قيمة التابع من اجل ضد معين (حل معين). و بما انه لا يوجد مجموعة مستضدات ليتم تمييزها فان مجموعة الأضداد Ab بالكامل ستكون مجموعة الذاكرة أي لن يكون هناك $Ab_m$ منفصلة. كما انه لن يتم انتقاء خلية واحدة كل مرة لتشكل الذاكرة المناعية لكن سيتم انتقاء N ضد لتشكل المجموعة Ab [3] [4] [5] [50].

تعمل خوارزمية opt-CLONALG كالتالي:
1. البداية: توليد مجموعة Ab من الحلول المقترحة ذات العدد N و ذلك بشكل عشوائي .
2. من اجل كل جيل نقوم بما يلي:
   1,2. نقيس مقدار التآلف لكل عنصر من Ab و ذلك بواسطة التابع $g(.)$ و نضع الناتج ضمن شعاع f .
   2,2. نختار أفضل n فرد (ضد) من المجموعة Ab لتشكل المجموعة Abn و هي الأضداد ذات التآلف الأكبر.
   3,2. نقوم باستنسال المجموعة المنتقاة (أفضل n فرد) و بذلك ينشأ لدينا مجموعة مؤقتة C.
   4,2. نطبق على C عمليات إنضاج للتآلف affinity maturate و ينتج عندها المجموعة *C .
   5,2. نقوم بقياس مقدار التآلف من جديد لعناصر *C و نضع الناتج ضمن شعاع *f .
   6,2. نختار من *C عدد N عنصر ذوات التآلف الأعلى لتشكل $Ab_{mc}$ كمرشح ليكون الذاكرة المناعية الجديدة .
   7,2. نستبدل الأضداد في Ab بالأضداد ذات التآلف الأعلى من $Ab_{mc}$ .
   8,2. توليد d عنصر (حل) عشوائياً و نستبدل بها اقل العناصر تآلفاً من Ab .
3. نعيد العملية 2 حتى نهاية الأجيال اللازمة.

بنهاية الخوارزمية نقوم باختيار الخلية ذات التآلف الأعلى من Ab بحيث تشكل الحل النهائي الناتج عن الخوارزمية.

تم ترميز الحلول بأرقام ثنائية بشكل مشابه للخوارزمية الأصلية المستخدمة لتمييز الأنماط لكن هذه الأرقام تعبر عن أرقام حقيقية ضمن مجال البحث المفروض لإيجاد الحل و للقيام بتقييم الحل و حساب مقدار التآلف يتم تحويل الحل إلى رقم حقيقي كالتالي:

الضد مؤلف من L رقم ثنائي كالتالي $m=[m_0, m_1, ..., m_{L-1}]$ نحسب $z'$ بحيث $z' = \sum_{i=0}^{L-1} m_i 2^i$ عندها القيمة الحقيقة المقابلة للحل هي $z = z_{min} + z' \frac{z_{max} - z_{min}}{2^L - 1}$ حيث $z_{max}$ و $z_{min}$ تعبر عن حدود مجال البحث. ثم يتم قياس التآلف وفقاً لتابع التقييم $g(.)$. آلية الاستنسال هي نفسها المستخدمة في خوارزمية CLONALG الأساسية و آلية الطفرة المعززة هي نفسها أيضاً و كلا العمليتين تتم بشكل متناسب مع مقدار التآلف [50].

هناك العديد من الخوارزميات الأخرى التي اعتمدت على مبدأ الانتقاء النسيلي من اجل الأمثلة و معظمها هو محاولة لتحسين الخوارزمية الأساسية CLONALG. اعتمدت CLONALG على معاملين أساسيين لتوجيه الحلول هما مبدأ الانتقاء النسيلي و آلية الطفرة المعززة. هناك من أضاف معاملات أخرى مثل خوارزمية opt-IA [56] التي استعملت ثلاث معاملات لتوجيه الحلول هي الاستنسال و الطفرة المعززة و معامل التقادم في السن aging operator و تم مقارنة أدائها مع CLONALG و تم إثباث أنها الأفضل [57]. ثم طورت نسخة معدلة من opt-IA هي opt-IMMALG [58] و هي بالإضافة إلى استعمالها المعاملات الثلاث السابقة تستعمل تمثيل حقيقي للحلول.

كما تم تطوير خوارزمية RECSA [59] التي استخدمت طريقة لتحرير (تعديل) المستقبلات بحيث تكوّن مع الطفرة المعززة آليات إنضاج التآلف إضافة إلى أنها استعملت آلية احتمالية لانتقاء النخبة. و يوضح الشكل (5-14) مخطط الخوارزمية. كما قدمت نسخة محسنة من خوارزمية الانتقاء النسيلي [60] وهي تستعمل متسلسلة عشوائية Chaotic لتوليد مجموعة الحلول العشوائية كما تستخدم آلية طفرة تعتمد على المسافة بين الجينات إضافة إلى عملية تحرير المستقبلات من اجل علمية إنضاج التآلف.

نسخة أخرى من خوارزمية الانتقاء النسيلي ناقشت فكرة مهمة و هي أن الانتقاء النسيلي عملياً لا يقوم بعملية العبور crossover مثل الخوارزميات الجينية و بالتالي لا يوجد مشاركة بالمعرفة بين الحلول و لتحقيق عملية مشاركة استعملت مبادئ من نظرية الشبكة المناعية و تم جعل

الطفرة تتعلق بالتآلف إضافة إلى مقدار التفعيل من قبل الخلايا الأخرى (هناك تفاعل بين الخلايا و يعتمد على الجينات) [61].

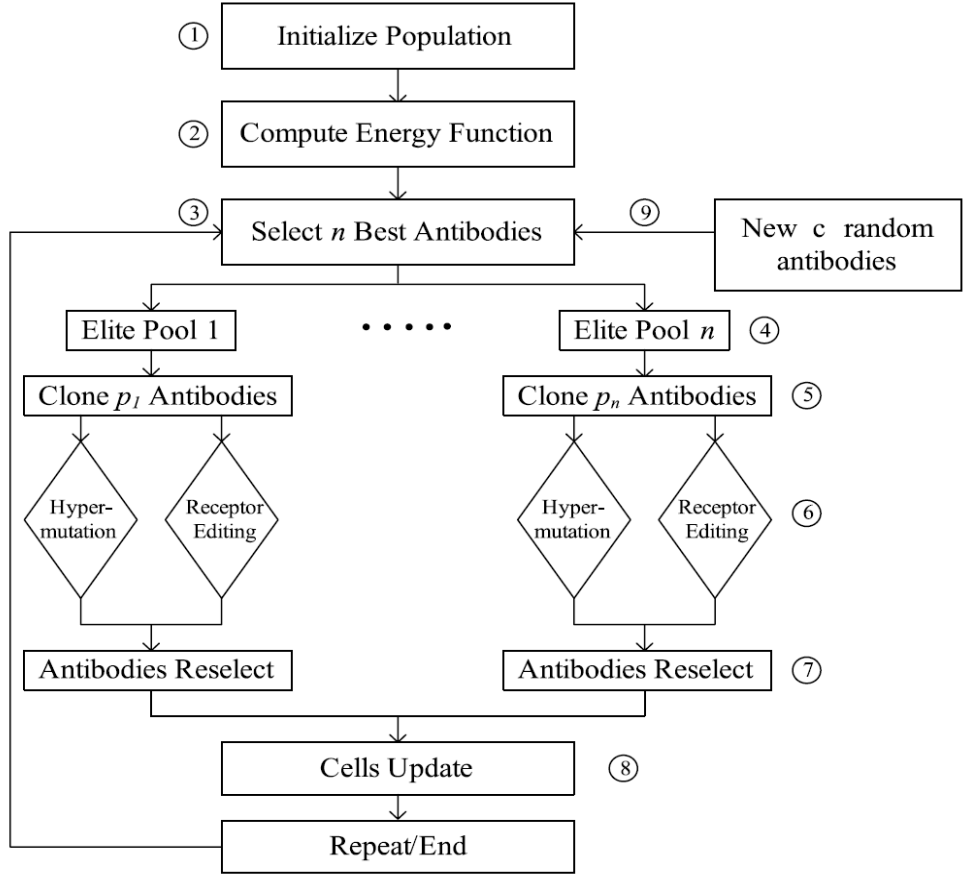

Flow diagram of RECSA.

الشكل (5-14) مخطط عمل خوارزمية RECSA [59]

قام Khilwani sh [62] بتطوير خوارزمية FCA حيث تم تعديل العديد من العمليات إذ استعملت الخوارزمية آلية طفرة تفرعية (PM) parallel mutation و هي تستعمل نوعين من الطفرة معاً : طفرة غوصية من اجل الطفرات صغيرة الخطوة و طفرة ذات توزع Cauchy و ذلك من اجل الطفرة ذات الخطوة الكبيرة. كما استعملت معادلة جديدة لتحديد مقدار التوسع النسيلي و معدل الطفرة و استخدمت أيضا متسلسلات عشوائية لتوليد الحلول العشوائية كما استخدمت آلية احتمالية للانتقاء الحلول و ذلك وفق مبدأ عجلة الروليت. يوضح الشكل (5-15) مخطط خوارزمية FCA.

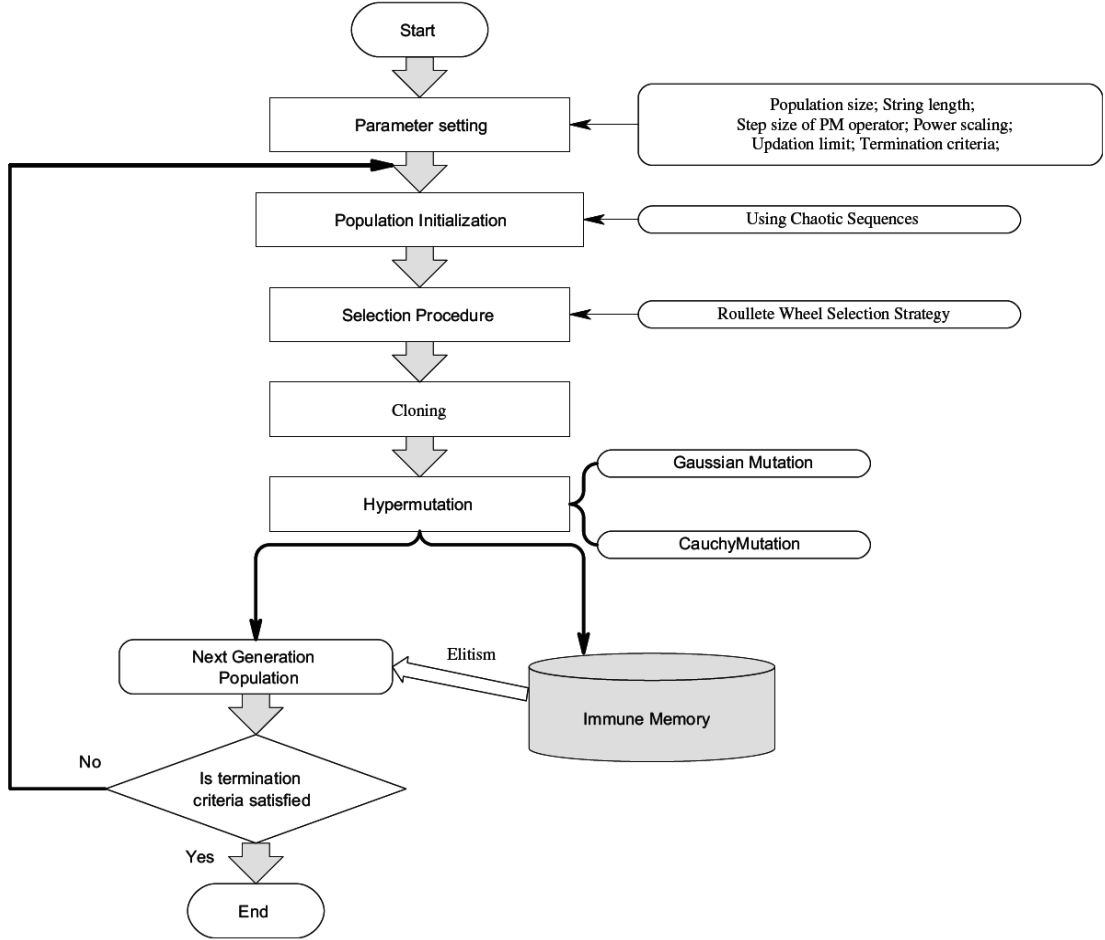

الشكل (5-15) مخطط عمل خوارزمية FCA [62]

## 5-6- الخاتمة:

كانت الغاية من هذا الفصل إعطاء فكرة شاملة عن معظم الخوارزميات المناعية التي تم تطويرها في مجال تعلم الآلة بشكل عام ، حيث احتوى هذا الفصل شرحاً مفصلاً لأهم الخوارزميات المناعية المستخدمة في مجال تمييز الأنماط بشكل عام و هي:

1. خوارزمية الانتقاء السلبي.
2. خوارزمية الانتقاء النسيلي.
3. الشبكات المناعية الاصطناعية.

كما تم شرح بعض الخوارزميات المناعية المستخدمة في مجال الأمثَلة بسبب ارتباط عملية الأمثَلة بعملية تمييز الأنماط.

# 6

# الفصل السادس: تصميم خوارزمية مناعية للقيام بعملية التصنيف غير الموجه

# 6- تصميم خوارزمية مناعية للقيام بعملية التصنيف غير الموجه:

## 6-1- مقدمة:

قمنا باختبار خوارزمية الانتقاء النسيلي إضافة إلى خوارزمية الشبكة المناعية الاصطناعية و دراسة اثر تغير معاملات تلك الخوارزميات على أداء الخوارزمية و مدى أهمية هذه المعاملات في تحسين الأداء. و بناءً على دراسة النظام المناعي الاصطناعي AIS و النماذج و الخوارزميات المناعية المطورة قمنا بتصميم خوارزمية مناعية تعتمد على مبدأ الانتقاء النسيلي في النظام المناعي البيولوجي للقيام بعملية تصنيف غير موجه للمعطيات unsupervised classification. الخوارزمية المطورة هي خوارزمية متكيفة مع المعطيات و تعمل علي تغيير معاملاتها أوتوماتيكياً مع المعطيات من اجل تحسين الأداء و تسريع الوصول إلى الحل و قد تم اختبار أدائها و مقارنته مع الخوارزميات المعروفة في مجال التصنيف غير الموجه مثل K-means حيث وجدنا من نتائج الاختبار أن الخوارزمية الجديدة كانت أكثر دقة في التصنيف و أكثر وثوقية.

ينقسم هذا الفصل إلى قسمين أساسيين: القسم الأول يضم الاختبارات التي أجريناها على الخوارزميات المناعية المستخدمة في تمييز الأنماط (خوارزميات الانتقاء النسيلي و خوارزمية aiNet) أما القسم الثاني يشرح الخوارزمية المناعية التي قمنا بتصميمها من اجل القيام بعملية التصنيف غير الموجه و الاختبارات التي أجريناها عليها حيث تم اختبارها على عدة مجموعات للمعطيات.

## 6-2- اختبار خوارزميات الانتقاء النسيلي:

تعتمد خوارزميتي CLONALG أو CLONCLAS على الكثير من المحددات المضبوطة من قبل الإنسان وهذه احد مساوئ هذه الخوارزميات. من هذه المحددات: حجم مجموعة الحلول N و عدد الحلول المنتقاة للاستنسال n و معامل الاستنسال β و عدد الأجيال gen و عتبة التآلف اللازمة $\varepsilon$ و معامل الهبوط $\rho$ و عدد الحلول العشوائية الجديدة d. و لذلك لا بد من دراسة تأثير تغير هذه المعاملات على أداء الخوارزمية. من اجل ذلك قمنا ببناء أدوات خاصة ضمن برنامج MATLAB [63] [64] [65] خصيصاً من اجل عملية الاختبار.

تم اختبار اثر تغيير المعاملات على CLONALG و على نسخة معدلة من CLONCLAS حيث تم تعديلها لدراسة اثر الاستبدال في الخطوة 8.1.2. من خطوات الخوارزمية المبينة في الفقرة 5-3-5- إذ بدلاً من استبدال كامل عناصر $Ab_r$ بأفضل الحلول من $C^*$ تم تعديل ذلك

ليصبح استبدال k أسوء حل من $Ab_r$ بأفضل k حل من *C و تم الإشارة للنسخة المعدلة بـ Improved CLONALG و بالتالي يدخل معامل جديد في الخوارزمية هو k.

قمنا بإجراء مجموعة من التجارب اعتماداً على توابع MATLAB و التي تم تطويرها خصيصاً لدراسة هذه الخوارزميات و ذلك على تطبيق تمييز مجموعة من الأرقام المرمزة ثنائياً و تم استخدام نفس الرموز التي استخدمت في [50] لاختبار CLONALG. يوضح الشكل (6-1) الرموز المستخدمة للاختبارات. تم جعل d=0 في جميع التجارب حيث وجدنا أن تغيير قيمته لم يؤثر على أداء الخوارزميات أثناء الاختبار.

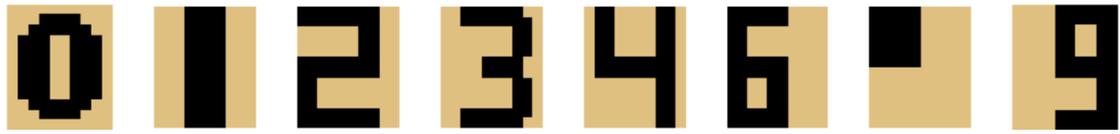

الشكل (6-1) الرموز المستخدمة للاختبارات

### 6-2-1- اختبار عام لـ CLONALG:

قمنا باختبار عام لـ CLONALG حيث تم ضبط محددات الخوارزمية كالتالي : N=10 و n=3 و d=0 و β=10 و 4.8=$\rho$ و عتبة التآلف $\varepsilon = 1$ و ذلك للوصول إلى التآلف التام. وجد أن خوارزمية CLONALG استغرقت اقل من 10 ثوان للوصول للحل و ذلك بعد اقل من 170 جيل.

### 6-2-2- اختبار عام لـ Improved CLONALG:

قمنا باختبار عام للنسخة المعدلة CLONALG Improved حيث تم ضبط محددات الخوارزمية كالتالي : N=10 و n=3 و d=0 و k=3 و β=10 و 4.8=$\rho$ و عتبة التآلف $\varepsilon = 1$ و ذلك للوصول إلى التآلف التام. وجد أن CLONALG Improved استغرقت اقل من 3 ثوان للوصول للحل وذلك بعد اقل من 60 جيل. الشيء الواضح من نتائج الاختبارين أن الخوارزمية المعدلة أسرع في الوصول للحل.

### 6-2-3- ازدياد التآلف مع مرور أجيال الخوارزمية:

الهدف من هذا الاختبار هو دراسة ازدياد التآلف مع مرور أجيال الخوارزمية حيث تم ضبط محددات الخوارزميات كالتالي : N=10 و n=3 و d=0 و k=3 و β=10 و 4.8=$\rho$ و عتبة التآلف $\varepsilon = 1$ و الحد الأقصى لعدد الأجيال هو gen=150. التآلف المقاس هو

متوسط قيم التآلف لكل الأضداد ضمن الذاكرة المناعية عند جيل معين. يبين الشكل (2-6) نتائج الاختبار حيث نلاحظ أن الخوارزمية المحسنة تصل للتآلف الكامل قبل خوارزمية CLONALG بكثير.

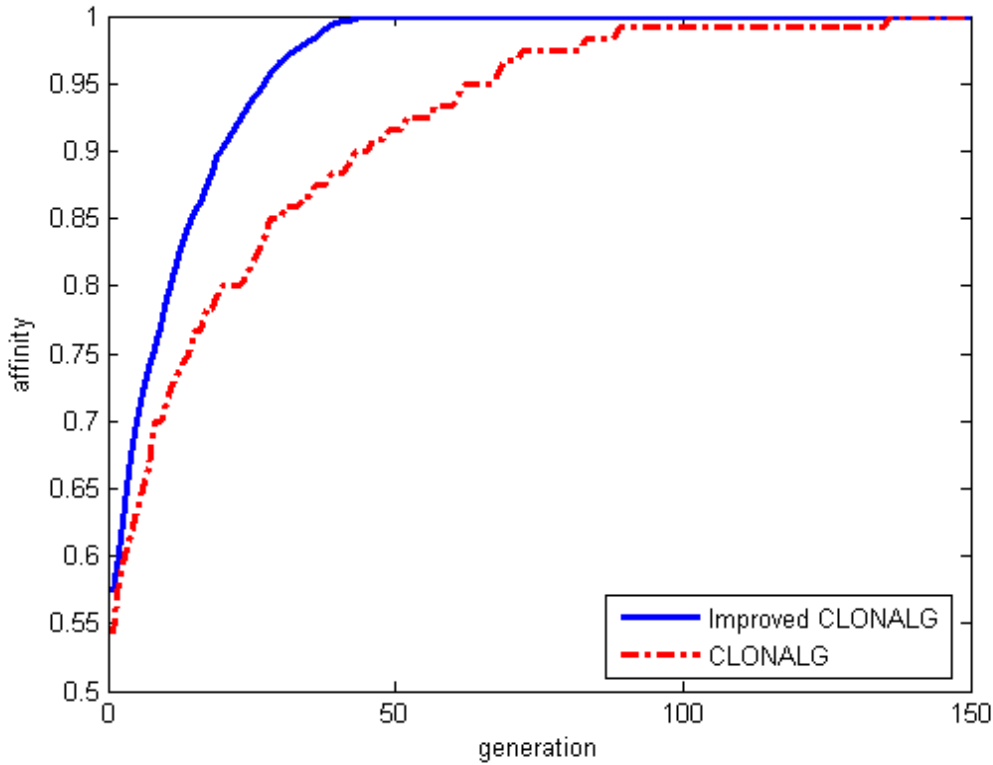

الشكل (2-6) ازدياد التآلف مع مرور الأجيال

## 6-2-4- اثر عتبة التآلف على أداء الخوارزميات:

الغاية من هذا الاختبار دراسة عدد الأجيال اللازمة من الخوارزميات للوصول إلى قيم مختلفة من عتبات التآلف حيث تم ضبط محددات الخوارزميات كالتالي : $N=10$ و $n=3$ و $d=0$ و $k=3$ و $\beta=10$ و $\rho=4.8$ تم الاختبار عند عدة قيم لعتبة التآلف $\varepsilon$. قمنا بتكرار التجربة عشرة مرات عند كل قيمة لـ $\varepsilon$ و من ثم تم اخذ المتوسط للقيم الناتجة (عدد الأجيال). يبين الشكل (6-3) نتائج الاختبار حيث نلاحظ أنه عند ازدياد عتبة التآلف فإن عدد الأجيال اللازم للوصول للحل يزداد بشكل كبير ، و نلاحظ أيضاً الأداء الأفضل للخوارزمية المعدلة.

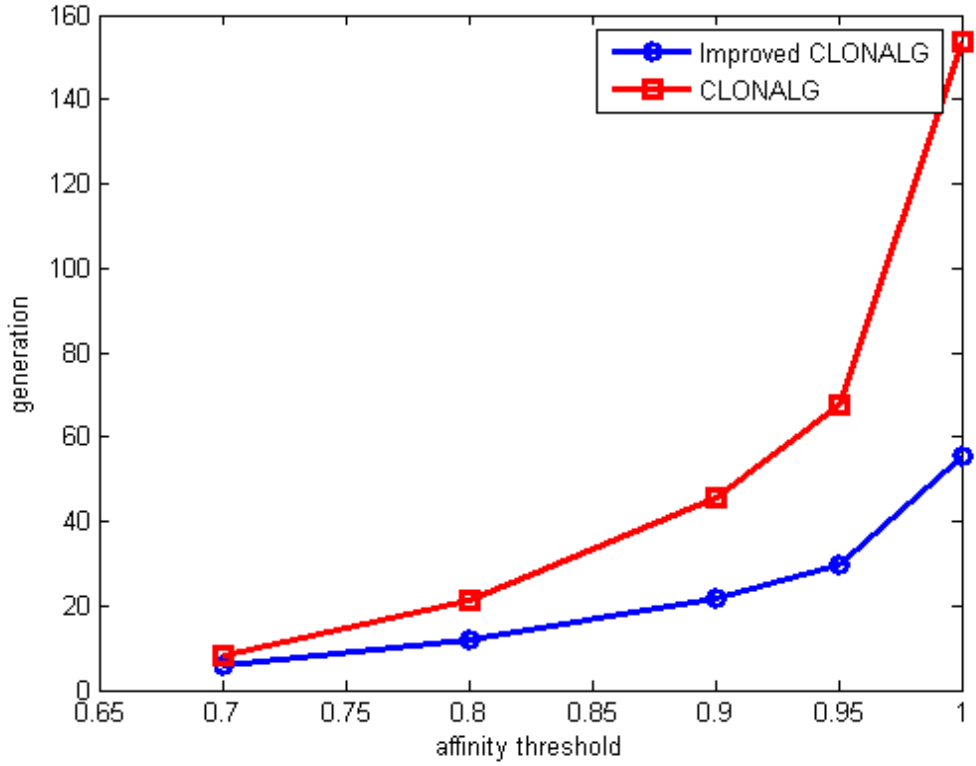

الشكل (3-6) عدد الأجيال اللازمة للوصول للحل مع عتبة التآلف

## 6-2-5- اثر معامل الاستنسال على أداء الخوارزميات:

الغاية من هذا الاختبار دراسة أداء الخوارزميات مع تغير معامل الاستنسال β و لدراسة ذلك تم تحديد عدد الأجيال اللازمة من الخوارزميات للوصول إلى التآلف الكامل من اجل قيم مختلفة لمعامل الاستنسال حيث تم ضبط محددات الخوارزميات كالتالي : N=10 و n=3 و d=0 و k=3 و $\rho$=4.8 تم الاختبار عند عدة قيم لمعامل الاستنسال. قمنا بتكرار التجربة عشر مرات عند كل قيمة لـ β و من ثم تم اخذ المتوسط للقيم الناتجة(عدد الأجيال). يبين الشكل (6-4) نتائج الاختبار حيث نلاحظ أن المعامل β مهم جداُ و هو يحدد فعلياً سرعة الخوارزمية و كلما ازدادت قيمة β قل عدد الأجيال اللازمة للوصول للحل. كما نلاحظ أن الخوارزمية المعدلة ذات أداء أفضل بكثير من CLONALG.

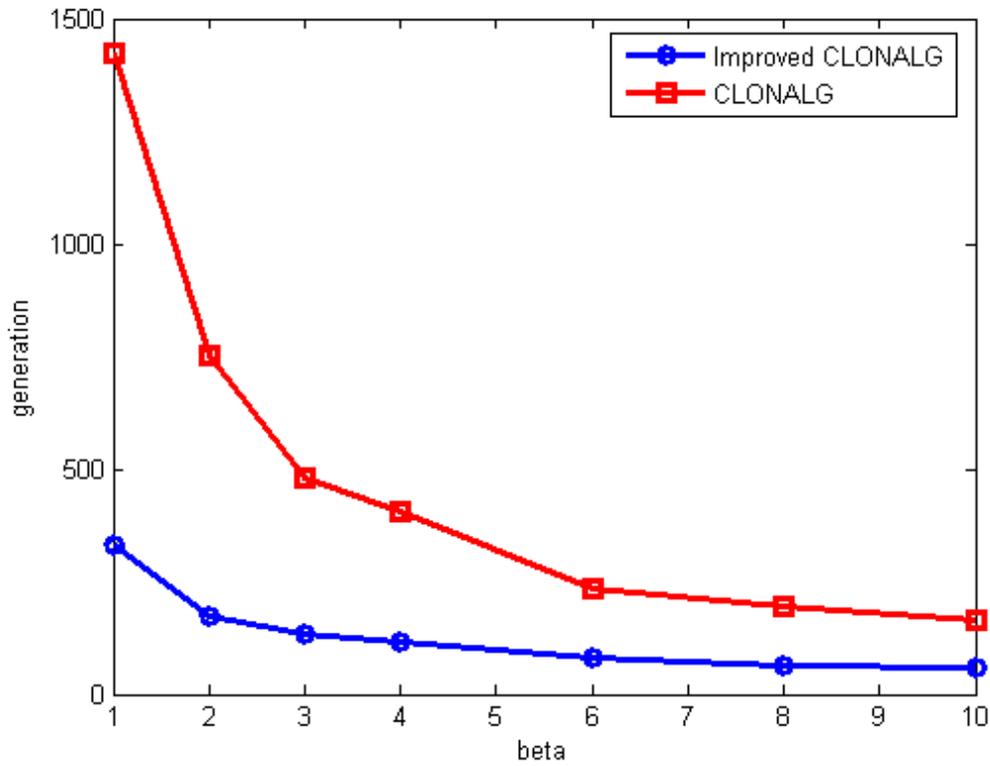

الشكل (4-6) عدد الأجيال اللازمة للوصول للحل مع معامل الاستنسال

### 6-2-6- اثر عدد الخلايا المنتقاة للاستنسال على أداء الخوارزميات:

الغاية من هذا الاختبار دراسة اثر تغير عدد الخلايا المنتقاة للاستنسال n على أداء الخوارزميات و لدراسة ذلك تم تحديد عدد الأجيال اللازمة من الخوارزميات للوصول إلى التآلف الكامل من اجل قيم مختلفة لـ n حيث تم ضبط محددات الخوارزميات كالتالي: N=10 و d=0 و k=3 و β=10 و $\rho$=4.8 تم الاختبار عند عدة قيم لـ n. تم تكرار التجربة عشر مرات عند كل قيمة لـ n و من ثم تم اخذ المتوسط للقيم الناتجة(عدد الأجيال). يبين الشكل (6-5) نتائج الاختبار حيث نلاحظ أن ازدياد n يسرع الخوارزميات و يقلل عدد الأجيال اللازمة للوصول للحل. كما نلاحظ أن الخوارزمية المعدلة ذات أداء أفضل من خوارزمية CLONALG كما أن تأثير تغير n اكبر على الخوارزمية المعدلة منه في خوارزمية CLONALG.

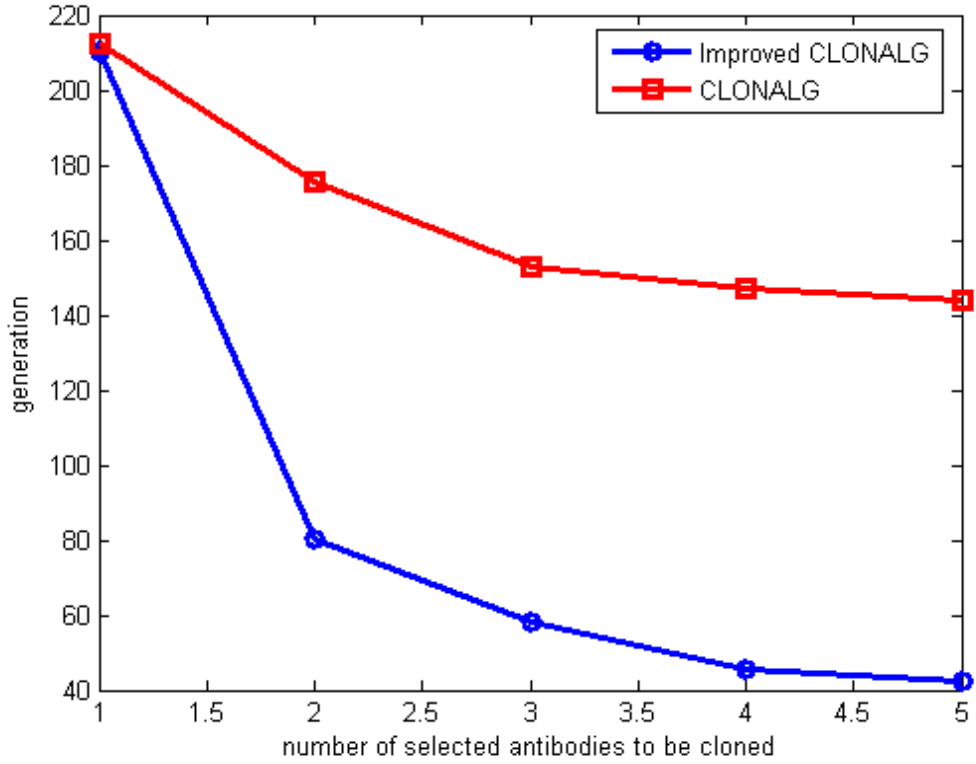

الشكل (5-6) الأجيال اللازمة للوصول للحل مع عدد الخلايا المنتقاة للاستنسال

### 6-2-7- اثر عدد الخلايا المستبدلة k على أداء الخوارزمية المعدلة:

الغاية من هذا الاختبار دراسة اثر استبدال أسوء الحلول في المجموعة $Ab_r$ بـ k أفضل حل من المجموع $C^*$ على أداء الخوارزمية المحسنة و لدراسة ذلك تم تحديد عدد الأجيال اللازمة من الخوارزمية للوصول إلى التآلف الكامل من اجل قيم مختلفة لـ k حيث تم ضبط محددات الخوارزميات كالتالي : $N=10$ و $n=3$ و $d=0$ و $\beta=15$ و $\rho=4.8$ تم الاختبار عند عدة قيم لـ k. قمنا بتكرار التجربة عشر مرات عند كل قيمة لـ k و من ثم تم اخذ المتوسط للقيم الناتجة(عدد الأجيال). يبين الشكل (6-6) نتائج الاختبار حيث نلاحظ أنه يكفي أن تكون $k=1$ لتصبح الخوارزمية المعدلة أسرع بكثير لكن فعليا من اجل زيادة k أكثر من الواحد فإنه لا يوجد تحسن كبير في الأداء و بالتالي نكتفي بجعل $k=1$.

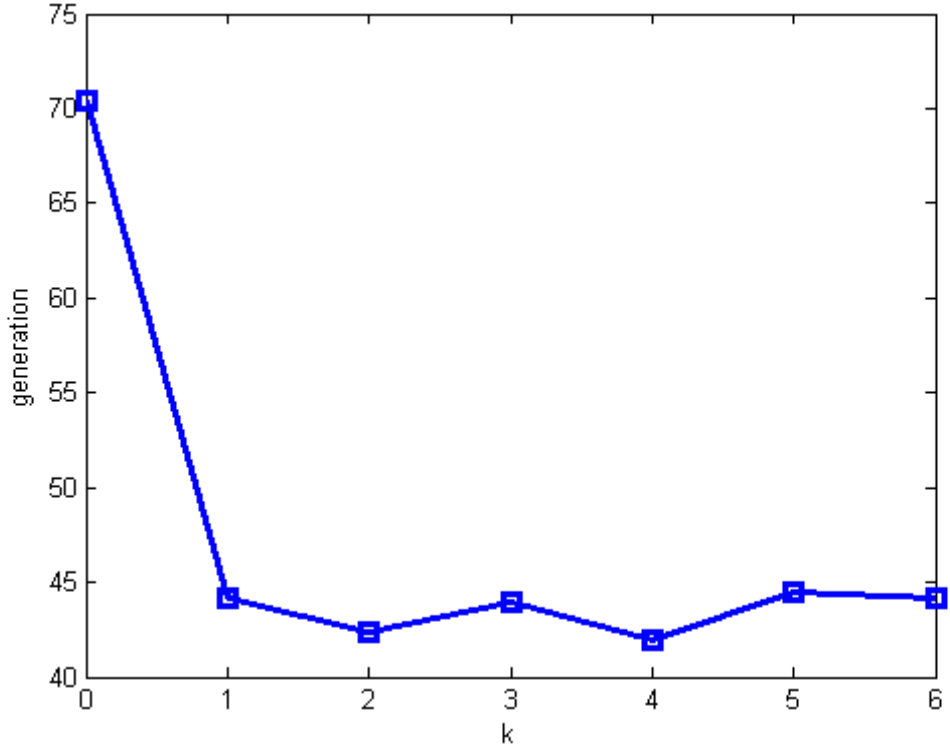

الشكل (6-6) الأجيال اللازمة للوصول للحل مع عدد الخلايا المستبدلة k

### 6-2-8- اثر معامل هبوط منحني احتمال الطفرة على أداء الخوارزميات:

الغاية من هذا الاختبار دراسة تأثير تغير $\rho$ على أداء الخوارزميات و لدراسة ذلك تم تحديد متوسط التآلف لمجموعة الأضداد المتشكلة ضمن الذاكرة المناعية من اجل قيم مختلفة لـ $\rho$ حيث تم ضبط محددات الخوارزميات كالتالي : N=10 و n=3 و d=0 و k=3 و β=15 و gen=50 تم الاختبار عند عدة قيم لـ $\rho$. تم تكرار التجربة عشر مرات عند كل قيمة لـ $\rho$ و من ثم تم اخذ المتوسط للقيم الناتجة(قيمة متوسط التآلف). يبين الشكل (6-7) نتائج الاختبار حيث نلاحظ أن $\rho$ هو معامل حرج جداً لأداء الخوارزمية و مجرد تغيره بشكل ضئيل يؤثر تأثير كبير على سرعة الخوارزمية و جودة الحل الناتج . و بالتالي يجب ضبطه بشكل دقيق و جيد للحصول على الأداء الأمثل. يعتبر هذا المعامل من احد سيئات الخوارزمية لصعوبة تحديده بدقة.

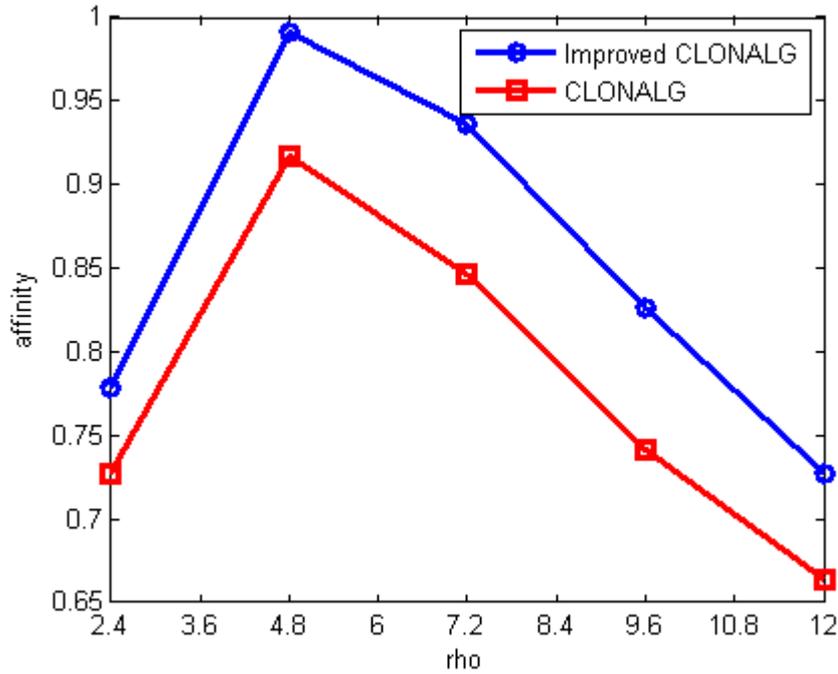

الشكل (7-6) التآلف مع معامل الهبوط

يمكن تلخيص نتائج الاختبارات على خوارزمية الانتقاء النسيلي في الجدول (1-6):

جدول (1-6) اثر تغيير معاملات خوارزمية الانتقاء النسيلي على أدائها

| الأثر على الخوارزمية | المعامل |
|---|---|
| زيادة عدد الأجيال تؤدي إلى ازدياد جودة الحل الناتج | gen عدد الأجيال |
| زيادة قيمة عتبة التآلف تؤدي إلى ازدياد الزمن اللازم للوصول للحل | $\varepsilon$ عتبة التآلف |
| زيادة قيمة معامل الاستنسال تؤدي إلى تقليل الزمن اللازم للوصول إلى لحل | $\beta$ معامل الاستنسال |
| زيادة عدد الخلايا المنتقاة للاستنسال تؤدي إلى تقليل الزمن اللازم للوصول إلى لحل | n عدد الخلايا المنتقاة للاستنسال |
| يكفي جعل k=1 لتسريع الخوارزمية بشكل كبير | K عدد الخلايا من المجموعة المؤقتة $C^*$ التي يتم إدخالها إلى $Ab_r$ |
| معامل الهبوط لمنحني احتمال الطفرة هو معامل حرج جداً يؤثر بشكل كبير على سرعة الخوارزمية و يصعب ضبطه | $\rho$ معامل الهبوط لمنحني احتمال الطفرة |

## 6-3- الاختبارات على aiNet:

الغاية من aiNet هو ضغط المعطيات و التخلص من المعطيات الفائضة و ذلك بغرض تسهيل معالجة هذه المعطيات دون ضياع المعلومة. بعد تطبيق aiNet و ضغط المعطيات يتم الحصول على العناقيد ضمن المعطيات (الحصول على المعلومة) بواسطة خوارزمية عنقدة هرمية.

قمنا باختبار aiNet على مجموعة من معطيات الاختبار الاصطناعية و كان الغرض هو التخلص من المعلومات الفائضة ثم عنقدة المعطيات. بالتالي من المهم أن تحتوي معطيات الاختبار على معلومات فائضة و إلا لا معنى من استخدام aiNet.

تم الاختبار على مجموعات المعطيات التالية:
1. مجموعة المعطيات الأولى عبارة عن حلزونين متداخلين و هي مستخدمة في [53] [54].
2. مجموعة المعطيات الثانية عبارة عن حلقتين متداخلتين في فضاء ثلاثي البعد و هي مستخدمة في [53] [54].
3. مجموعة المعطيات الثالثة قمنا بتوليدها على شكل دائرتين احدهما بداخل الأخرى.

يوضح الشكل (6-8) معطيات الاختبار و الشبكة الناتجة عن تطبيق aiNet.
تم ضبط محددات aiNet كالتالي:

- من اجل مجموعة المعطيات الأولى: $\sigma_d = 1$ و n=4 و $\zeta = 10$ و $\sigma_s = 0.07$ و عدد التكرارات N$_{gen}$=20.
- من اجل مجموعة المعطيات الثانية: $\sigma_d = 1$ و n=4 و $\zeta = 10$ و $\sigma_s = 0.2$ و عدد التكرارات N$_{gen}$=40.
- من اجل مجموعة المعطيات الثالثة: $\sigma_d = 1$ و n=4 و $\zeta = 10$ و $\sigma_s = 0.1$ و عدد التكرارات N$_{gen}$=40.

يوضح الجدول (6-2) نسب ضغط المعطيات الناتجة عن تطبيق aiNet على المعطيات حيث نلاحظ أن الخوارزمية قامت بالتخلص من المعطيات الفائضة بشكل فعال دون أن يضر ذلك بعملية التصنيف.

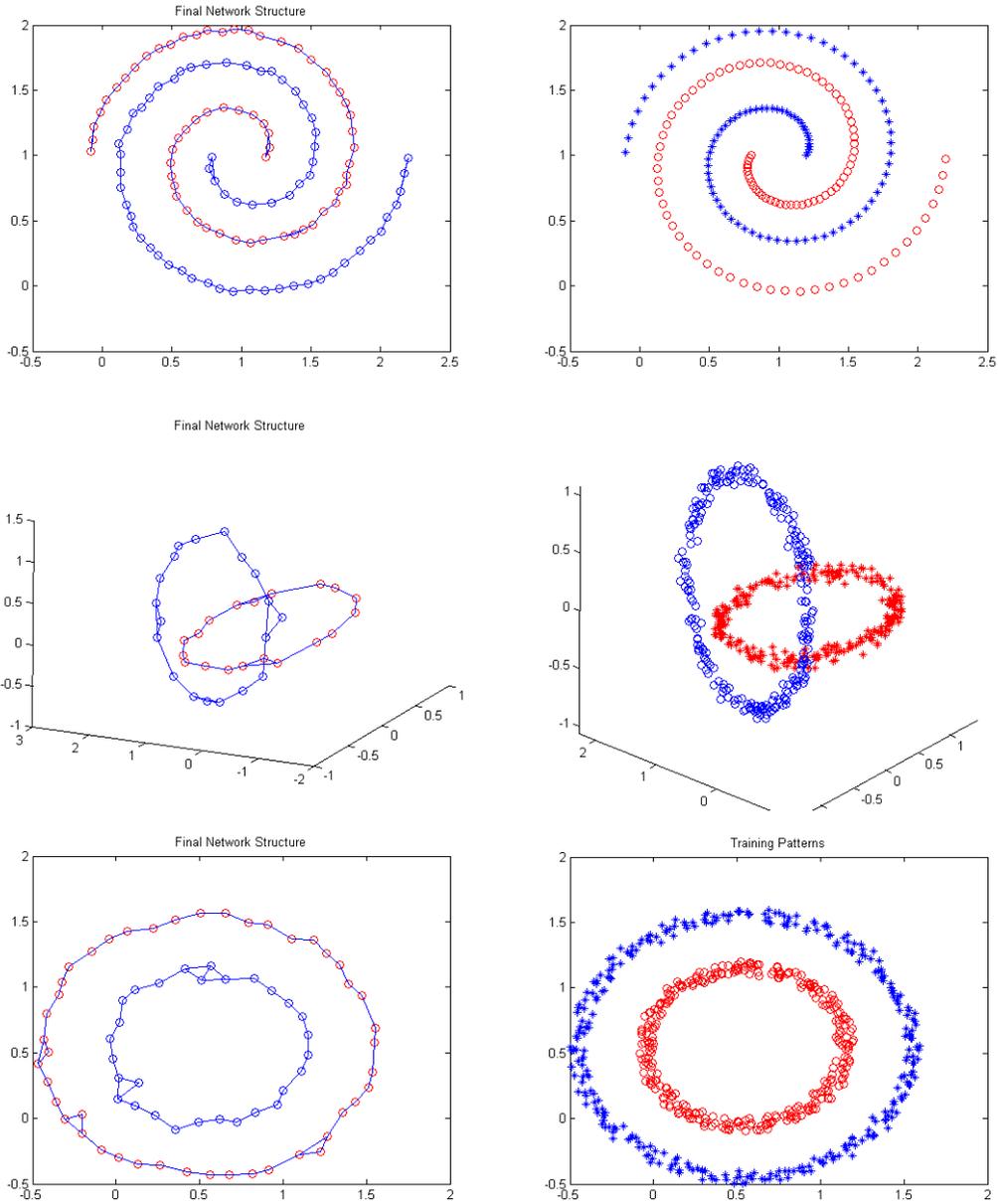

الشكل (8-6) معطيات الاختبار لـaiNet و الشبكة الناتجة

جدول (2-6) نسب الضغط الناتجة عن تطبيق aiNet

| مجموعة المعطيات | حجم المعطيات | عدد خلايا الشبكة الناتجة | نسبة الضغط |
|---|---|---|---|
| مجموعة المعطيات الأولى | 190 | 120 | 36.84% |
| مجموعة المعطيات الثانية | 500 | 35 | 93% |
| مجموعة المعطيات الثالثة | 630 | 70 | 88.89% |

كما قمنا بدراسة اثر تغيير معاملات aiNet على عمل الخوارزمية و كانت النتائج مطابقة تماماً لما تم الحصول عليه في [53] و يمكن تلخيص النتائج في الجدول التالي:

جدول (3-6) اثر تغيير معاملات خوارزمية aiNet على أدائها

| المعامل | الأثر على الخوارزمية |
|---|---|
| n عدد الخلايا المنتقاة للاستنسال | زيادة n عدد الخلايا المنتقاة للاستنسال تؤدي إلى تقليل عدد الأجيال اللازمة لتقارب الخوارزمية (التعلم) لكن تزيد من حجم الشبكة النهائية الناتجة |
| $\zeta$% نسبة الخلايا المنتقاة لتشكيل الذاكرة المناعية | زيادة $\zeta$ نسبة الخلايا المنتقاة لتشكيل الذاكرة المناعية لا تؤثر على حجم الشبكة النهائية الناتجة لكن تبطئ الخوارزمية و تزيد عدد الأجيال اللازمة للتقارب |
| $\sigma_d$ عتبة الموت الطبيعي | $\sigma_d$ عتبة الموت الطبيعي تقلل عدد الأضداد في الشبكة في الجيل الأول ثم لا تؤثر في الأجيال التالية |
| $\sigma_s$ عتبة التثبيط | $\sigma_s$ عتبة التثبيط تتحكم في حجم الشبكة النهائية الناتجة و هي معامل حرج جداً زيادة عتبة التثبيط قد تؤدي إلى ضغط شديد للمعطيات و ضياع المعلومة و تفسير خاطئ للعناقيد |

رغم قوة الخوارزمية في تحليل المعطيات و لكن لاحظنا من الاختبارات التي أجريناها انه لديها العديد من السيئات:

1. العدد الكبير من المحددات المضبوطة من قبل الإنسان.
2. حساسية الخوارزمية لبعض هذه المحددات مثل عتبة التثبيط التي تؤثر بشكل كبير على النتائج.
3. الحمل الحسابي الكبير للخوارزمية.

## 6-4- تصميم خوارزمية مناعية للقيام بعملية التصنيف غير الموجه للمعطيات:

بناءً على ما قمنا به في هذا البحث من دراسة للنظام المناعي الاصطناعي و النماذج و الخوارزميات المناعية المطورة قمنا بتصميم خوارزمية مناعية تعتمد على مبدأ الانتقاء النسيلي

في النظام المناعي البيولوجي للقيام بعملية تصنيف غير موجه للمعطيات. الخوارزمية الجديدة هي خوارزمية تصنيف الانتقاء النسيلي غير الموجه Unsupervised Clonal Selection Classification و سنرمز لها بـ UCSC.

للقيام بتصميم UCSC قمنا باعتماد الهيكلية التي تم شرحها في الفصل الثالث و ذلك للانتقال من مجال المسألة و هو التصنيف غير الموجه إلى مجال الخوارزمية المناعية. يوضح الشكل (6-9) منهجية التصميم و الاختبار التي اعتمدنا عليها.

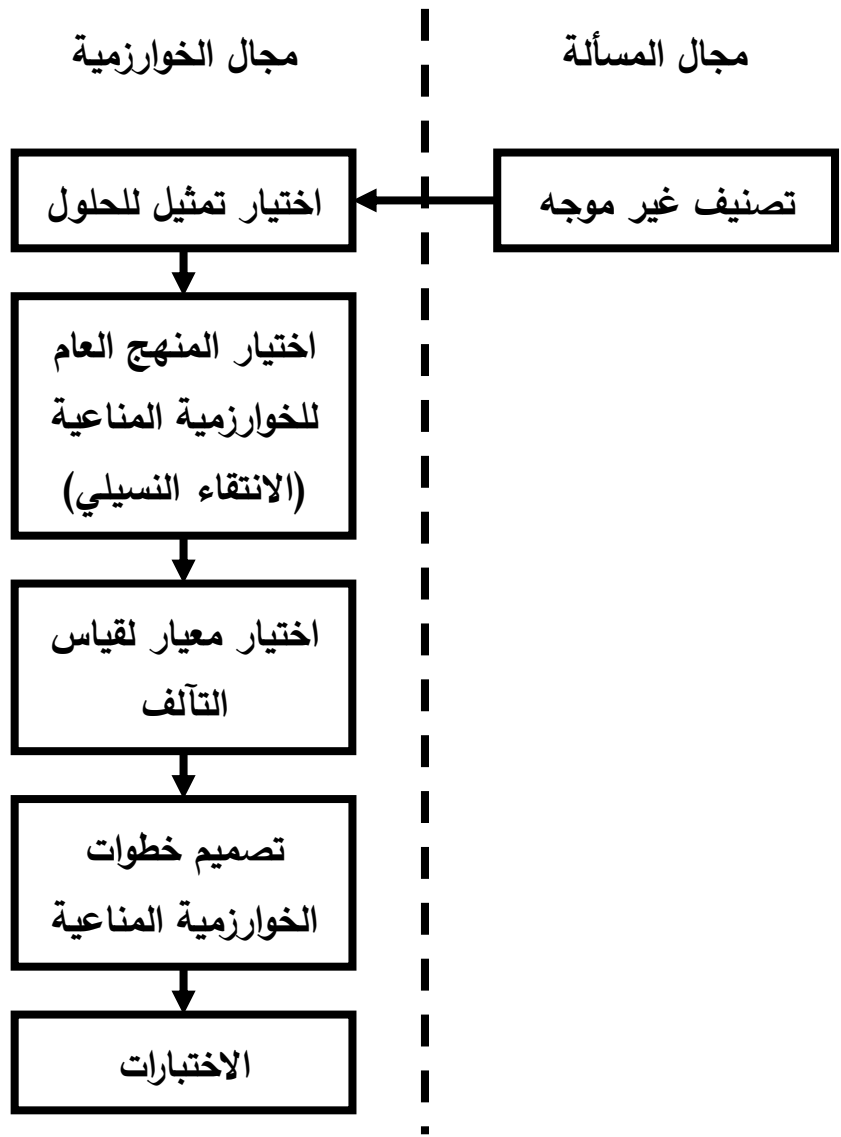

الشكل (6-9) منهجية تصميم UCSC و اختبارها

عملياً و أثناء التصميم فإن العمليات التي ستؤثر على أداء الخوارزمية بشكل كبير هي [62]:

1. تمثيل الحلول.
2. آلية التوسع النسيلي.
3. آلية الطفرة المعززة.
4. آلية توليد الحلول العشوائية.
5. تكوين الذاكرة المناعية (آلية انتقاء النخبة).

إن أي تغيير و لو كان صغيراً على هذه الآليات سيؤثر بشكل كبير على أداء الخوارزمية.

أثناء التصميم حاولنا تخفيض عدد محددات خوارزمية UCSC المضبوطة من قبل الإنسان و ذلك بربط معظم المحددات المهمة بالمعطيات التي يتم تصنيفها بحيث يتم تحديدها أوتوماتيكياً بناءً على المعطيات.

### 6-4-1- المبدأ الأساسي لـ UCSC:

تستخدم UCSC مبدأ العنقدة التقسيمية Partitional Clustering حيث ينظر إلى عملية التصنيف غير الموجه أو عملية التجميع Clustering على أنها عملية أمثلة optimization و الغاية من العملية تحقيق التقسيم الأمثل لمجموعات المعطيات بحيث تكون نتيجة العملية هي عناقيد من المعطيات المضغوطة قدر الإمكان و بحيث تحقق معيار معين للتماثل فيما بينها. معيار العنقدة المستخدم هو معيار لقياس مقدار الانتشار ضمن العنقود.

تم افتراض أن عدد العناقيد ضمن المعطيات معروف (عدد الأصناف) و هو $K$ و الغاية من الخوارزمية هي تحديد مراكز العناقيد $C_1, C_2, ..., C_K$ التي تحقق معيار العنقدة (الأمثلة) التالي: و هو تحقيق اقل مسافة بين المعطيات ضمن العنقود الواحد (الصنف) و مركز هذا العنقود , أي المسافة الداخلية ضمن العنقود اقل ما يمكن. بالتالي معيار العنقدة هو وفق المعادلة (6-1) حيث $x_j$ تمثل المعطيات و $m_i$ هي مركز العنقود $C_i$.

$$D = \sum_{i=1}^{k} \sum_{x_j \in C_i} \left\| x_j - m_i \right\| \quad (1-6)$$

و يصبح عمل خوارزمية الانتقاء النسيلي هو إيجاد مراكز العناقيد الأمثل الذي يجعل $D$ اقل ما يمكن.

### 6-4-2- خطوات خوارزمية UCSC:

تتبع UCSC الخطوات العامة لخوارزميات الانتقاء النسيلي من اجل عملية الأمثلة. خطوات الخوارزمية تتلخص كالتالي:

1. توليد مجموعة $P$ ذات العدد $n$ من الأضداد (مجموعة الحلول المقترحة) بشكل عشوائي.
2. من اجل كل جيل القيام بما يلي:
   2.1. قياس التآلف لكل الأضداد ضمن $P$.
   2.2. استنسال $P$ مولدين المجموعة $P_C$.
   2.3. تطبيق آلية الطفرة المعززة على $P_C$ مولدين المجموعة $P_m$.
   2.4. دمج المجموعتين $P$ و $P_m$.
   2.5. قياس التآلف لكل الأضداد ضمن المجموعة الجديدة.
   2.6. اختيار $n$ حل من المجموعة الجديدة ذات التآلف الأعلى لتشكل $P$.
   2.7. توليد $d$ حل جديد عشوائياً.
   2.8. استبدال الحلول ل ذات التآلف الأدنى في $P$ بالحلول الجديدة المولدة.
3. اختيار أعلى الأضداد تآلف من $P$ ليشكل الحل النهائي للخوارزمية.

### 6-4-3- تمثيل الحلول:

تم تمثيل كل ضد بمجموعة من الأرقام الحقيقية التي تمثل مراكز العناقيد. من اجل معطيات ضمن فضاء ذو $L$ بعد عندها يكون الضد هو مصفوفة مؤلفة من $L*K$ عنصر بحيث تمثل تلك العناصر إحداثيات المراكز مرتبة بالتتالي.

$$Ab=[Ab_1, Ab_2, \ldots, Ab_{L*k}]=[m_{11}, m_{12}, \ldots, m_{1L}, m_{21}, \ldots, m_{KL}]$$

بحيث تمثل أول $L$ عنصر إحداثيات المركز الأول و ثاني $L$ عنصر إحداثيات المركز الثاني و هكذا حتى المركز $K$. المثال التالي يوضح التمثيل المستخدم: من اجل فضاء ذو بعدين و $K=2$. ترميز الضد بالشكل $Ab=[0.1\ 2\ 0.5\ 3]$ يدل على المركزين $m1=[0.1\ 2]$ و $m2=[0.5\ 3]$.

### 6-4-4- معيار التآلف:

يتم حساب التآلف لكل ضد كالتالي يتم أولا تشكيل العناقيد وفقاً لمراكز العناقيد المرمزة ضمن الضد و يتم ذلك بإسناد كل نقطة من المعطيات إلى العنقود (الصنف) المقابلة لأقرب مركز من النقطة. بعد عملية العنقدة يتم حساب مراكز العناقيد المتشكلة من العملية و ذلك بحساب المتوسط لإحداثيات النقاط المشكلة للعنقود ، يليه حساب المسافة الداخلية لكل عنقود $D_{\text{intra}}^{(C_j)}$ وفق المعادلة (6-2) و عندها المسافة الكلية $D$ هي مجموع المسافات و عندها نحسب التآلف وفق العلاقة (6-4)

$$D_{\text{intra}}^{(C_j)} = \sum_{x_i \in C_j} \|x_i - m_j\| \qquad (2\text{-}6)$$

$$D = \sum_{j=1}^{k} D_{\text{intra}}^{(C_j)} \qquad (3\text{-}6)$$

$$aff = \frac{1}{D} \qquad (4\text{-}6)$$

بالتالي فأن التآلف الأعلى يكافئ المسافة الداخلية الأقل. في حال تشكل عناقيد خالية من المعطيات فانه يتم إسناد القيمة صفر إلى التآلف ليشير إلى أن الحل المقابل مرفوض.

### 6-4-5- الاستنسال:

تم اعتماد آلية الاستنسال المتبعة في CLONALG [50] حيث تم افتراض أن مجموعة الأضداد تم فرزها تنازلياً و ذلك وفقاً للتآلف و من ثم تم تطبيق المعادلة (5-6) لتحديد عدد النسخ المولدة عن كل ضد ضمن المجموعة.

$$nc_i = round(\frac{\beta \cdot n}{i}) \qquad (5\text{-}6)$$

حيث $\beta$ هو معامل الاستنسال. إن آلية الاستنسال المستخدمة تمثل مبدأ الانتقاء النسيلي.

### 6-4-6- آلية الطفرة المعززة:

يتم تعريض كل ضد ضمن المجموعة $P_c$ إلى آلية طفرة متناسبة عكساً مع مقدار التآلف و يتم ذلك وفقاً للمعادلات التالية:

$$Ab^* = Ab + \alpha\, N(0,1) \qquad (6\text{-}6)$$

$$\alpha = \rho\, e^{-aff} \qquad (7\text{-}6)$$

حيث $Ab^*$ هي الضد الناتج عن تطبيق الطفرة على الضد $Ab$. أما $N(0,1)$ هي مصفوفة من $L*K$ متحول غوصي بمتوسط قدره صفر و انحراف معياري يساوي الواحد. aff هو التآلف المقاس بعد أن يتم تقيسه ضمن المجال $[0,1]$.

المعامل $\alpha$ يعمل على تقييس مقدار الطفرة الغوصية الحاصلة من المتحول الغوصي و هو متناسب عكساً مع التآلف ، $\rho$ هو معامل سيحدد مجال $\alpha$. لتسريع عمل الخوارزمية و تسريع

وصولها للحل تم ربط الطفرة بالمجال الذي تشغله المعطيات ضمن الفضاء متعدد الأبعاد وفق المعدلة (6-8).

$$\rho = (\max_{data} - \min_{data})/10 \qquad (6-8)$$

حيث $\max_{data}$ و $\min_{data}$ تمثل اكبر قيمة و اصغر قيمة للخواص المشكلة للمعطيات و ذلك من اجل كل الأبعاد.

بالتالي أصبحت الطفرة متعلقة بمقدار التآلف و مجال البحث (المجال الذي تشغله المعطيات) الذي نبحث ضمنه فعلياً عن مراكز العناقيد و كلما زاد مجال البحث زاد مجال الطفرة لتسريع عملية البحث. و بالتجربة وجد أن هذا الربط سرع الخوارزمية و جعلها متكيفة مع المعطيات إذ ليس هناك داعي لضبط المعاملات المحددة للطفرة بل تحدد أوتوماتيكيا بناءً على المعطيات و التي تحدد فعلياً مجال البحث.

### 6-4-7- مولد الحلول العشوائية:

إن مولد الحلول العشوائية مهم جداً لضمان البحث الكامل عن الحل و لصيانة التنوع ضمن مجموعة الحلول. تم أولا تحديد مجال البحث (المجال الذي تشغله المعطيات) و ذلك بإيجاد القيم العليا و الدنيا للمعطيات على كل الأبعاد.

$$UL_{data} = [UL_1, UL_2, \cdots, UL_L] \qquad (6-9)$$

$$LL_{data} = [LL_1, LL_2, \cdots, LL_L] \qquad (6-10)$$

حيث $UL_i$ هي الحدود العليا للخواص المحددة للمعطيات و $LL_i$ هي الحدود الدنيا للخواص المحددة للمعطيات ، و بناءً على مجال البحث تم توليد الحلول الجديدة عشوائياً وفق المعادلة (6-11).

$$Ab_{new} = LL_{data} + (\text{diag}((UL_{data} - LL_{data})^T \times rand))^T \qquad (6-11)$$

حيث $rand$ هي مصفوفة من $L*K$ متحول عشوائي ذو توزع احتمالي متساوي ضمن المجال $[0,1]$. إن هذه الصيغة لمولد الحلول العشوائية تضمن سرعة و دقة الأداء للخوارزمية حيث يضمن التقارب السريع للخوارزمية لأن جميع الحلول المولدة هي ضمن مجال البحث.

### 6-5- الاختبارات و التجارب التي تمت على UCSC:

قمنا باختبار الخوارزمية على مجموعة من المعطيات الاصطناعية التي قمنا بتوليدها و على مجموعة من المعطيات الحقيقية المأخوذة من الواقع و تم مقارنة النتائج مع خوارزمية K-means [20] [24] المشروحة سابقاً.

سبب المقارنة مع خوارزمية K-means هو أنها تستخدم معيار الخطأ التربيعي كمعيار للعنقدة و هو مجموع مربع المسافات الأقليدية بين المعطيات ضمن العنقود الواحد (الصنف) و مركز هذا العنقود و فعلياً هذا المعيار يعمل على قياس مقدار الانتشار ضمن العنقود بالتالي هو مكافئ من حيث المبدأ للمعيار المستخدم من قبل UCSC مع الانتباه إلى أن المعيار المستخدم في UCSC هو مجموع المسافات الأقليدية و ليس مجموع مربع المسافات.

### 6-5-1- معطيات الاختبار:

### 6-5-1-1- المعطيات الاصطناعية:

لاختبار الخوارزمية قمنا بتشكيل ثلاث مجموعات من المعطيات الاصطناعية كالتالي:

- **مجموعة المعطيات الأولى:**

مؤلفة من صنفين متداخلين بحيث كل صنف مكون من 100 نمط ثنائي البعد ذات توزع غوصي ضمن الأصناف.

محددات توزع الأصناف هي, $m_1=(0.1, 0.1)$, $m_2=(0.35, 0.1)$

$\sum_1 = \sum_2 = \begin{bmatrix} 0.11 & 0 \\ 0 & 0.1 \end{bmatrix}$, حيث $m_2$, $m_1$ هي مراكز الأصناف و $\sum_2, \sum_1$ هي مصفوفات التغاير.

يوضح الشكل (6-10) مجموعة المعطيات.

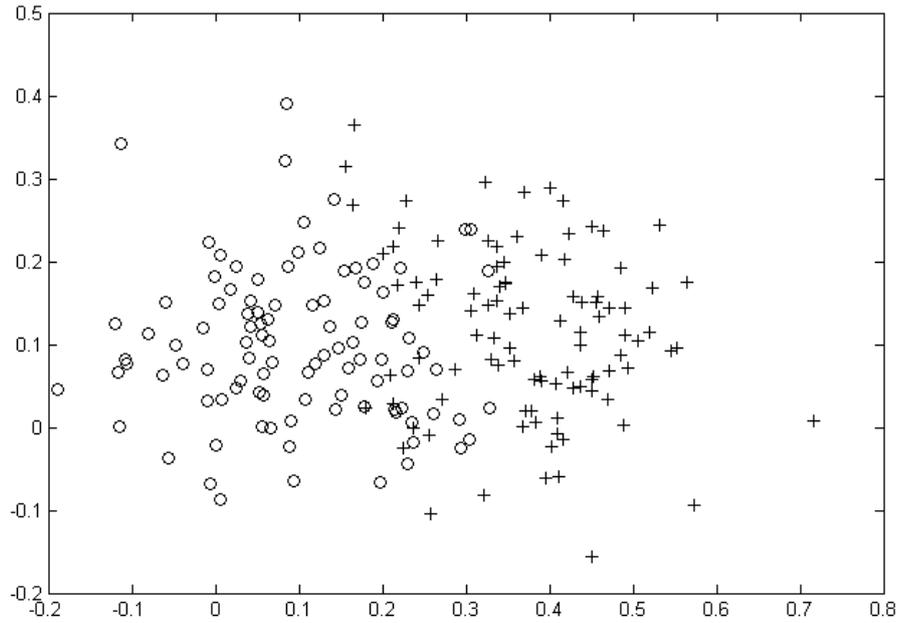

الشكل (6-10) مجموعة المعطيات الأولى

- **مجموعة المعطيات الثانية:**

مؤلفة من تسع أصناف بحيث كل صنف مكون من 25 نمط ثنائي البعد ذات توزع غوصي ضمن الأصناف.

محددات توزع الأصناف $m_1=(0.1, 0.1)$, $m_2=(0.1, 0.5)$, $m_3=(0.1, 0.9)$, $m_4=(0.5, 0.1)$, $m_5=(0.5, 0.5)$, $m_6=(0.5, 0.9)$, $m_7=(0.9, 0.1)$, $m_8=(0.9, 0.5)$, $m_9=(0.9, 0.9)$

$$\sum\nolimits_1 = \sum\nolimits_2 = \sum\nolimits_3 = \ldots = \sum\nolimits_9 = \begin{bmatrix} 0.08 & 0 \\ 0 & 0.08 \end{bmatrix}$$

حيث $m_i$ هي مراكز الأصناف و $\sum\nolimits_i$ هي مصفوفات التغاير.

يوضح الشكل (6-11) مجموعة المعطيات.

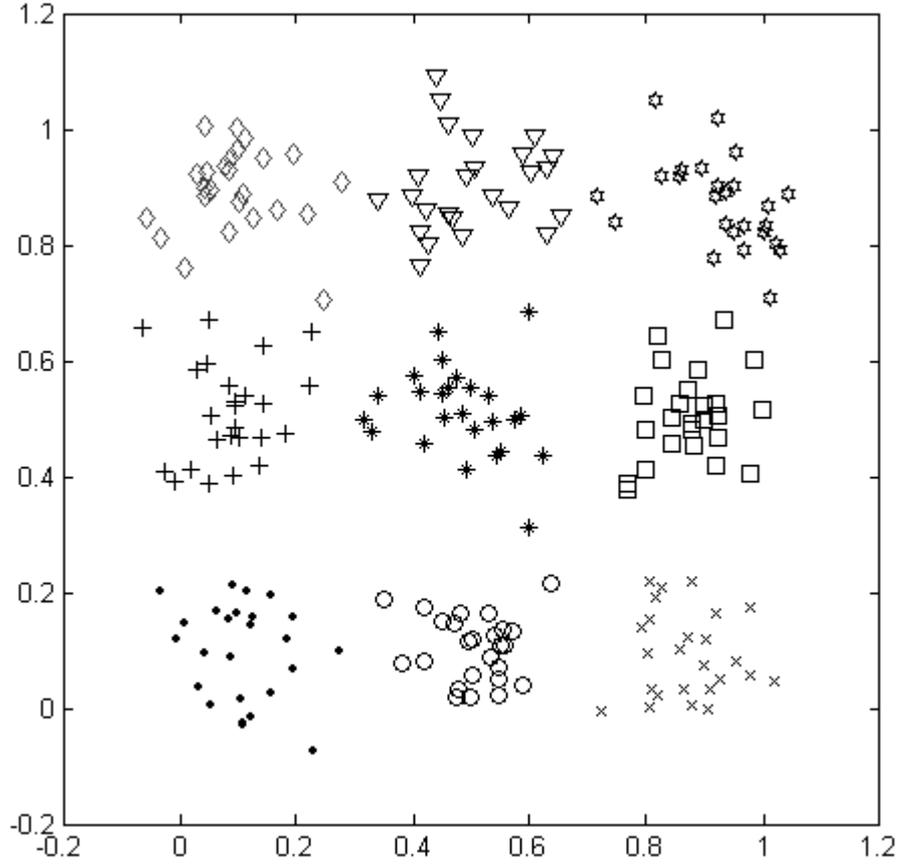

الشكل (6-11) مجموعة المعطيات الثانية

- **مجموعة المعطيات الثالثة:**

مؤلفة من ثلاث أصناف بحيث كل صنف مكون من 50 نمط ثلاثي البعد ذات توزع غوصي ضمن الأصناف.

محددات توزع الأصناف $m_1=(1, 1, 1)$, $m_2=(2, 2.5, 2.5)$, $m_3=(2, 3, 3)$

$\sum_1 = \sum_2 = \sum_3 = \begin{bmatrix} 0.3 & 0 & 0 \\ 0 & 0.3 & 0 \\ 0 & 0 & 0.3 \end{bmatrix}$ حيث $m_i$ هي مراكز الأصناف و $\sum_i$ هي مصفوفات التغاير.

يوضح الشكل (6-12) مجموعة المعطيات.

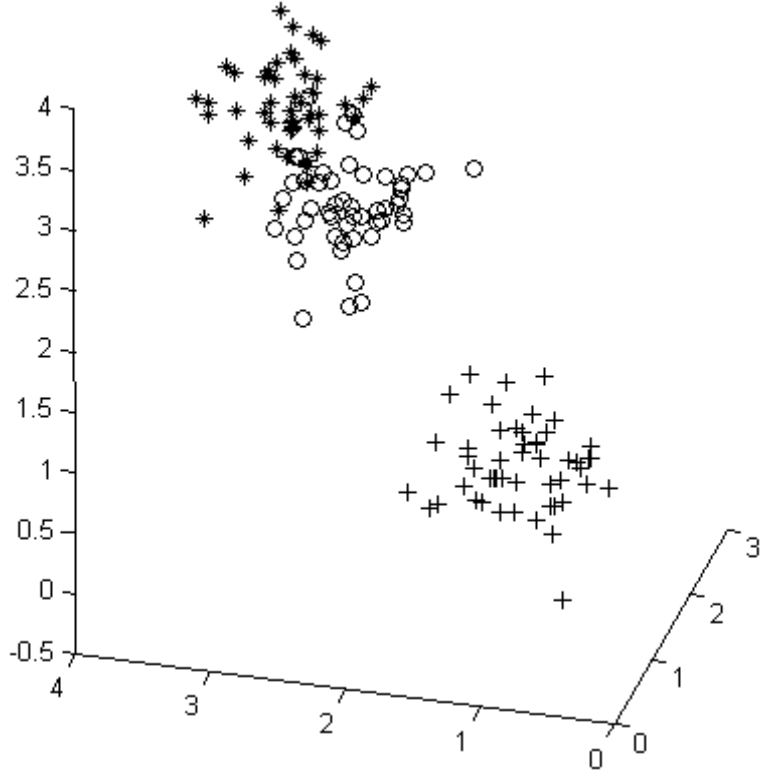

الشكل (6-12) مجموعة المعطيات الثالثة

### 6-5-1-2- المعطيات الحقيقية:

- **مجموعة المعطيات Iris**:

مؤلفة من ثلاث أصناف [66] كل صنف مؤلف من 50 نمط رباعي البعد تمثل الأصناف ثلاث أنواع مختلفة من زهرة السوسن و كل نمط مؤلف من أربع خواص (واصفات) للزهرة و الخواص هي : طول الكأس و عرض الكأس و طول التويجة و عرض التويجة بالسنتيمتر. أسماء أنواع زهرة السوسن هي Iris Virginica , Iris Versicolor, Iris Setosa.

- **مجموعة معطيات لسرطان الثدي Wisconsin Breast Cancer**:

و هي مؤلفة من صنفين [67] ورم حميد Benign و ورم خبيث Malignant الصنف الأول مؤلف من 458 نمط و الصنف الثاني مؤلف من 241 نمط. الأنماط ذات تسع أبعاد كل بعد يمثل خاصية الخواص هي: سماكة التكتل , التجانس في حجم الخلايا , التجانس في شكل الخلايا , التلاصق الحدي , حجم الخلية الظهارية الوحيدة , النواة العارية , الكروماتين الخفيف , التفتلات , النواة الطبيعية.

## 6-5-2- نتائج الاختبارات:

عند اختبار UCSC تم ضبط معاملاتها كما يلي:

$n=10, β =5, d=4, gen=20$ تم إعادة اختبار الخوارزميات 100 مرة و ذلك من اجل مجموعة حلول أولية مختلفة كل مرة و الغرض من تكرار الاختبارات هو القيام بدراسة إحصائية لقدرة الخوارزميات على إيجاد الحل. النتائج موضحة في الجداول.

يوضح الجدول (6-4) أفضل الحلول التي وصلت إليها كل خوارزمية ممثلة بمعيار العنقدة المستخدم.

جدول(6-4) قيم $D$ التي حصلت عليها الخوارزميات من اجل مجموعات المعطيات

| Dataset | UCSC | *K*-means |
|---|---|---|
| Dataset 1 | 25.141 | 25.166 |
| Dataset 2 | 21.597 | 21.906 |
| Dataset 3 | 70.628 | 70.653 |
| Iris dataset | 97.101 | 97.205 |
| Breast Cancer dataset | 3048.2 | 3051.3 |

و كما نرى من الجدول فان UCSC استطاعت الحصول على حل أفضل من الحل الذي حصلت عليه *K*-means و ذلك من اجل جميع معطيات الاختبار.

توضح الجداول التالية نتائج التصنيف (أفضل دقة تصنيف تم الحصول عليها خلال تكرار الاختبارات 100 مرة):

جدول(6-5) نتائج التصنيف التي حصلت عليها *K*-mean من اجل مجموعة المعطيات الأولى

|  | Class 1 | Class 2 |
|---|---|---|
| **Accuracy** | 77% | 95% |
| **Overall accuracy** | 86% ||
| **Value of *D*** | 25.166 ||

جدول(6-6) نتائج التصنيف التي حصلت عليها UCSC من اجل مجموعة المعطيات الأولى

|  | Class 1 | Class 2 |
|---|---|---|
| **Accuracy** | 86% | 90% |
| **Overall accuracy** | 88% ||
| **Value of *D*** | 25.141 ||

نلاحظ من النتائج الموضحة في الجدولين السابقين أن الدقة الكلية للتصنيف التي حصلت عليها UCSC من اجل مجموعة المعطيات الأولى هي أعلى من الدقة الكلية التي حصلت عليها K-mean حيث حصلت UCSC على دقة أعلى من اجل الصنف الأول لكن على دقة اقل من اجل الصنف الثاني.

جدول(6-7) نتائج التصنيف التي حصلت عليها K-mean من اجل مجموعة المعطيات الثانية

|  | Class 1 | Class 2 | Class 3 | Class 4 | Class 5 | Class 6 | Class 7 | Class 8 | Class 9 |
|---|---|---|---|---|---|---|---|---|---|
| **Accuracy** | 100% | 96% | 96% | 100% | 92% | 100% | 100% | 100% | 92% |
| **Overall accuracy** | 97.33% | | | | | | | | |
| **Value of D** | 21.609 | | | | | | | | |

جدول(6-8) نتائج التصنيف التي حصلت عليها UCSC من اجل مجموعة المعطيات الثانية

|  | Class 1 | Class 2 | Class 3 | Class 4 | Class 5 | Class 6 | Class 7 | Class 8 | Class 9 |
|---|---|---|---|---|---|---|---|---|---|
| **Accuracy** | 100% | 96% | 100% | 100% | 92% | 100% | 100% | 100% | 92% |
| **Overall accuracy** | 97.78% | | | | | | | | |
| **Value of D** | 21.597 | | | | | | | | |

نلاحظ من النتائج الموضحة في الجدولين السابقين أن الدقة الكلية للتصنيف التي حصلت عليها UCSC من اجل مجموعة المعطيات الثانية هي أعلى من الدقة الكلية التي حصلت عليها K-mean حيث حصلت UCSC على دقة أعلى من اجل الصنف الثالث.

جدول(6-9) نتائج التصنيف التي حصلت عليها K-mean من اجل مجموعة المعطيات الثالثة

|  | Class 1 | Class 2 | Class 3 |
|---|---|---|---|
| **Accuracy** | 100% | 84% | 90% |
| **Overall accuracy** | 91.33% | | |
| **Value of D** | 70.653 | | |

جدول(6-10) نتائج التصنيف التي حصلت عليها UCSC من اجل مجموعة المعطيات الثالثة

|  | Class 1 | Class 2 | Class 3 |
|---|---|---|---|
| **Accuracy** | 100% | 84% | 90% |
| **Overall accuracy** | 91.33% | | |
| **Value of** $D$ | 70.628 | | |

نلاحظ من النتائج الموضحة في الجدولين السابقين أن كلاً من UCSC و K-mean حصلتا على نفس الدقة الكلية من اجل مجموعة المعطيات الثالثة.

جدول(6-11) نتائج التصنيف التي حصلت عليها K-mean من اجل مجموعة المعطيات Iris

|  | Setosa | Versicolor | Virginica |
|---|---|---|---|
| **Accuracy** | 100% | 96% | 72% |
| **Overall accuracy** | 89.33% | | |
| **Value of** $D$ | 97.205 | | |

جدول(6-12) نتائج التصنيف التي حصلت عليها UCSC من اجل مجموعة المعطيات Iris

|  | Setosa | Versicolor | Virginica |
|---|---|---|---|
| **Accuracy** | 100% | 96% | 74% |
| **Overall accuracy** | 90% | | |
| **Value of** $D$ | 97.101 | | |

نلاحظ من النتائج الموضحة في الجدولين السابقين أن الدقة الكلية للتصنيف التي حصلت عليها UCSC من اجل مجموعة المعطيات Iris هي أعلى من الدقة الكلية التي حصلت عليها K-mean حيث حصلت UCSC على دقة أعلى من اجل الصنف الثالث (**Virginica**).

جدول(6-13) نتائج التصنيف التي حصلت عليها K-mean
من اجل مجموعة المعطيات لسرطان الثدي

|  | Benign | Malignant |
|---|---|---|
| **Accuracy** | 98.03% | 93.36% |
| **Overall accuracy** | 95.7% | |
| **Value of** $D$ | 3051.3 | |

جدول(6-14) نتائج التصنيف التي حصلت عليها UCSC

من اجل مجموعة المعطيات لسرطان الثدي

|  | **Benign** | **Malignant** |
|---|---|---|
| **Accuracy** | 98.03% | 94.19% |
| **Overall accuracy** | 96.11% | |
| **Value of *D*** | 3048.2 | |

نلاحظ من النتائج الموضحة في الجدولين السابقين أن الدقة الكلية للتصنيف التي حصلت عليها UCSC من اجل مجموعة المعطيات لسرطان الثدي هي أعلى من الدقة الكلية التي حصلت عليها K-mean حيث حصلت UCSC على دقة أعلى من اجل الصنف الثاني (الورم الخبيث).

النتائج السابقة هي من اجل أفضل الحلول التي تم الحصول عليها أثناء إعادة تنفيذ الخوارزميات 100 مرة و الفكرة المهمة جداً هي أن خوارزمية K-means لم تحصل على تلك النتائج الجيدة الموضحة في الجداول في كل التجارب بل أعطت عدة حلول محلية غير جيدة أما UCSC فقد أعطت نفس النتائج من اجل كل التجارب التي اختبرت فيها. يوضح الجدولين (6-15) و (6-16) نسب تكرار أفضل الحلول التي تم الحصول عليها من كلا الخوارزميتين أثناء التنفيذ 100 مرة.

جدول(6-15)نسب تكرار أفضل الحلول أثناء تنفيذ *K*-mean

| **Dataset** | ***D*** | **percents** |
|---|---|---|
| Dataset 1 | 25.166 | 100% |
| Dataset 2 | 21.906 | 40% |
| Dataset 3 | 70.653 | 75% |
| Iris dataset | 97.205 | 80% |
| Breast Cancer dataset | 3051.3 | 100% |

جدول(6-16)نسب تكرار أفضل الحلول أثناء تنفيذ UCSC

| **Dataset** | ***D*** | **percents** |
|---|---|---|
| Dataset 1 | 25.141 | 100% |
| Dataset 2 | 21.597 | 100% |
| Dataset 3 | 70.628 | 100% |
| Iris dataset | 97.101 | 100% |
| Breast Cancer dataset | 3048.2 | 100% |

و كما نرى من النتائج فان خوارزمية *K*-means قامت بإيجاد الحل الجيد ولكن بنسب صغيرة و ذلك في مجموعات المعطيات الثانية و الثالثة و Iris أي أنها كانت تعلق ضمن حلول محلية حتى من اجل المعطيات البسيطة مثل المجموعة الثانية. أما UCSC أعطت الحل الجيد في كل التجارب بالتالي فإن UCSC هي أكثر وثوقية من *K*-means.

توضح الأشكال التالية الحلول التي أعطتها *K*-means من اجل مجموعات المعطيات الثانية و الثالثة و Iris و مقدار تكرارها.

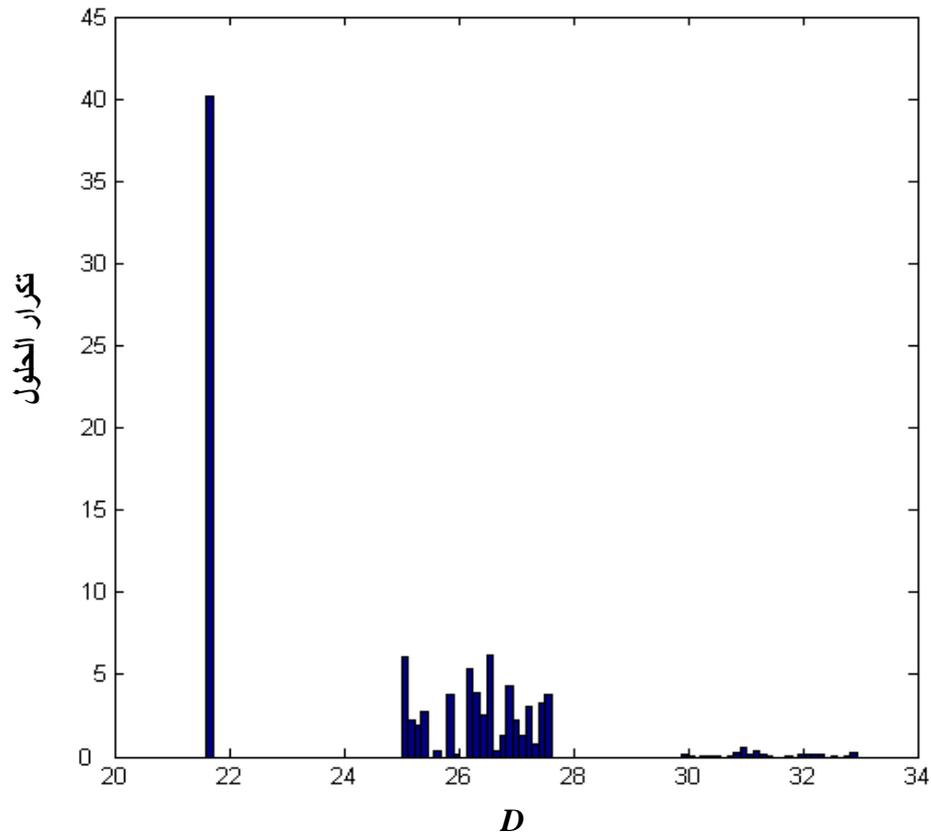

شكل(6-13) مخطط تكرار الحلول من اجل مجموعة المعطيات الثانية

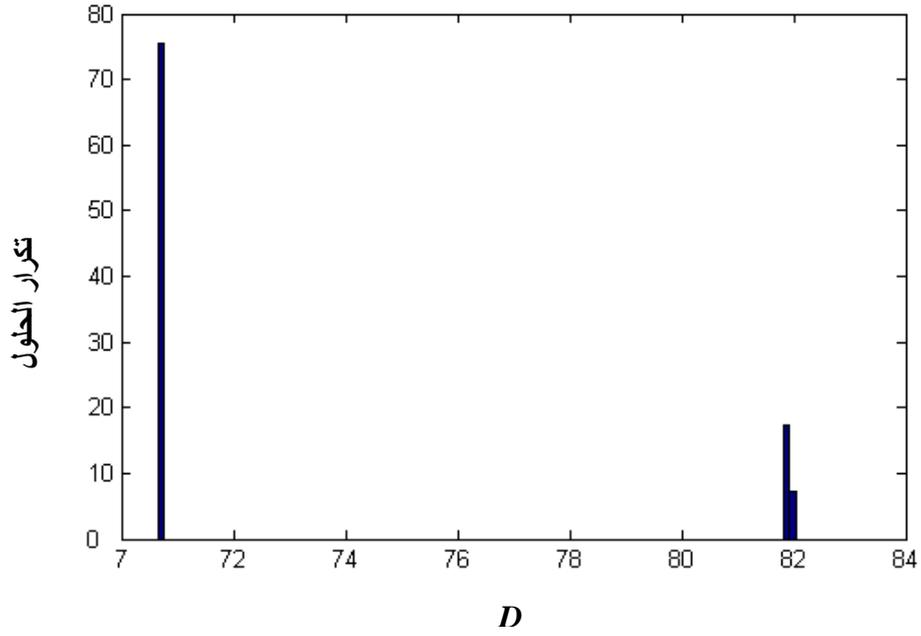

شكل(6-14) مخطط تكرار الحلول من اجل مجموعة المعطيات الثالثة

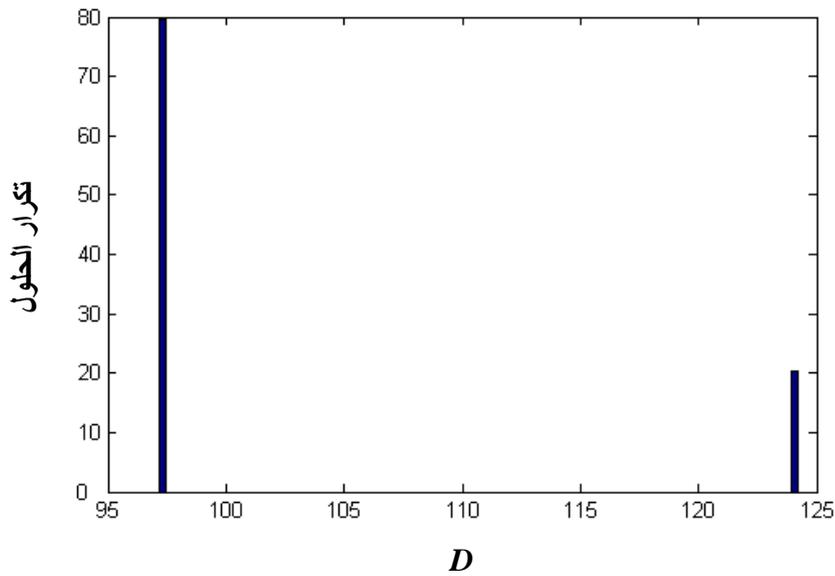

شكل(6-15) مخطط تكرار الحلول من اجل مجموعة المعطيات Iris

من الأمور المهمة التي يجب ملاحظتها أن UCSC أعطت الحل المطلوب خلال 20 جيل فقط و هذا يدل على أنها من أسرع الخوارزميات التطورية . كما أن حجم مجموعة الحلول المستخدم هو n=10 و هو أمر مهم لتقليل الحمل الحسابي حيث أن معظم الخوارزميات التطورية تحتاج إلى حجم مجموعة حلول لا يقل عن 100 للتمكن من إيجاد الحل.

## 6-6- مميزات UCSC:

يمكن تلخيص مميزات UCSC المستخلصة من التجارب كالتالي:
1. استخدام ترميز حقيقي للحلول.
2. متكيفة مع المعطيات و تعدل معاملاتها أوتوماتكياً لتسريع الوصول للحل.
3. دقة تصنيف اكبر.
4. وثوقية عالية في إيجاد الحل.
5. عدد أجيال قليل للوصول للحل.
6. حجم مجموعة حلول صغير.

## 6-7- سيئة UCSC :

سيئة UCSC هو الحمل الحسابي الكبير لتنفيذها حيث تستغرق الخوارزمية زمناً طويلا نسبياً للوصول إلى الحل و ذلك مقارنة مع خوارزميات التصنيف المماثلة مثل K-means. لدراسة زمن تنفيذ الخوارزمية تم استخدام MATLAB Profiler و تم ضبط عدد الأجيال على جيل واحد فقط و تم الاختبار على مجموعة المعطيات الثانية (ذات الفضاء المؤلف من 9 أبعاد) بالنتيجة استغرق تنفيذ UCSC 1.625s و هو زمن طويل. حيث وجد أن الحمل الحسابي الكبير للخوارزمية ناتج عن تابع حساب التآلف حيث حجز 99% من زمن التنفيذ و كان الزمن اللازم له على مدار الخوارزمية هو 1.609s حيث استدعي هذا التابع أربع مرات أثناء التنفيذ. تم الاختبار على حاسب P4 بسرعة 1600MHz.

## 6-8- الخاتمة:

احتوى هذا الفصل على الاختبارات التي أجريناها على خوارزمية الانتقاء النسيلي و خوارزمية الشبكة المناعية الاصطناعية حيث تم تلخيص نتائج الاختبارات و اثر تغيير معاملات الخوارزميات على أدائها ضمن جداول ، ثم تم شرح خوارزمية UCSC التي قمنا بتصميمها للقيام بعملية التصنيف غير الموجه للمعطيات ، إضافة إلى الاختبارات التي أجريناها عليها حيث تم مقارنة أدائها مع خوارزمية K-means و هي من أكثر الخوارزميات استعمالا في هذا المجال و كانت نتائج خوارزمية UCSC المقترحة هي الأفضل.

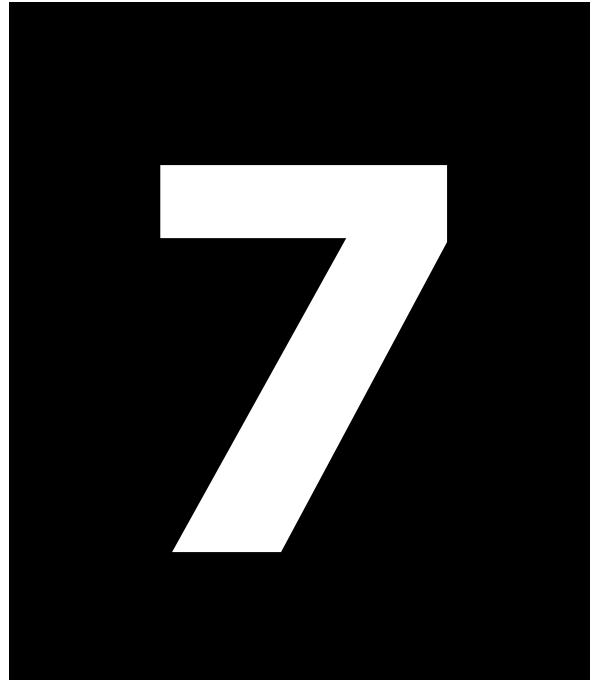

# الخاتمة و الأعمال المستقبلية

# 7- الخاتمة و الأعمال المستقبلية:

## 7-1- الخاتمة:

قدم هذا البحث دراسة لأحدث التقنيات في مجال الذكاء الحوسبي و هي النظام المناعي الاصطناعي إذ قدمت الرسالة فكرة عن ماهية النظام المناعي و كيفية الاستفادة منه في تصميم خوارزميات حاسوبية في مجال تمييز الأنماط ، حيث عرضت الرسالة عدداً من أهم النماذج و الخوارزميات المناعية المستخدمة في مجال تمييز الأنماط و منها: خوارزمية الانتقاء السلبي و خوارزمية الانتقاء النسيلي و الشبكات المناعية الاصطناعية. تضمن البحث شرحاً كاملاً لتلك الخوارزميات و تطبيقاتها و التحسينات التي تمت عليها ، كما قمنا أيضاً ضمن إطار هذا البحث باختبار تلك الخوارزميات المناعية مثل خوارزمية الانتقاء النسيلي و الشبكة المناعية الاصطناعية إضافة إلى دراسة اثر تغير معاملات تلك الخوارزميات على أداء الخوارزمية و مدى أهمية هذه المعاملات في تحسين الأداء. و بناءً على دراسة النظام المناعي الاصطناعي و النماذج و الخوارزميات المناعية المطورة تم تصميم خوارزمية مناعية تعتمد على مبدأ الانتقاء النسيلي في النظام المناعي البيولوجي للقيام بعملية تصنيف غير موجه للمعطيات.

في الفصل الثاني تم شرح النظام المناعي البيولوجي و مكوناته و طبقات الحماية التي يوفرها ومن ثم آليات عمله و النظريات الأساسية للاستجابة المناعية حيث شكل مدخلاً إلى الفصل الثالث الذي يعطي فكرة عن النظام المناعي الاصطناعي و مجالات استخدامه و كيفية تصميم نظام مناعي اصطناعي لتطبيق معين ، في الفصل الرابع قمنا بشرح الطرق الأساسية لتمييز الأنماط ، في الفصل الخامس تمت دراسة الخوارزميات المناعية المستخدمة في مجال تمييز الأنماط ، في الفصل السادس تم وضع نتائج الاختبارات التي قمنا بها على الخوارزميات المناعية المستخدمة في مجال تمييز الأنماط و من ثم قمنا بشرح الخوارزمية المناعية التي قمنا بتصميمها لتقوم بعملية التصنيف غير الموجه كما تم اختبار أدائها و مقارنته مع أداء الخوارزميات المستعملة في هذا المجال حيث أعطت الخوارزمية الجديدة نتائج جيدة.

قمنا باختبار أداء الخوارزمية الجديدة UCSC و مقارنتها مع احد الخوارزميات المعروفة في مجال التصنيف غير الموجه و هي K-means حيث وجدنا من نتائج الاختبار أن الخوارزمية الجديدة كانت أكثر دقة في التصنيف و أكثر وثوقية.

تتلخص مميزات UCSC المستخلصة من التجارب فيما يلي: استخدام ترميز حقيقي للحلول ، متكيفة مع المعطيات و تعدل معاملاتها أوتوماتكياً لتسريع الوصول للحل ، دقة تصنيف اكبر ،

وثوقية عالية في إيجاد الحل ، عدد أجيال قليل للوصول للحل ، حجم مجموعة حلول صغير. لكن سيئة UCSC الوحيدة هي الحمل الحسابي الكبير اللازم لتنفيذها.

مجالات استخدام UCSC: إن الخوارزمية صممت لتكون خوارزمية تصنيف عامة تصلح لأكثر من تطبيق و ليس لتطبيق مخصص لكن حسب معيار العنقدة المستخدم في UCSC و الذي هو فعلياً معيار يقيس مقدار الانتشار ضمن العنقود فإن UCSC تلائم الحالات ذات الأصناف المفصولة عن بعضها بشكل جيد و تكون فيها الأصناف (العناقيد) مضغوطة قدر الإمكان و تأخذ شكل شبه كروي في الفضاء متعدد الأبعاد. إذاً لضمان عمل الخوارزمية و ملاءمتها لتطبيق ما يجب العمل على انتقاء و استخراج خواص من المسألة بحيث تكون العناقيد مفصولة بشكل جيد و مضغوطة بشكل شبه كروي. من أهم المجالات التي يمكن تطبيق خوارزمية UCSC عليها هو مجال تحليل صور الأقمار الصناعية متعددة الأطياف.

## 7-2- الأعمال المستقبلية:

أظهرت الخوارزمية الجديدة UCSC أداء جيد و يمكن تحسين هذا الأداء و تطوير الخوارزمية. هناك العديد من المجالات لتطوير خوارزمية UCSC نذكر منها:

1. استخدام أفكار الشبكة المناعية ضمن الخوارزمية إضافة إلى مبدأ الانتقاء النسيلي المستخدم و ذلك بإدخال فكرة التفاعل بين الحلول الناتجة و تفعيل و تثبيط قسم من الحلول.

2. محاولة حل مشكلة الحمل الحسابي الكبير للخوارزمية و يمكن ذلك بدراسة إمكانية تنفيذها بشكل تفرعي و تخفيض الحسابات أثناء الجيل الواحد و تسريع الخوارزمية قدر الإمكان.

3. من الأمور المهمة التي يمكن دراستها هي إمكانية استخدام معايير عنقدة أخرى حيث استخدمت UCSC مجموع المسافات الأقليدية بين المعطيات ضمن العنقود الواحد (الصنف) و مركز هذا العنقود و يمكن تطبيق معايير أخرى و دراسة اثر تغير معيار العنقدة على نتائج عملية التصنيف.

4. يمكن دمج تقنيات أخرى مع خوارزمية UCSC مثل المنطق العائم و ذلك بتغير معيار العنقدة لتصبح عملية العنقدة أو التقسيم بين المعطيات و فق مبادئ المنطق العائم.

بما يخص مجال النظام المناعي الاصطناعي AIS ما تزال هذه التقنية في بداياتها و هي تشكل مجالاً واسعاً للبحث و ما يزال البحث مستمراً في هذه النظام و آليات تطبيقه في مختلف المجالات. و يمكن حصر إمكانيات البحث في مجال AIS بما يلي:

1. اختبار و تحسين وتطوير الخوارزميات المناعية المعمولة سابقاً و دراسة إمكانية استخدامها من اجل تطبيق محدد.
2. دراسة استخدام آليات مناعية جديدة لم تستعمل بعد في الخوارزميات المناعية بغرض تطوير خوارزميات مناعية جديدة.
3. دراسة إمكانية دمج الخوارزميات المناعية مع تقنيات الذكاء الحوسبي الأخرى مثل الشبكات العصبونية و الأنظمة العائمة بغرض تحسين أداء تلك الأنظمة.

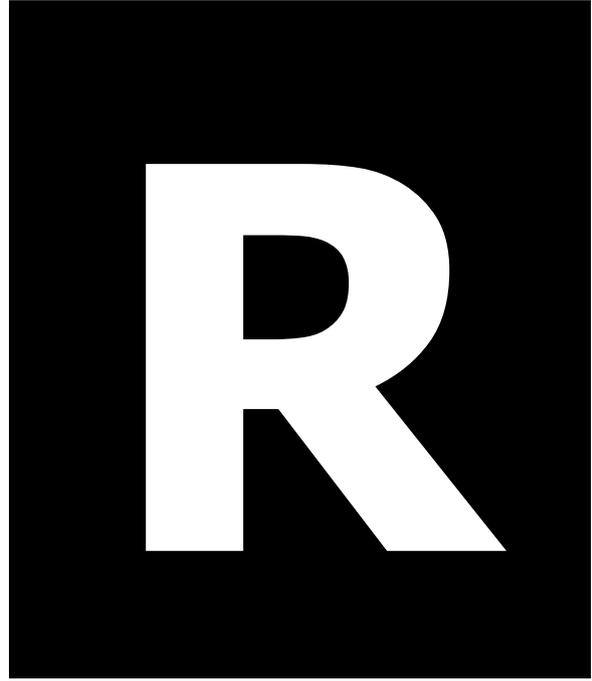

# المراجع

# المصطلحات

| | |
|---|---|
| adaptive | تلاؤمي , متكيف |
| affinity | ألفة , تآلف |
| anergy | استعطال |
| antibody | الضد , الجسم المضاد |
| antigen | مستضد , مولد الضد |
| apoptosis | استماتة |
| associative | ترابطي |
| autoaggresive | عدوانية ذاتية |
| bind | يرتبط |
| binding site | مقر رابط |
| bone marrow | نقي العظم |
| cancerous | سرطاني |
| chromosome | صبغي |
| chronic | مزمن |
| clonal | نسيلي |
| clonal expansion | التوسع النسيلي |
| clonal selection | الانتقاء النسيلي |
| complement system | جملة المتممة |
| conflict | صراع |
| cytokines | سيتوكينات |
| cytotoxic | سام للخلايا |
| dendritic cell | خلية تغصنية |
| differentiation | تمايز |
| elitism | سيطرة النخبة |
| endocrine system | جهاز الغدد الصم |
| epitope | حاتمة |
| exogenous infection | العدوة ذات المنشأ الخارجي |
| fitness | الكفاءة , الجودة |
| genome | مجين , مجموع الجينات في الكائن |
| germinal centers | المراكز المنتشة |
| granule cell | خلية حبيبية |
| granulocyte | الخلية المحببة |
| histocompatibility | توافق نسيجي |
| hypermutation | الطفرة المعززة |
| idiotope | مكنان |
| idiotype | نمط ذاتي |
| idiotypic | اختلاف الأنماط الذاتية |
| immature | غير ناضج |
| immune repertoire | الذخيرة المناعية |
| immune system | النظام المناعي |
| immunocompetent | مؤهل مناعياً |
| immunogen | مستمنع |
| immunoglobulin | غلوبولين مناعي |
| induce | محدث , محرض |
| ingestion | ابتلاع |
| innate | فطري , خلقي |
| lesion | آفة |
| leukocyte | الكرية البيضاء |
| ligand | لجين |
| lymphocyte | الخلة اللمفاوية |
| lymphoid organs | أعضاء لمفية |
| lymphokines | لمفوكينات |
| macrophage | بالعات كبيرة |
| maturation | نضوج |
| membrane | غشاء |
| metadynamic | الديناميكية المتبادلة |
| metaphors | مجاز |
| microbe | مكروب |
| microorganism | مكروب |
| molecules | جزيئات |
| monocyte | الوحيدة |
| multimodal | متعدد الأنماط |
| natural killer cell | خلية فاتكة طبيعية |
| neutralize | يحيد |
| niching method | طرق التعايش |
| offspring | النسل , الذرية |
| organism | متعضي , الكائن الحي |
| paratope | مستوقع |
| pathogen | ممراض , العامل الممرض |
| peptide | بيبتيد |
| periphery | محيط |
| phagocyte | بلعمية |
| plasmocyte | خلية بالزماوية |
| primed | مبرمج |
| proliferation | تكاثر |
| promote | يعزز |
| resting cell | خلية راقدة |
| scavenger cell | خلية كاسحة |
| secrete | يفرز |
| secretion | إفراز |
| selective theory | نظرية الانتقاء |
| self-antigen | مستضد ذاتي |
| self-tolerance | تحمل ذاتي |
| soluble | ذواب , قابل للذوبان |
| somatically | جسدي |
| specificity | مختص |
| stimulate | ينبه |
| suppression | الكبت |
| suppressor | كابت |
| thymus | تيموس |
| tumors | أورام |
| vaccination | لقاح |
| vertebrate | فقاري |
| viral | فيروسي |